\documentclass[11pt]{article}

\usepackage{arxiv}
\usepackage{natbib}
\usepackage{amsmath}
\usepackage{amsthm}
\usepackage{hyperref}
\newtheorem{definition}{Definition}
\newtheorem{theorem}{Theorem}
\newtheorem{corollary}{Corollary}
\newtheorem{lemma}{Lemma}

\usepackage{microtype}
\usepackage{graphicx}
\usepackage{subcaption}
\usepackage[export]{adjustbox}
\usepackage{booktabs} % for professional tables
\usepackage{color}
\usepackage{multirow}

\usepackage{algorithmic}
\usepackage{algorithm}
\usepackage{amsmath}
\usepackage{amssymb}
\usepackage{comment}
\usepackage{cases}
\usepackage{enumerate}

\def\E{\mathbb{E}}

\DeclareMathOperator*{\argmin}{arg\,min}
\DeclareMathOperator*{\argmax}{arg\,max}
\def\tr{{\bf tr}}
\newcommand\independent{\protect\mathpalette{\protect\independenT}{\perp}}
\def\independenT#1#2{\mathrel{\rlap{$#1#2$}\mkern2mu{#1#2}}}

\newcommand{\doop}{\textrm{do}}

\def\grds{GRDS}
\def\sras{SRAS}

\begin{document}

\title{Reinforcement Learning with Exogenous States and Rewards}

\author{George Trimponias\thanks{Equal contributions}\\ 
       Intercom\\
       Dublin, Ireland\\ 
       \texttt{george.trimponias@intercom.io}\\
       \And
       Thomas G. Dietterich$^*$\\
       Collaborative Robotics and Intelligent Systems (CoRIS) Institute\\
       Oregon State University, Corvallis, OR 97331 USA\\
       \texttt{tgd@cs.orst.edu}
     	}

\renewcommand{\shorttitle}{Reinforcement Learning with Exogenous States and Rewards}

\maketitle

\begin{abstract}
Exogenous state variables and rewards can slow reinforcement learning by injecting uncontrolled variation into the reward signal. This paper formalizes exogenous state variables and rewards and shows that if the reward function decomposes additively into endogenous and exogenous components, the MDP can be decomposed into an exogenous Markov Reward Process (based on the exogenous reward) and an endogenous Markov Decision Process (optimizing the endogenous reward). Any optimal policy for the endogenous MDP is also an optimal policy for the original MDP, but because the endogenous reward typically has reduced variance, the endogenous MDP is easier to solve. We study settings where the decomposition of the state space into exogenous and endogenous state spaces is not given but must be discovered. The paper introduces and proves correctness of algorithms for discovering the exogenous and endogenous subspaces of the state space when they are mixed through linear combination. These algorithms can be applied during reinforcement learning to discover the exogenous subspace, remove the exogenous reward, and focus reinforcement learning on the endogenous MDP. Experiments on a variety of challenging synthetic MDPs show that these methods, applied online, discover large exogenous state spaces and produce substantial speedups in reinforcement learning.

\end{abstract}

\section{Introduction} \label{sec:introduction}

In many practical settings, the actions of an agent control only a limited part of the environment. For example, in a wireless cellular network, although the cell tower base stations have many parameters that can be adjusted, a control policy for managing the network cannot modify the behavior of the cellular network customers or the radio propagation properties of the atmosphere. Nonetheless, both the reward function and the policy usually depend on these factors. In cellular networks specifically, a typical reward is the negative of the number of customers who are suffering low bandwidth. 

Now consider applying reinforcement learning to optimize a control policy for cell tower management. We can formulate a Markov Decision Process where the state includes both \textit{endogenous variables} (e.g., the cell tower parameters) and \textit{exogenous variables} (e.g., the number and spatial distribution of customers). The exogenous variables can be highly stochastic because of traffic accidents, sporting events, storms, and so on.  This high degree of stochasticity can confuse the reinforcement learning algorithm.  During exploration, for example, the benefit of performing action $a$ in state $s$ is hard to determine---a policy change that increases expected reward may appear bad in a single trial because exogenous factors cause a drop in the reward.  To obtain an accurate estimate of the expected reward, we can average over many trials. This can be implemented in temporal difference algorithms, such as Q-learning, by reducing the learning rate, and it can be implemented in policy gradient methods by increasing the number of Monte Carlo trials when estimating the gradient (or, equivalently, making the gradient update step size very small). This makes reinforcement learning very slow. 

In this paper, we pursue a different approach. Assume for a moment that we know which variables are exogenous and which are endogenous. If $S$ denotes the vector of all state variables, we can decompose it into $S=(E,X)$, where $E$ is the vector of endogenous variables and $X$ is the vector of the exogenous variables. Our approach starts by collecting a training set of experience tuples of the form $\langle s_t,a_t,r_t,s_{t+1}\rangle$ following some exploration policy in the MDP. Then we train a regression model to predict the immediate reward at each step as a function of only the exogenous variables: $r_t \approx f(x_t)$. Of course if the agent's actions have any effect on the reward, then $f$ will only be able to predict part of the reward. We call this part the \textit{exogenous reward} and rename $f(x_t)$ to be $\hat{R}_{exo}(x_t)$, where the hat indicates that this is an estimate of the true exogenous reward. The prediction residual, $r_t - \hat{R}_{exo}(x_t)$, is the part of the reward that is \textit{not} explained by the exogenous variables. We refer to it as the \textit{endogenous reward}, $\hat{R}_{end}(e_t, x_t)$. We can now define a modified MDP where the original reward function is replaced by $\hat{R}_{end}$. Assuming the estimated $\hat{R}_{end}$ equals the true endogenous reward, $R_{end}$, we prove that any optimal policy for this modified MDP is an optimal policy for the original MDP. Experimentally, we show that in the modified MDP with the estimated endogenous reward, $\hat{R}_{end}$, RL learns much faster and, hence, achieves higher expected rewards for the same amount of training, than RL applied to the original MDP.   

In real-world problems, the state variables that can be measured in practice may be mixtures of underlying endogenous and exogenous variables. To apply our method, we must first execute a form of causal discovery to find a representation in which the exogenous and endogenous subspaces are separated. Then we can perform the reward regression to define $\hat{R}_{exo}$. 

To understand the nature of this causal discovery problem, we analyze the structure of exogenous and endogenous subspaces of the state space for general MDPs. We prove that there is a unique maximal exogenous subspace, but that not all subspaces of this maximal space are valid exogenous subspaces. 

We then formalize the discovery problem as a constrained optimization problem where the objective is to maximize the dimensionality of the exogenous subspace subject to conditional mutual information (CMI) constraints that enforce exogeneity. We introduce two algorithms for solving this optimization problem. One algorithm, \grds{}, starts by hypothesizing that all dimensions of the state space are exogenous and iteratively shrinks the number of dimensions until the CMI constraints are satisfied. The other algorithm, \sras{}, starts by hypothesizing that none of the dimensions are exogenous and progressively adds one dimension at a time while checking that the CMI constraints are satisfied. \grds{} provably finds the maximal exogenous subspace, but it must solve a series of high-dimensional optimization problems. \sras{} solves a series of one-dimensional optimization problems, so it has the potential to be faster. However, while it is sound, it may not find the maximal exogenous subspace. This is due to several factors including our earlier finding that not all subspaces of the maximal exogenous space satisfy the CMI constraints. Fortunately, even a non-maximal exogenous subspace may be able to learn an $\hat{R}_{exo}$ that removes a meaningful fraction of the exogenous noise from the reward function and thereby accelerates reinforcement learning.

A strength of our work is that, unlike previous works, we do not assume that the MDP dynamics factor into independently-evolving exogenous and endogenous subspaces---that is, we do not assume $P(E',X'|E,X,A) = P(E'|E,A)P(X'|X)$, where $A=\pi(E)$ is the action chosen by the endogenous policy that depends only on $E$, and $E'$ and $X'$ are the endogenous and exogenous components of the resulting state. In many problems, such as our cell tower management problem, the optimal policy depends critically on the exogenous factors, and hence, the dynamics cannot be factored in this way. Another strength is that we do not assume the exogenous/endogenous decomposition $S=(E,X)$ is given. A limitation is that our algorithms make two assumptions. First, they assume that the exogenous subspace is a linear projection of the original state space. Second, they assume that the conditional mutual information can be approximated by a quantity we call the Conditional Correlation Coefficient (CCC). Our experiments show that even when these assumptions are violated, our methods still yield very substantial speedups in reinforcement learning. 

This paper is organized as follows. We begin in Section 2 with a review of previous research on exogenous state variables in MDPs. In Section 3, we derive structural results on MDPs that have exogenous state variables. We define exogenous state variables based on a causal foundation and then introduce 2-Exogenous State MDPs, which capture a broad class of MDPs with exogenous states. We characterize the space of all 2-Exogenous State MDPs in terms of constraints on the structure of the corresponding two-time step dynamic Bayesian network. The paper analyzes the properties of exogenous subspaces of the state space and proves that every 2-Exogenous State MDP has a unique maximal exogenous subspace that contains all other exogenous subspaces. The paper then shows that, under the assumption that the reward function decomposes additively into exogenous and endogenous components, the Bellman optimality equation for the original MDP decomposes into two equations: one for an exogenous Markov reward process (Exo-MRP) and the other for an endogenous MDP (Endo-MDP). Importantly, every optimal policy for the Endo-MDP is an optimal policy for the full MDP. 

In Section 4, the paper formulates the problem of discovering the exogenous/endogenous decomposition as a constrained optimization problem where the objective is to find the exogenous subspace subject to conditional mutual information constraints that enforce the 2-time step Exogenous State MDP structure.  We also introduce an approximation to the conditional mutual information called the conditional correlation coefficient (CCC).

Section 5 presents our two algorithms, \grds{} and \sras{}, for solving the subspace discovery problem and studies their soundness and optimality. 

Finally, Section 6 studies the performance of our algorithms experimentally on a variety of challenging MDPs. The main finding is that the decomposition algorithms can discover large exogenous subspaces that yield substantial speedups in reinforcement learning. We also explore the behavior of the algorithms on MDPs with nonlinear rewards, nonlinear dynamics, and discrete states. 

The main contributions of this paper are
\begin{itemize}
    \item Definitions and structural results for MDPs with exogenous state variables,
    \item Two practical and provably sound algorithms for discovering the decomposition of the state space into exogenous and endogenous subspaces under the assumption that these subspaces are mixed linearly, and
    \item Experimental demonstration that these algorithms work well on a variety of difficult MDPs including MDPs with non-linear dynamics and non-linear rewards, MDPs with combinatorial action spaces, and MDPs with discrete states and actions.
\end{itemize}

\section{Prior Work}
All prior work has focused on special cases of the general problem of exogenous variables in MDPs. \cite{efroni2022provably} introduce the Exogenous Block Markov Decision Process (EX-BMDP) setting to model environments with exogenous noise. Under the assumption that the endogenous state transition dynamics are nearly deterministic, they propose the Path Predictive Elimination (PPE) algorithm. PPE learns a form of multi-step inverse dynamics \citep{paster2021}. It can recover the latent endogenous model and runs in a reward-free setting without making any assumptions about the reward function.
In subsequent work, \cite{Lamb2023} introduce the Agent Control-Endogenous State Discovery algorithm (AC-State) for the EX-BMDP setting, which is guaranteed to discover the minimal control-endogenous latent state that contains all of the information necessary for controlling the agent. AC-State is also based on multi-step inverse dynamics and employs an information bottleneck to limit the size of the control-endogenous latent state. Multi-step inverse models for offline RL were studied in \citep{islam23a}.

In a different work, \cite{efroni2022sample-efficient} introduce the ExoMDP setting, which is a finite state-action variant of the EX-BMPD where the state directly decomposes into a set of endogenous and exogenous factors, while the reward only depends on the endogenous state. Similar models have been outlined in \citep{dietterich2018, mao2018variance,powell2022reinforcement}. They propose ExoRL, an algorithm that learns a near-optimal policy for the ExoMDP with sample complexity polynomial in the number of endogenous state variables and logarithmic in the number of exogenous components. Because the reward does not depend on the exogenous state variables, those variables are not only exogenous but also irrelevant, and they can be ignored during reinforcement learning. In our work, in contrast, the exogenous variables can still be important for the policy, so the RL algorithm must consider them as inputs when learning the policy. We note that equivalence results between ExoMDPs and discrete MDPs as well as near-optimal algorithms have been recently investigated in \citep{wan2024exploiting}. Furthermore, \cite{sinclair23a} develop efficient hindsight learning algorithms for the ExoMDP based on the insight that, upon getting new samples of exogenous state variables, past decisions can be revisited in hindsight to infer counterfactual consequences that can accelerate policy improvements.

\cite{chitnis2020} address the problem of learning a compact model of an MDP with endogenous and exogenous components for the purpose of planning. They consider only reducing the exogenous part of the state, assuming it induces the most noise, and they assume that the reward function decomposes as a sum over the individual effects of each exogenous state variable. They introduce an algorithm based on the mutual information among the exogenous state variables that generates a compact representation, and they provide conditions for the optimality of their method. Our assumption of additive reward decomposition is weaker than their per-state-variable assumption.
% Although I suppose if we are using linear exo reward regression, we are assuming each exo state variable makes an independent contribution to R_exo

Another line of work related to exogenous information concerns curiosity-driven exploration by \cite{pathak2017}, where the goal is to learn a reward to enable the agent to explore its environment better in the presence of very sparse extrinsic rewards. It falls under the well-studied class of methods with intrinsic rewards \citep{bellemare2016,haber2018,houthooft2016,oh2015,ostrovski2017,badia2020}. This work employs a self-supervised inverse dynamics model, which encodes the states into representations that are trained to predict the action. As a result, the learned representations do not include environmental features that cannot influence or are not influenced by the agent's actions. Based on these representations, the agent can learn an exploration strategy that removes the impact of the uncontrollable aspects of the environment. Note that the inverse dynamics objective is generally not sufficient for control and does not typically come with theoretical guarantees \citep{rakelly2021}. Finally, we mention that reinforcement learning with exogenous information has also been studied from a more empirical standpoint in real problems such as learning to drive \citep{chen2021}.

Our work shares similarities with other work on modeling the controllable aspects in the environment \citep[e.g.,][]{choi2019,song2020,burda2018,bellemare2012,corcoll2022,thomas2018,thomas2017}. The main difference is that in 2-Exogenous state MDPs, we capture the controllable versus uncontrollable aspects via the exo/endo factorization, where endo (respectively, exo) states correspond to the controllable (resp., uncontrollable) aspects. 
%Combining the two families of approaches could be an avenue for future research. 
We also note the work by \cite{Yang2022}, which proposes the dichotomy of control for return-conditioned supervised learning. This is accomplished by conditioning the policy on a latent variable representation of the future and introducing two conditional mutual information constraints that remove any information from the latent variable that has to do with randomness in the environment. However, unlike ours, their work is not concerned with exogenous state variables and decompositions.

Finally, we discuss the connections of our work to representation learning. In the representation learning literature for reinforcement learning, most works can be categorized into three classes based on the mutual information objective that they maximize \citep{rakelly2021}: (i) approaches based on forward dynamics maximize $I(Z_{t+1};Z_t,A_t)$, where $Z$ is the learned representation; (ii) approaches based on inverse dynamics maximize $I(A_t;Z_{t+k}\mid Z_t)$, with $k\geq 1$; and (iii) approaches based on state-only transition dynamics maximize $I(Z_{t+k};Z_t)$, with $k\geq 1$. Interestingly, only the forward dynamics is generally sufficient for control. Contrary to such approaches, our approach minimizes a conditional mutual information metric that captures the distributional properties of the exogenous state representations. Note, however, that our starting point is also the forward transition dynamics model, where we exploit the specific factorization of the endo and exo state dynamics under exogenous information. Our work shows that removing the learned exo representations does not change the optimal policy, so our approach comes with theoretical guarantees. Notice that curiosity-driven exploration by \cite{pathak2017} tries to remove the exogenous components by making use of an inverse dynamics model, which nevertheless does not enjoy theoretical guarantees. 
Finally, we emphasize that in the presence of exogenous information, the state-only transition dynamics that excludes the action from the objective can be particularly problematic, since it will be inclined to include all exogenous components in the representation while potentially ignoring endogenous ones.

\section{Definitions and Structural Properties of Exogenous-State MDPs}

\begin{table}[t!]
\centering
\begin{tabular}{|c|c|} 
 \hline
 \textbf{Symbol} & \textbf{Meaning}\\  
 \hline
 endo, exo & endogenous, exogenous\\  
 \hline
  CMI & Conditional Mutual Information\\
 \hline
  CCC & Conditional Correlation Coefficient\\
 \hline
 $\mathcal{S},\mathcal{E},\mathcal{X},\mathcal{A}$ & State, Endo State, Exo State, and Action Spaces\\
 \hline
 $S/S'$ & Random Vector for Current/Next State\\
 \hline
 $s,a,r$ & Realizations of the State, Action, and Reward\\
  \hline
  $\mathcal{S}_i,S_i,s_i$ & The $i^{th}$ component of $\mathcal{S},S,s$\\
  \hline
  $\mathbf{S,S',A,R}$ & Observational data for state, next state, action, reward\\
  \hline
 $\mathcal{P(B)}$ & Space of probability distributions over space $\mathcal{B}$\\
  \hline
 $S=(E,X)$ & Decomposition of $\mathcal{S}$ into endo and exo sets $E$ and $X$\\
 \hline
  $\subseteq, \subset$ & Subset of, Strict (proper) subset of\\
 \hline
 $[d]$ & The set $\{1,\dots, d\}$\\
 \hline
 $\mathcal{I},\mathcal{I}^c$ & Index set ($\subseteq[d]$), Complement of $\mathcal{I}$ ($=[d] \setminus \mathcal{I}$)\\
 \hline
 $\mathcal{S}[\mathcal{I}],S[\mathcal{I}],s[\mathcal{I}]$ & $(\mathcal{S}_i)_{i\in\mathcal{I}},(S_i)_{i\in\mathcal{I}},(s_i)_{i\in\mathcal{I}}$\\
 \hline
 $I(A;B\mid C)$ & CMI of random vectors $A$ and $B$ given $C$\\
 \hline
 $A\independent B\mid C$ & Conditional independence of $A$ and $B$ given $C$\\
\hline
 $\mathbb{R},\mathbb{R}^d$ & Set of real numbers, Real $d$-dimensional vector space\\
\hline
 $\mathbb{R}^{m\times n}$ & The vector space of $m\times n$ matrices over $\mathbb{R}$\\
  \hline
 $S\stackrel{W}{=}[E,X]$ & Linear decomposition of $S$ into endo/exo parts $E$ and $X$ via $W$ \\ % check whether we actually use this notation in the final paper
\hline
 $\mathcal{A}+\mathcal{B},\mathcal{A}\oplus\mathcal{B}$ & Sum (Direct Sum) of vector spaces $\mathcal{A}$ and $\mathcal{B}$\\
 \hline
 % Tom: I'm pretty sure we don't need this
$\mathcal{A}\sqsubseteq\mathcal{B},\mathcal{A}\sqsubset\mathcal{B}$ & $\mathcal{A}$ is a vector subspace (proper subspace) of vector space $\mathcal{B}$\\
\hline
 $\mathcal{A}^{\perp}$ & Orthogonal complement of subspace $\mathcal{A}$ of vector space $\mathcal{S}$ \\
\hline
 $\dim(\mathcal{A})$ & Dimension of vector subspace $\mathcal{A}$\\
\hline
 $\mathbb{I}_{n},\mathbf{0}^{m\times n}$ & Identity matrix of size $n$, Matrix of zeros of size $m\times n$\\
\hline
 $W_{exo}$ & Matrix that defines the linear exogenous subspace\\
\hline
 $\tr(A),A^{\top},\det(A)$ & Trace of matrix $A$, Transpose of $A$, Determinant of $A$ \\
\hline
 $\|u\|_2$ &Euclidean norm of vector $u$ \\
\hline % check that we use the Frobenius norm in the final paper
 $\Sigma_{AA}, \Sigma_{AB}$ & Covariance matrix of $A$, Cross-covariance matrix of $A,B$ \\
\hline
$\mathcal{N}(\mu,\sigma^2)$ & Gaussian distribution with mean $\mu$ and variance $\sigma^2$ \\
\hline
\end{tabular}
\caption{Symbols and Abbreviations.}
\label{table:symbols}
\end{table}

In this section, we provide a causal definition of exogenous variables in MDPs and then identify conditions on the structure of the MDP causal graph (and the corresponding probabilistic graphical model) that are necessary and sufficient for state variables to be exogenous. This leads to a definition of valid exogenous-endogenous decompositions of MDPs about which we then prove properties that are important for designing algorithms for finding decompositions. Finally, we show that if the reward function can be decomposed additively into exogenous and endogenous components, then a reduced MDP---the endogenous MDP---can be defined for which any optimal policy is an optimal policy for the original MDP. 

We study discrete time stationary MDPs with stochastic rewards and stochastic transitions \citep{Puterman1994,Sutton1998}; the state and action spaces may be either discrete or continuous. For tractability, our analysis is restricted to episodic MDPs with fixed horizon $H$. (Our experiments do not make this assumption.)  Notation: state space $\cal{S}$, action space $\cal{A}$, reward distribution $R\!:\cal{S}\times \mathcal{A}\mapsto \mathcal{P}(\mathbb{R})$ (where $\mathcal{P}(\mathbb{R})$ is the space of probability distributions over the real numbers), transition function $P\!: \cal{S} \times \cal{A} \mapsto \mathcal{P(S)}$ (where $\mathcal{P(S)}$ is the space of probability distributions over $\cal{S}$), starting state distribution $P_0\! \in \mathcal{P(S)}$, fixed horizon $H$, and discount factor $\gamma \in (0,1]$. We assume that for all $(s,a)\in {\cal S}\times {\cal A}$, $R(s,a)$ has expected value $m(s,a)$ and finite variance $\sigma^2(s,a)$.  We denote random variables by capital letters ($S$, $A$, etc.)\ and their corresponding values by lower case letters ($s$, $a$, etc.). Table~\ref{table:symbols} summarizes the notation employed in this paper.

Let the state space $\mathcal{S}$ take the form $\mathcal{S}=\times_{i=1}^d\mathcal{S}_i$, where $\mathcal{S}_i$ defines the domain of the $i^{th}$ state variable. In our problems, the domain of each variable is either the real numbers or a finite, discrete set of values.
Each state $s \in \mathcal{S}$ can then be written as a $d$-tuple of the values of these state variables $s=(s_1,\dots,s_d)$, with $s_i\in\mathcal{S}_i$. 
We refer to $s_i$ as the value of the $i^{th}$ state variable. 
We denote by $S_t=\times_{i=i}^d S_{t,i}$ the random vector for the state at time $t$. Similarly, $A_t$ is the random variable for the action at time $t$, and $R_t$ is the random variable for the reward at time $t$. In some formulas, instead of indexing by time, we will use ``prime'' notation. For example, $S$ and $S'$ denote the current and next states, respectively (and analogously for $A$ and $A'$, $R$ and $R'$).  When it is clear from the context, we will also refer to $S_i$ as the $i$-th state variable. In the terminology of \cite{Koller2009}, $S_i$ is a ``template variable'' that refers to the family of random variables that correspond to $\mathcal{S}_i$ at all time steps: $\{S_{1,i}, S_{2,i},\ldots,S_{H,i}\}$. 

We are interested in problems where the set of state variables $S$ can be decomposed into endogenous and exogenous sets $E$ and $X$. In the simplest case, this can be accomplished by variable selection. Following the notation of \cite{efroni2022sample-efficient}, define an index set $\mathcal{I}$ as a subset of $[d]=\{1,\dots,d\}$ and $\mathcal{I}^c=[d]\setminus \mathcal{I}$ as its complement. The variable selection formulation aims to discover an index set $\mathcal{I}$ so that the state vector $S=\times_{i=1}^dS_i$ can be decomposed into two disjoint sets of state variables $X=\times_{i\in{\mathcal{I}}}S_i=S[\mathcal{I}]$ and $E=\times_{i\in{\mathcal{I}}^c}S_i=S[\mathcal{I}^c]$.  We will also denote the corresponding exogenous and endogenous state spaces as $\mathcal{X}=\times_{i\in\mathcal{I}}\mathcal{S}_i=\mathcal{S}[\mathcal{I}]$ and $\mathcal{E}=\times_{i\in\mathcal{I}^c}\mathcal{S}_i=\mathcal{S}[\mathcal{I}^c]$.

In many problems, the given state variables do not neatly separate into endogenous and exogenous subsets. Instead, we must discover a mapping $\xi: U \mapsto V$ such that the first $d_{exo}$ dimensions of $\xi(\mathcal{S})$ provide the exogenous state variables, and the remaining $d-d_{exo}$ dimensions give the endogenous state variables. In this paper, we study the case where $\xi$ is a full-rank linear transformation. 

%In general, we will argue that $\xi$ should a diffeomorphism so that no information in the original space ${\cal S}$ is lost in the new space $\xi(\mathcal{S})$. 

\subsection{Causally-Exogenous State Variables}
\label{sec:exogenous-state-variables}

The notion of exogeneity is fundamentally causal: a variable is exogenous if it is impossible for our actions to affect its value. We formalize this in terms of Pearl's do-calculus \citep{pearl2009}. 

\begin{definition}[Causally-Exogenous Variables]\label{def:exogenous}
A set of state variables $X = S[\mathcal{I}]$ is \emph{causally exogenous} for MDP $\mathcal{M}$ with causal graph $\mathcal{G}$ if and only if for all times $t<H$, graph $\mathcal{G}$ encodes the conditional independence
\begin{equation}\label{eq:causal-exogeneity}
P(X_{t+1}, \ldots, X_H\mid X_t, \textnormal{do}(A_t=a_t)) = P(X_{t+1}, \ldots, X_H\mid X_t) \quad\forall a_t \in \mathcal{A}.
\end{equation}
\end{definition}

Causal exogeneity is a qualitative property that depends only on the structure of the causal graph $\mathcal{G}$ and not on the parameters of the probability distributions associated with each node. If there is a parameterization of the probability distributions that violates \eqref{eq:causal-exogeneity}, then the state variables $X$ are not causally exogenous.

% \begin{figure}[t!]
%     \centering
%         \includegraphics[scale=0.8]{fully-connected-diagram-cropped.pdf}
%         \caption{A fully-connected dynamic Bayesian network for state decomposition $S=(E,X)$ (reward function not shown).}
%         \label{fig:fully-connected}
%\end{figure}

We now introduce a condition on the structure of the causal graph $\mathcal{G}$ that is necessary and sufficient to ensure that the variables in $X$ are causally exogenous. We begin by considering the case where $X$ consists of a single state variable $S = \mathcal{S}[\{i\}]$.

\begin{definition}[Action-Disconnected State Variable]\label{def:structural}
A state variable $S = \mathcal{S}[\{i\}]$ is \emph{action-disconnected} if the causal graph $\mathcal{G}$ for the MDP contains no directed path of the form $A_t\to\cdots\to S_{\tau}, \forall t\in\{0,\dots,H-1\}, \forall \tau\in\{t+1,\dots,H\}$.
\end{definition}

\begin{theorem}\label{theorem:generic-causal-exogeneity}
    A state variable $S$ is {\em causally exogenous} if and only if it is action-disconnected.    
\end{theorem}

We defer the proof to Appendix\ref{app:causal-exogeneity-beyond-DBNs}.

This result can be generalized to establish a structural criterion for causal exogeneity in MDPs:
\begin{corollary}\label{cor:X-causal-exogeneity}
    A set $X$ of state variables is causally exogenous in MDP $\mathcal{M}$ if and only if each variable $S \in X$ is action-disconnected in the causal graph of $\mathcal{M}$. 
\end{corollary}

\begin{figure}[b!]
    \centering
    \includegraphics[scale=0.8]{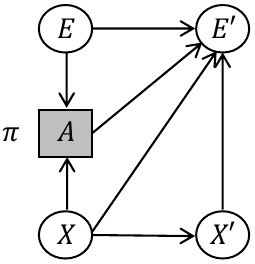}
    \caption{Restricted 2-Time Step Dynamic Bayesian Network sufficient to establish that $X$ is exogenous.}
    % We call this the ``full setting'' because it includes the synchronic edge from $X'$ to $E'$.
    \label{fig:full-setting}
\end{figure}

This structural criterion requires examining the entire causal graph $\mathcal{G}$. A convenient special case is to consider only stationary MDPs. These can be represented by 2-Time Step Dynamic Bayesian Networks (DBNs, \cite{Koller2009}). We now state conditions on the structure of a 2-Time Step DBN that are sufficient to ensure that a set $X$ of state variables is causally exogenous. The intuition is that if the causal graph matches Figure~\ref{fig:full-setting} (or any subset of it), then there is no directed path from the action $A$ and endogenous variables $E$ to any future exogenous variables $X'$. 

\begin{theorem}[Causally Exogenous DBN]
\label{theorem:exo-DBN}
Any DBN with a causal graph matching the structure of Figure~\ref{fig:full-setting} or matching any DBN graph obtained by deleting edges from this structure, when unrolled for $H$ time steps, yields an $H$-horizon MDP for which $X$ is causally exogenous. 
\end{theorem}

\begin{figure}
     \centering
     \includegraphics[scale=0.8]{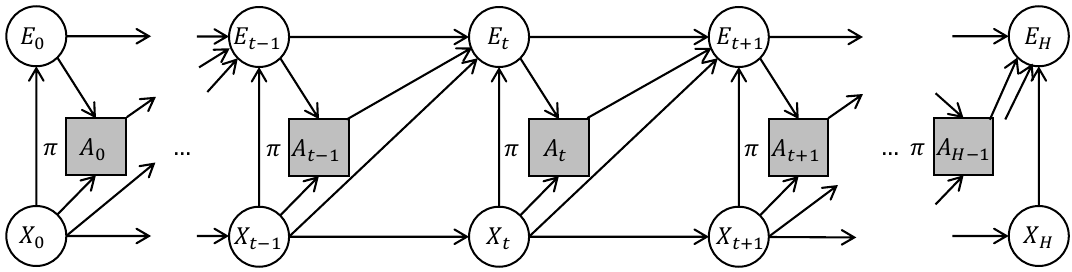}
     \caption{Unrolled state transition diagram for the full exo DBN.}
     \label{fig:unrolled-setting}
\end{figure}

\begin{proof}
Let $\mathcal{G}$ be the full causal graph obtained by unrolling the DBN causal graph over the $H$-step horizon, as in Figure~\ref{fig:unrolled-setting}. Note that all edges from action nodes and $X$ nodes point directly or indirectly to $E$ nodes. Hence, every state variable in $X$ is action-disconnected.  
\end{proof}
Note that $E$ may contain additional action-disconnected variables. Hence, this structural condition is only sufficient but not necessary.

\begin{figure}[b!]
     \centering
         \includegraphics[scale=0.8]{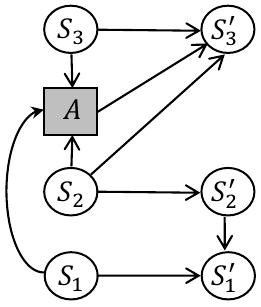}
         \caption{State transition diagram of MDP with 3 state variables.}
         \label{fig:example_MDP1}
\end{figure}

\subsection{Probabilistically-Exogenous State Variables}
When applying reinforcement learning to an unknown MDP, we do not know the causal graph. Instead, we can estimate the probability distribution for the state transition dynamics $P(S' \mid S, A)$.  
Inspired by this observation, we define the following probabilistic analog of the 2-Time Step Causally Exogenous MDP: 

\begin{definition}[2-Exogenous State MDP]\label{def:2-exogeneous-state-MDP}
Any MDP whose conditional distribution\\ $P(E',X' \mid E, X, A)$ can be factored to match the DBN structure of Figure~\ref{fig:full-setting} is called a \emph{2-Time Step Exogenous State MDP with exogenous set $X$}. We will shorten this to \emph{2-Exogenous State MDP} throughout the paper. 
\end{definition}

\begin{definition}[Valid Exo/Endo Decomposition]\label{def:valid-exo/endo-decomposition}
    A state decomposition $S = (E,X)$ is said to be a (probabilistically) \emph{valid exo/endo decomposition} if it permits the conditional distribution of the MDP to be factored as 
    \begin{equation} \label{eqn:full-factorization}
        P(E',X'\mid E,X,A) = P(E' \mid E, X, A, X') P(X'\mid X).
    \end{equation}
\end{definition}

Note that this definition does not require that the variables in $X$ are \textit{causally} exogenous. For purposes of accelerating reinforcement learning, it is sufficient to obtain a probabilistically valid endo/exo decomposition without proving that it corresponds to a correct causal model.

We now examine some properties of exo/endo decompositions that will be important for designing algorithms for decomposing MDPs into exogenous and endogenous components. These properties are all stated and proved with respect to the DBN structure viewed as a factorization of the conditional probability distribution describing the MDP dynamics. However, these properties also apply to causal graphs that satisfy the same DBN structure. 

We first note that it is possible for an MDP with exogenous state variables to accept multiple valid exo/endo decompositions. In Figure \ref{fig:example_MDP1}, the decompositions $(X_1=\{S_2\},E_1=\{S_1,S_3\})$ and $(X_2=\{S_1,S_2\},E_3=\{S_3\})$  are both valid decompositions, since they match the full DBN template of Figure \ref{fig:full-setting}. This shows that the set $E$ in an exo/endo decomposition $(E,X)$ may contain additional exogenous state variables not in $X$. 

\begin{theorem}[Union of Exo/Endo Decompositions]\label{theorem:decomposition-union}
Assume that a 2-Exogenous State MDP accepts two valid exo/endo decompositions $(E_1,X_1)$ and $(E_2,X_2)$. Define the union of the two decompositions as the state decomposition $(E,X)$ with $X=X_1\cup X_2$ and $E=E_1\cap E_2$. Then $(E,X)$ is a valid exo/endo decomposition with exo state set $X$.
\end{theorem}
\begin{proof}
Because $(E_1,X_1)$ is a valid decomposition, there are no edges from any of the state variables in $E_1$ and no edges from the action $A$ to any variable in $X_1$. Similarly, there are no edges from nodes in $E_2$ or $A$ to nodes in $X_2$. Hence, the union $X=X_1\cup X_2$ also has no incoming edges from $E_1$, $E_2$, or $A$, and therefore it has no incoming edges from $E$. This shows that the decomposition $(E,X)$ satisfies the conditions of Definitions \ref{def:2-exogeneous-state-MDP} and \ref{def:valid-exo/endo-decomposition}. 
\end{proof}

On the other hand, not every subset of exogenous state variables can yield a valid exo/endo decomposition. 

\begin{theorem}[Exogenous Subsets] \label{theorem:bad-subsets}
Let $(X,E)$ be a valid exo/endo decomposition and $X_1$ and $X_2$ be non-empty, disjoint proper subsets of $X$, $X_1 \cap X_2 = \emptyset$, such that their disjoint union gives back $X$: $X_1 \cup X_2 = X$. Then $(X_1, X_2 \cup E)$ is not necessarily a valid exo/endo decomposition.
\end{theorem}
\begin{proof} By example. In Figure~\ref{fig:example_MDP1}, the decomposition $X_3=\{S_1\},E_3=\{S_2,S_3\}$ is not a valid exo/endo decomposition due to the edge from $E'_3$ to $X'_3$ (specifically the edge $S'_2\to S'_1$). %This is true even though $X_3$ is a proper subset of the maximal exogenous set $\{S_1,S_2\}$. 
\end{proof}

For 2-Exogenous state MDPs, we will be interested in the exo/endo decomposition where the exo set $X$ is as large as possible. This is formalized in the next definition:

\begin{definition}[Maximal Exo/Endo Decomposition]
Given a 2-Exogenous state MDP, a valid exo/endo decomposition $(E,X)$ is maximal if there is no other valid exo/endo decomposition $(\tilde{E},\tilde{X})$ with $|\tilde{X}|>|X|$. We denote the maximal decomposition by $(E_m,X_m)$ and call $X_m$ the \emph{maximal exo set}.
\end{definition}

\begin{corollary}[Uniqueness of Maximal Exo/Endo Decomposition]\label{cor:decomposition-uniqueness}
The maximal \\exo/endo decomposition of any MDP is unique.
\end{corollary}
\begin{proof}
By contradiction. Suppose that there are two distinct maximal decompositions $(E_1,X_1)$ and $(E_2,X_2)$. By Theorem~\ref{theorem:decomposition-union}, their union would be a valid exo/endo decomposition with an exo state set of higher cardinality. This contradicts the assumption that $(E_1,X_1)$ and $(E_2,X_2)$ are maximal.
\end{proof}

\begin{corollary}[Containment for Maximal Exo/Endo Decomposition]\label{cor:decomposition-containment}
For the maximal exo/endo decomposition $(E_m,X_m)$, it holds that $X_m\supseteq X$, where $X$ is the exo set of any valid exo/endo decomposition $(E,X)$.
\end{corollary}
\begin{proof}
By contradiction. Suppose there exists a decomposition $(E,X)$ so that $X\not\subseteq X_m$. By Theorem~\ref{theorem:decomposition-union}, we could take the union of $(E,X)$ and $(E_m,X_m)$ to get a new valid exo/endo decomposition with exo set $X\cup X_m\supset X$ of higher cardinality than $X$. 
\end{proof}

\subsection{Additive Reward Decomposition}

In this paper, we identify and analyze a case where reinforcement learning can be accelerated even when the exogenous variables are all relevant to the policy, the dynamics, and the reward. This case arises when the reward function can be decomposed additively into two functions, $R_{exo}$, which only depends on $X$, and $R_{end}$, which can depend on both $X$ and $E$. 

\begin{definition}[Additively Decomposable 2-Exogenous State MDP]\label{def:additive-MDP}
An \emph{Additively Decomposable 2-Exogenous State MDP} is a 2-Exogenous State MDP whose reward function can be decomposed into the sum of two terms
\[R(x,e,a) = R_{exo}(x) + R_{end}(x,e,a),\]
where  $R_{exo}\!:\!{\mathcal X} \mapsto \mathcal{P}(\mathbb{R})$ is the \emph{exogenous reward function} and $R_{end}\!:\!{\mathcal E} \times {\mathcal X} \times {\mathcal A} \mapsto \mathcal{P}(\mathbb{R})$ is the \emph{endogenous reward function}. If the reward function is defined as a distribution over $\mathcal{S}\times\mathcal{A}\times\mathcal{S}$, we instead consider the decomposition $R(x,e,a,x',e') = R_{exo}(x,x') + R_{end}(x,e,a,x',e')$.
\end{definition}
Let $m_{exo}(x)$ and $\sigma^2_{exo}(x) < \infty$ be the mean and variance of the exogenous reward distribution in state $x$. Similarly, let $m_{end}(e,x,a)$ and $\sigma^2_{end}(e,x,a) < \infty$ be the mean and variance of the endogenous reward distribution for state-action pair $(e,x,a)$. 

\begin{theorem}\label{theorem:MDP-decomposition} For any Additively Decomposable 2-Exogenous State MDP with valid exo/endo decomposition $S=(E,X)$, the $H$-step finite-horizon Bellman optimality equation (Equation \ref{eqn:bellman}) for the value function $V(e,x;h)$ can be decomposed into two separate equations, one for a Markov Reward Process involving only $X$ and $R_{exo}$ (Equation \ref{eqn:exo}) and the other for an MDP (the endo-MDP) involving $X$, $E$, and $R_{end}$ (Equation \ref{eqn:end}). The sum of $V_{exo}$ and $V_{end}$ obtains the original optimal value function $V$ (Equation \ref{eqn:sum}):
\begin{align}
    V(e,x;h) &= \max_{a} \mathbb{E}[R(e,x)] + \gamma \E_{x'\sim P(x'\mid x); e'\sim P(e'\mid e,x,x',a)}[V(e',x';h-1)] \label{eqn:bellman}\\
    V_{exo}(x;h) &= m_{exo}(x) + \gamma \E_{x'\sim P(x'\mid x)}[V_{exo}(x';h-1)] \label{eqn:exo}\\
    V_{end}(e,x;h) &= \max_{a} m_{end}(e,x,a) + \E_{x'\sim P(x'\mid x); e'\sim P(e'\mid e,x,x',a)}[V_{end}(e',x';h-1)]. \label{eqn:end}\\
    V(e,x;h) &= V_{exo}(x;h) + V_{end}(e,x;h) \label{eqn:sum}
\end{align}
Here $V(e,x;h)$ is the value (according to the optimal policy) of the state $(e,x)$ with $h$ steps remaining until the horizon $H$ is reached. $V_{exo}(x;h)$ and $V_{endo}(e,x;h)$ are the corresponding value functions for the exogenous Markov Reward Process (MRP) and the optimal policy of the endo-MDP. 
\end{theorem}
\begin{proof}
Proof by induction on the horizon $H$. Note that the expectations could be either sums (if $\mathcal{S}$ is discrete) or integrals (if $\mathcal{S}$ is continuous).

\vspace*{0.1in}
\noindent {\bf Base case:} $H=1$; we take one action and terminate. 
\[V(e,x;1) =  m_{exo}(x) + \max_a m_{end}(x,a).\]
The base case is established by setting $V_{exo}(x;1) = m_{exo}(x)$ and $V_{end}(e,x;1) = \max_a m_{end}(e,x,a)$. 

\vspace*{0.1in}
\noindent {\bf Recursive case:} $H=h$. 
\begin{align*}
    V(e,x;h) = m_{exo}(x) + \max_a \{&m_{end}(e,x,a)\; + \\
    & E_{x'\sim P(x'\mid x); e'\sim P(e'\mid e,x,x',a)}[V_{exo}(x';h-1) + V_{end}(e',x';h-1)]\}.
\end{align*}
Distribute the expectation over the sum in brackets and simplify. We obtain
\begin{align*}
V(e,x;h) =\; &m_{exo}(x) + \gamma \E_{x'\sim P(x'\mid x)}[V_{exo}(x';h-1)]\; +\\
           &\max_a \{m_{end}(e,x,a) + \gamma \E_{x'\sim P(x'\mid x); e'\sim P(e'\mid e,x,x',a)}[V_{end}(e',x';h-1)]\}.
\end{align*}
The result is established by setting
\begin{align*}
    V_{exo}(x;h) &= m_{exo}(x) + \gamma \E_{x'\sim P(x'\mid x)}[V_{exo}(x';h-1)]\\
    V_{end}(e,x;h) &= \max_{a}\{ m_{end}(e,x,a) + \gamma \E_{x'\sim P(x'\mid x); e'\sim P(e'\mid e,x,x',a)}[V_{end}(e',x';h-1)]\}.
\end{align*}
\end{proof}

\begin{corollary}\label{corollary:optimal-policy}
Any optimal policy for the endo-MDP of Equation~\eqref{eqn:end} is an optimal policy for the full 2-Exogenous state MDP.
\end{corollary}
\begin{proof}
Because $V_{exo}(s;H)$ does not depend on the policy, the optimal policy can be computed simply by solving the endo-MDP.
\end{proof}  

Bray (\citeyear{Bray2017}) proves a result similar to Theorem~\ref{theorem:MDP-decomposition}. He also identifies conditions under which value iteration and policy iteration for a fully-specified Endo-MDP can be accelerated by computing the eigenvector decomposition of the endogenous transition matrix. While such techniques are useful for MDP planning with a known transition matrix, we do not know how to exploit them in reinforcement learning where the MDP is unknown.
In other related work, McGregor et al.~(\citeyear{McGregor2017}) show how to remove known exogenous state variables in order to accelerate the Model Free Monte Carlo algorithm \citep{Fonteneau2012}. Their experiments obtain substantial speedups in policy evaluation and reinforcement learning.

\section{Decomposing a 2-Exogenous State MDP: Optimization Formulations}
\label{sec:optimization-formulations}
We now turn our attention to formalizing the task of discovering the exogenous variables (or exogenous subspace) of an MDP as an optimization problem. Then in Section~\ref{sec:algorithms} we will introduce algorithms for solving (approximately) these optimization problems. 

Our overall strategy is shown in Algorithm~\ref{alg:framework}. We assume we are applying an online reinforcement learning algorithm, such as PPO or Q-learning. As the RL algorithm interacts with the environment, it collects $\langle s_t, a_t, r_t, s_{t+1} \rangle$ experience tuples into a dataset $D$. After $L$ tuples have been collected, we apply an exogenous space discovery algorithm (see Section~\ref{sec:algorithms}) to find a function $\xi_{exo}$, so that $x_t = \xi_{exo} (s_t)$ computes the exogenous state $x_t$ from state $s_t$. By applying $\xi_{exo}$ to each $s_t$ of the experience tuples, we assemble a supervised training set $D_{exo}$ of the form $\{(x_t, r_t)\}$. We then solve a regression problem to predict as much of the reward $r_t$ from $x_t$ as possible. The resulting fitted function $\hat{m}_{exo}$ is our estimate of the mean of the exogenous reward function. Because of the additive decomposition, we can therefore estimate the endogenous rewards\footnote{Note that $\hat{r}_{end,t}$ includes any zero-mean random noise in $R_{end}$.} as $\hat{r}_{end,t} := r_t - \hat{m}_{exo}(s_t)$. We then convert the set of experience tuples into modified tuples $\langle s_t, a_t, \hat{r}_{end,t}, s_{t+1}\rangle$ by replacing the original reward $r_t$ values with our estimate of the endogenous reward and resume running the online reinforcement learning algorithm. Depending on the algorithm, we may need to re-initialize the data structures using the modified experience tuples. In any case, as the algorithm collects additional full experience tuples $\langle s_t, a_t, r_t, s_{t+1} \rangle$, each is converted to an experience tuple for the endo-MDP by replacing $r_t$ by $\hat{r}_{end,t}$. In our experiments, we find that there is some benefit to repeating the reward regression at regular intervals, so we also add $(\xi_{exo}(s_i), r_i)$ to $D_{exo}$ at each time step. However, we do not observe similar benefits from rerunning the exogenous space discovery algorithm, especially when the state transition function is linear, so we only execute it once. 

\begin{algorithm}[t!]
\begin{algorithmic}[1]
\STATE {\bf Inputs:} Decomposition steps $L$, Exogenous reward update steps $M$, Policy update steps $K$, Total steps $N$
\STATE {\bf Datasets:} RL training tuples $D$, Exogenous reward examples $D_{exo}$
\STATE {\it //Phase 1}
\STATE Initialize policy and/or value networks randomly
\FOR {$i=1$ \textbf{to} $L$}
\STATE Run RL to collect a new transition $\langle s_i,a_i,r_i,s'_i\rangle$ and add it to $D$
\IF {$i \mod K == 0$}
\STATE Update the policy using the last $K$ observations with observed rewards $r_i$
\ENDIF
\ENDFOR
\STATE Run a state decomposition algorithm (see Section \ref{sec:algorithms}) on $D$ to compute the exogenous state mapping $\xi_{exo}$
\STATE Let $D_{exo} = \{(\xi_{exo}(s_i), r_i)\}_{i=1}^L$ be the exogenous reward training set
\STATE Fit the exogenous reward function $\hat{m}_{exo}$ to $D_{exo}$
\STATE Update $D$ by replacing each tuple $\langle s_i,a_i,r_i,s'_i\rangle$ by $\langle s_i, a_i, r_i - \hat{m}_{exo}(\xi_{exo}(s_i)),s'_i\rangle$
\STATE {\it //Phase 2}
\FOR {$i=L+1$ \textbf{to} $N$}
\STATE Run RL to collect a new transition $\langle s_i,a_i,r_i,s'_i\rangle$ 
\STATE Add $(\xi_{exo}(s_i), r_i)$ to $D_{exo}$
\STATE Add $\langle s_i, a_i, r_i - \hat{m}_{exo}(\xi_{exo}(s_i)), s'_i\rangle$ to $D$
\IF {$i \mod K == 0$}
\STATE Update the policy using last $K$ observations in $D$
\ENDIF
\IF {$i \mod M == 0$}
\STATE Update the estimate of the exogenous reward function $\hat{m}_{exo}$ with the last $M$ observations in $D_{exo}$
\ENDIF
\ENDFOR
\end{algorithmic}
\caption{Practical RL with the Exo/Endo Decomposition}
\label{alg:framework}
\end{algorithm}

The heart of our approach is the exogenous state discovery problem. We will first describe this for the case where we are given a fixed set of state variables and our goal is to determine which state variables are exogenous and which are endogenous. We will then consider the more general case where the given state variables are a linear mixture of the underlying exogenous and endogenous variables and we must unmix them as well. 

\subsection{Variable Selection Formulation}\label{sec:variable-selection-formulation}
The variable selection formulation aims to discover an index set $\mathcal{I}$ with the following two properties:
\begin{itemize}
\item $\mathcal{I}$ decomposes the set of state variables $S=\times_{i=1}^d S_i$  into two disjoint sets $X=\times_{i\in{\mathcal{I}}}S_i$ and $E=\times_{i\in{\mathcal{I}}^c}S_i$ that satisfy the structure of Figure~\ref{fig:full-setting}.
\item The squared error of the exogenous reward regression,
$\sum_t [\hat{m}_{exo}(x_t) - r_t]^2$ is minimized, where $\hat{m}_{exo}$ regresses $r_t$ onto $X=\times_{i\in{\mathcal{I}}}S_i$. 
\end{itemize}
We can express the first property as the following conditional mutual information (CMI) constraint:
\begin{equation}\label{eq:full-mi-constraint}
I(X';[E,A] \mid X) = 0,
\end{equation}
where $X=S[\mathcal{I}]$ and $E=S[\mathcal{I^C}]$. This says that if we know the value of $X$, then the combination of the endogenous state variables and the action carries no additional information about $X'$, the value of the exogenous state in the next time step. 

We can then formulate the exogenous variable selection problem as 
\begin{equation}\label{opt:abstract_formulation_with_reward}
\begin{split}
\mathcal{I}^*,\hat{m}_{exo}^*=&\argmin_{\mathcal{I}\subseteq[d],\hat{m}_{exo}:\mathcal{X} \mapsto \mathbb{R}} \E[(\hat{m}_{exo}(X) - R)^2]\\
&\mbox{subject to } I(X'; [E, A] \mid X)=0,X=S[\mathcal{I}],E=S[\mathcal{I}^c].
\end{split}
\end{equation}
This formulation considers all subsets of the variables (as indexed by $\mathcal{I}$), checks whether they satisfy the CMI constraint, and then chooses the subset that yields the smallest squared error in the reward regression. The optimal set $\mathcal{I}^*$ is not necessarily the maximal subset. This is a difficult optimization problem. 

We instead consider the following two phase formulation:
\begin{align}
\hat{m}_{exo}^*=\argmin_{\hat{m}_{exo}:{\mathcal X}^* \mapsto \mathbb{R}} &\E[(\hat{m}_{exo}(X)-R)^2], \textrm{ where } X^*=S[{\mathcal{I}^{*}}] \mbox{ and}\label{opt:decoupled_abstract_formulation_a}\\
\mathcal{I}^*=&\argmax_{\mathcal{I}\subseteq[d]} |\mathcal{I}|\label{opt:decoupled_abstract_formulation_b}\\
&\mbox{subject to } I(X'; [E, A] \mid X)=0,X=S[\mathcal{I}],E=S[\mathcal{I}^c].\label{opt:decoupled_abstract_formulation_c}
\end{align}
The first phase, subproblem \eqref{opt:decoupled_abstract_formulation_b}-\eqref{opt:decoupled_abstract_formulation_c}, computes the maximal exogenous subspace to produce $\mathcal{I}^*$. The second phase, subproblem \eqref{opt:decoupled_abstract_formulation_a}, then performs the reward regression. A potential weakness of this approach is that $\mathcal{I}^*$ may include exogenous variables that provide no benefit when predicting $R$. Such irrelevant variables can damage the accuracy of a regression estimate. Hence, some form of feature selection or regularization should be applied when solving \eqref{opt:decoupled_abstract_formulation_a}.

The advantage of this two-phase formulation is that we can exploit the structural results from Section~\ref{sec:exogenous-state-variables} to efficiently solve the first phase.

In Algorithm~\ref{alg:framework}, we must estimate the conditional mutual information from the collected set $D$ of observed state transitions and immediate rewards. A question for future research is to identify conditions on the exploration policy and the sample size under which the algorithm finds a good approximation to the optimal policy.  In Appendix~\ref{app:soundness}, we derive conditions that ensure asymptotic correctness of Algorithm~\ref{alg:framework}. 

\subsection{Unmixing Formulation}\label{sec:continuous-formulation}
We turn now to the situation in which the state space does not come already decomposed into state variables that are either exogenous or endogenous. Instead, the state space $\mathcal{S}\subseteq\mathbb{R}^{d}$ is an open subset $U \subseteq \mathbb{R}^{d}$, and each $s \in \mathcal{S}$ is thus a $d$-dimensional real-valued vector. The discovery problem is to find a mapping $\xi: U \mapsto V$ where $V \subseteq \mathbb{R}^d$ is also an open set such that the endogenous and exogenous state spaces can be readily extracted. Without loss of generality, we stipulate that the first $d_{exo}$ components in $\xi(\mathcal{S})$ define the exogenous subspace, while the remaining $d_{end}=d-d_{exo}$ components define the endogenous subspace. Hence, $\mathcal{I} = [d_{exo}]$, so that $\mathcal{X}=\xi(\mathcal{S})[\mathcal{I}]$ and $\mathcal{E}=\xi(\mathcal{S})[\mathcal{I}^c]$. 

With this change of notation, we can re-express the two-phase optimization problem as follows:
\begin{align}
\hat{m}_{exo}^*=&\argmin_{\hat{m}_{exo}:\mathcal{X}^* \mapsto \mathbb{R}} \E[(\hat{m}_{exo}(X^*)-R)^2], \textrm{ where }\mathcal{X}^*=S[{\mathcal{I}^{*}}], \mathcal{I}^*=[d_{exo}^*], \mbox{ and }\label{opt:general_formulation_decoupled_a}\\
& \xi^*,d_{exo}^*=\argmax_{\xi,d_{exo}\in\{0,\dots,d\}} d_{exo} \label{opt:general_formulation_decoupled_b}\\
& \hspace{2cm} \mbox{subject to }I(X'; [E, A] \mid  X) = 0,\\
& \hspace{2cm} \textrm{where } \mathcal{I} = [d_{exo}],\; \mathcal{X}=\xi(\mathcal{S})[\mathcal{I}],\mbox{ and }\mathcal{E}=\xi(\mathcal{S})[\mathcal{I}^c].\label{opt:general_formulation_decoupled_c}
\end{align}
In phase 1, Equations \eqref{opt:general_formulation_decoupled_b}-\eqref{opt:general_formulation_decoupled_c} compute the mapping $\xi^*$ that defines the maximal exogenous subspace $\mathcal{X}^*$ of dimension $d^*_{exo}$. 
In phase 2, \eqref{opt:general_formulation_decoupled_a} performs the reward regression. 

What conditions must the mapping $\xi$ satisfy? To ensure that the resulting decomposed MDP is equivalent to the original MDP and that we can evaluate the CMI constraint, it is convenient that $\xi$ preserves probability densities. A sufficient condition is that $\xi$ be a \textit{diffeomorphism} \citep{diffeomorphisms}. A diffeomorphism from an open subset $U\subseteq\mathbb{R}^d$ to an open subset $V\subseteq\mathbb{R}^d$ is defined as a bijective map $\xi: U\mapsto V$ so that both $\xi$ and its inverse $\xi^{-1}$ are continuously differentiable. We will restrict our attention to the case where $\xi$ belongs to the orthogonal group---the space of linear transformations defined by an orthonormal matrix $W$. These are diffeomorphisms that exactly preserve probability densities because $|\det W|=1$. 

We construct the matrix $W$ as follows. First, we define a linear mapping $\xi_{exo}$ from the full state space to the exogenous state space. We then define the endogenous state space as its orthogonal complement. Let $\xi_{exo}$ be specified by a matrix $W_{exo}\in\mathbb{R}^{d\times d_{exo}}$ with $0\leq d_{exo}\leq d$, where $d_{exo}$ is the dimension of subspace $\mathcal{X}$. We require the columns of $W_{exo}$ to be orthonormal vectors. Given a state $s$, its projected exogenous state is $W_{exo}^{\top}\cdot s$. The endogenous subspace is the orthogonal complement of $\mathcal{X}$ of dimension $d_{end}=d-d_{exo}$, written $\mathcal{E}=\mathcal{X}^{\perp}$ and defined by some matrix $W_{end}$, again with orthonormal column vectors. The endogenous state $e$ contains the components of $s$ in subspace $\mathcal{E}$. In the linear setting, $\mathcal{E}$ and $\mathcal{X}$ are vector subspaces of vector space $\mathcal{S}$, and we can write $\mathcal{S}=\mathcal{E}\oplus\mathcal{X}$, with $\dim(\mathcal{S})=\dim(\mathcal{E})+\dim(\mathcal{X})$. We will use the notation
\[S\stackrel{W}{=}[E,X],
\]
to denote the state-space decomposition defined by $W=(W_{exo},W_{end})$. 

Because diffeomorphisms preserve mutual information \citep{kraskov2004}, the conditional mutual information constraint holds in the mapped state space 
\[I(W_{exo}\cdot S';[W_{end}\cdot S,A] \mid W_{exo}\cdot S)=0
\]
if and only if it holds in the original space,
\[I(X';[E,A] \mid X)=0.\] 
Note that it is possible for two different matrices $W_{exo}$ and $W_{exo}'$ to satisfy the conditional mutual information constraint if they are related by an invertible matrix $U\in\mathbb{R}^{d_{exo}\times d_{exo}}$ as $W_{exo} = W_{exo}' \cdot U$ (and similarly for $W_{end}$). Hence, when we solve the maximal exogenous subspace optimization (Equations \eqref{opt:general_formulation_decoupled_a}-\eqref{opt:general_formulation_decoupled_c}), the exogenous subspace is only identified up to multiplication by an invertible matrix $U$. This can make it difficult to interpret the discovered subspace. %We found it convenient to put $M_{exo}$ into block-diagonal canonical form, so that to the extent possible, the coordinates of the mapped space are the same as in the original space.

To develop the linear formulations of the discovery problem, let us consider the database $D$ of $\{(s_i,a_i,r_i,s'_i)\}_{i=1}^N$ sample transitions collected in Algorithm~\ref{alg:framework}. We start by centering  $s_i$ and $s'_i$ by subtracting off the mean of the observed states. 
Let $\mathbf{S}\in\mathbb{R}^{N\times d},\mathbf{A}\in\mathbb{R}^{N\times k},\mathbf{R}\in\mathbb{R}^{N\times 1},\mathbf{S}'\in\mathbb{R}^{N\times d}$ be the  matrices containing the observations of $s_i,a_i,r_i,$ and $s'_i$, respectively. These are samples from the random variables $S,A,R,S'$, and we will estimate the expectations required in the optimization formulations (expected squared reward regression error and CMI values) from these samples.
Given observation matrices $\mathbf{S},\mathbf{S}'$, we can write the corresponding exogenous and endogenous states as $\mathbf{X}=\mathbf{S}\cdot W_{exo},\mathbf{X}'=\mathbf{S}'\cdot W_{exo},\mathbf{E}=\mathbf{S}-\mathbf{S}\cdot W_{exo}\cdot W_{exo}^{\top},$ and $\mathbf{E}'=\mathbf{S}'-\mathbf{S}'\cdot W_{exo}\cdot W_{exo}^{\top}$. 

The exogenous reward regression problem takes a particularly simple form if we adopt linear regression. Let $\hat{\mathbf{X}}= \mathbf{S}\cdot W_{exo}$ be the matrix of estimated exogenous state vectors, and let $w_R^*$ be the fitted coefficients of the linear regression. This coefficient vector can be computed as the solution of the usual least squares problem
\begin{equation}\label{opt:linear_reward_regression}
w_R^*=\argmin_{w_R\in\mathbb{R}^{d_{exo}}} \|\hat{\mathbf{X}}\cdot w_R-\mathbf{R}\|_2^2=\argmin_{w_R\in\mathbb{R}^{d_{exo}}} \{(\hat{\mathbf{X}}\cdot w_R-\mathbf{R})^\top\cdot(\hat{\mathbf{X}}\cdot w_R-\mathbf{R})\}=(\hat{\mathbf{X}}^{\top}\hat{\mathbf{X}})^{-1}\hat{\mathbf{X}}^{\top}\mathbf{R}.
\end{equation}
This gives us the optimization objective for the linear formulation. Now let us consider how to express the conditional mutual information constraints. 

Estimating mutual information (and conditional mutual information) has been studied extensively in machine learning. Recent work exploits variational bounds \citep{donsker1983, nguyen2007, nowozin2016, barber2003, blei2017} to enable differentiable end-to-end estimation of mutual information with deep nets \citep{belghazi2018, poole2019, alemi2018, hjelm2019, oord2018}. Despite their promise, mutual information estimation by maximizing variational lower bounds is challenging due to inherent statistical limitations \citep{McAllester2020}. Alternative approaches for estimating the mutual information include k-nearest neighbors \citep{kraskov2004}, ensemble estimation \citep{moon2017}, jackknife estimation \citep{zeng2018}, kernel density estimation \citep{kandasamy2015,han2020}, and Gaussian copula methods \citep{singh2017}. All of these require substantial computation, and some of them also require delicate hyperparameter tuning. Extending them to estimate conditional mutual information raises additional challenges. 

We have chosen instead to replace conditional mutual information with a quantity we call the conditional correlation coefficient (CCC). To motivate the CCC, assume that variables $X,Y,Z$ are distributed according to a multivariate Gaussian distribution. In this case, it is known \citep{baba2004partial} that $X$ and $Y$ are conditionally independent given $Z$, if and only if \[\Sigma_{XY}-\Sigma_{YZ}\Sigma^{-1}_{ZZ}\Sigma_{XZ}=\mathbf{0},\] where $\Sigma_{AA}$ is the covariance matrix of $A$ and $\Sigma_{AB}$ is the cross-covariance matrix of $A$ and $B$. We can normalize the above expression to obtain the normalized cross-covariance matrix 
\begin{equation} \label{eq:CCC}
V(X,Y,Z)=\Sigma_{XX}^{-1/2}(\Sigma_{XY} - \Sigma_{XZ}\Sigma_{ZZ}^{-1}\Sigma_{ZY})\Sigma_{YY}^{-1/2}=\mathbf{0}.
\end{equation}
It is not hard to see that Equation~\eqref{eq:CCC} holds if and only if
\[\tr(V^\top(X,Y,Z)\cdot V(X,Y,Z))=0,\]
where $\tr(\cdot)$ is the trace function.
We call the quantity $\tr(V^\top(X,Y,Z)\cdot V(X,Y,Z))$ the \textit{conditional correlation coefficient} (CCC), and we denote it by $CCC(X,Y\mid Z)$.\footnote{In \cite{dietterich2018}, we referred to this quantity as the Partial Correlation Coefficient (PCC), but this was an error. While the PCC can be used to determine conditional independence for Gaussians, it is a different quantity.} Because \eqref{eq:CCC} involves matrix inversion, we apply Tikhonov regularization \citep{tikhonov1977} to all inverse matrices with a small positive constant $\lambda>0$ for numerical stability. For instance, $\Sigma_{XX}^{-1/2}$ becomes $(\Sigma_{XX}+\lambda\cdot\mathbb{I}_n)^{-1/2}$, where $n$ is the size of random vector $X$. We note that the CCC is inspired by the conditional cross-covariance operator in Reproducing Kernel Hilbert Space (RKHS), which serves as a measure of conditional dependence of random variables \citep{fukumizu2008kernel,fukumizu2004dimensionality}.

We can now express the two-phase exogenous discovery problem as
\begin{equation}\label{opt:linear-decoupled-formulation}
\begin{split}
&\min_{w_R\in \mathbb{R}^{d_{exo}^*}}\|\mathbf{S}\cdot W_{exo}^*\cdot w_R-\mathbf{R}\|_2^2 \\
& \mbox{where }\\
& d_{exo}^*,W_{exo}^*=\argmax_{d_{exo}\in\{0,\dots,d\},W_{exo}\in \mathbb{R}^{d\times d_{exo}}} d_{exo} \\
&\hspace{2cm} \mbox{subject to } W_{exo}^{\top}W_{exo}=\mathbb{I}_{d_{exo}}\\
&\hspace{2cm} CCC(\mathbf{S}'W_{exo}; [\mathbf{S}-\mathbf{S}W_{exo}W_{exo}^{\top}, \mathbf{A}] \mid  \mathbf{S}W_{exo})<\epsilon.
\end{split}
\end{equation}

Although we have written the outer objective in terms of linear regression, this is not essential. Once the optimal linear projection matrix $W_{exo}^*\in\mathbb{R}^{d\times d_{exo}}$ has been determined and the reward regression dataset $D_{exo}$ has been constructed, any form of regression---including nonlinear neural network regression---can be employed.

\section{Algorithms for Decomposing a 2-Exogenous State MDP into Exogenous and Endogenous Components}\label{sec:algorithms}
This section introduces two practical algorithms for addressing the exogenous subspace discovery problem. The first algorithm, \grds{} (Global Rank Descending Scheme), is based on the linear hierarchical formulation of Equation~\eqref{opt:linear-decoupled-formulation}. It initializes $d_{exo}$ to $d$ and decreases $d_{exo}$ one dimension at a time until it can find a $W_{exo}$ matrix whose CCC is near zero. We refer to it as a ``global'' scheme, because it must solve a series of global manifold optimization problems. The second algorithm, \sras{} (Stepwise Rank-Ascending Scheme), starts with $d_{exo} := 0$ and constructs the $W_{exo}$ matrix by adding one column at a time as long as it can keep CCC near zero. \sras{} only needs to solve one-dimensional manifold optimization problems, so it has the potential to be faster.

\subsection{\grds{}: Global Rank Descending Scheme}
Algorithm~\ref{alg:Global} gives the pseudo-code for the global rank descending scheme, \grds. \grds{} solves the inner objective (Equation~\ref{opt:linear-decoupled-formulation}) by iterating from $d_{exo} := d$ down to zero. Instead of treating the $CCC < \epsilon$ condition as a constraint, we put $CCC$ into the objective and minimize it (line 6). % I tried to use a cross-reference for the 6, but it didn't work
If the optimization finds a $W_{exo}$ with $CCC<\epsilon$, we know that this gives the maximum value,  $d^*_{exo}$. Hence, we can halt and return $W_{exo}$ as the solution. 

One might hope that we could use a more efficient search procedure, such as binary search, to find $d^*_{exo}$. Unfortunately, because not all subsets of the maximal exogenous subspace are valid decompositions (Theorem~\ref{theorem:bad-subsets}), it is possible for an exogenous subset with $\hat{d} < d^*_{exo}$ to violate the $CCC < \epsilon$ constraint. Hence, binary search will not work.

The orthonormality constraint in the minimization (line 6) forces the weight matrix $W_{exo}$ to lie on a Stiefel manifold \citep{Stiefel1935}. Hence, line 6 seeks to minimize a function on a manifold, a problem to which we can apply familiar tools for Euclidean spaces such as gradient descent, steepest descent and conjugate gradient. Several optimization algorithms exist for optimizing on Stiefel manifolds \citep{Jiang2015,Absil2007,Edelman1999}. 
Manifold optimization on a Stiefel manifold has previously been considered by \cite{bach2003} in the context of Independent Component Analysis, but in their case the linearly projected data are subsequently mapped to an RKHS.

\begin{algorithm}[t!]
\begin{algorithmic}[1]
\STATE {\bf Inputs:} A database of transitions $\{(s_i,a_i,s'_i)\}_{i=1}^N$ provided as matrices $\mathbf{S},\mathbf{A},\mathbf{S'}$
\STATE {\bf Output:} The exogenous state projection matrix $W_{exo}^*$ 
\FOR {$d_{exo}=d$ \textbf{down to} $1$}
\STATE Set \;\;\;\; $Y(W_{exo})\gets [\mathbf{S}-\mathbf{S}W_{exo}W_{exo}^{\top}, \mathbf{A}]$ 
\STATE Solve the following optimization problem
\begin{align*}
W_{exo}^* := \quad\quad&\\
\argmin_{W_{exo}\in \mathbb{R}^{d\times  d_{exo}}}&CCC(\mathbf{S}'W_{exo}; Y(W_{exo}) \mid  \mathbf{S}W_{exo})\\
\textrm{subject to } &W_{exo}^{\top}W_{exo}=\mathbb{I}_{d_{exo}}
\end{align*}
\STATE Set \;\;\;\; $CCC \leftarrow CCC(\mathbf{S}'W_{exo}^*; Y(W_{exo}^*) \mid  \mathbf{S}W_{exo}^*)$ 
\IF {$CCC < \epsilon$}
\RETURN $W_{exo}^*$
\ENDIF
\ENDFOR
\RETURN null projection $\mathbf{0}$
\end{algorithmic}
\caption{\grds{}: Global Rank-Descending Scheme}
\label{alg:Global}
\end{algorithm}

\subsection{Analysis of the Global Rank-Descending Scheme}
\label{sec:global-analysis}
In this section, we study the properties of the global rank-descending scheme. We assume that we fit the exo reward function using linear regression  \eqref{opt:linear_reward_regression}, since this simplifies our analysis. Directly analyzing Algorithm \ref{alg:Global} is hard, because it (i) uses the CCC objective as a proxy for conditional independence, (ii) involves estimation errors due to having a finite number of samples, and (iii) involves approximations in the optimization (both from numerical errors and from the $\epsilon$ threshold). To side-step these challenges, we analyze an \textit{oracle variant} of our setting where the MDP is tabular (i.e., states and actions are discrete), we have access to the true joint distribution $P(S,A,S')$, and we have perfect algorithms for solving all optimization problems (including the exo reward linear regression). Access to $P(S,A,S')$ is equivalent to having an infinite training sample collected by visiting all states and executing all actions infinitely-many times so that estimation errors vanish when computing the conditional mutual information and the expected value of the residual error in \eqref{opt:linear_reward_regression}. We formalize this with the following two definitions:

\begin{definition}[Fully Randomized Policy] \label{def:fully-randomized}
An exploration policy $\pi_x$ is \emph{fully randomized} for a tabular MDP with action space $\mathcal{A}$ if $\pi_x$ assigns non-zero probability to every possible action $a\in \mathcal{A}$ in every state $s \in \mathcal{S}$.
\end{definition}

\begin{definition}[Admissible MDP] \label{def:admissible-mdp}
A tabular MDP is \emph{admissible} if every fully-randomized policy will visit every state in the MDP infinitely often. 
\end{definition}
Note that if an MDP is admissible, then the exogenous Markov Reward Process must be ergodic so that it visits every exogenous state infinitely often (and is aperiodic). Otherwise, it would be impossible for the fully-randomized policy to visit every state in the full MDP infinitely often.

Under this oracle setting, we prove that the global rank-descending scheme returns the unique exogenous subspace of maximum rank. In practice, if we have a sufficiently representative sample of $\langle s, a, s', r\rangle$ tuples and the CCC captures conditional independence reasonably well, we can hope that our methods will still give useful results.

Algorithm~\ref{alg:rank-descending-oracle} shows the oracle version of \grds{}. It is identical to \grds{} except that the optimization step of minimizing the CCC is replaced by the following feasibility problem:
% todo: rename this equation, as it is no longer the simplified objective
\begin{equation}\label{opt:simplified-objective}
\begin{gathered}
\mbox{Find }W_{exo}\in \mathbb{R}^{d \times d_{exo}} \mbox{ such that:}\\
W_{exo}^{\top}W_{exo}=\mathbb{I}_{d_{exo}}\\
I(S'W_{exo}; [S-SW_{exo}W_{exo}^{\top}, A] \mid  SW_{exo})=0.
\end{gathered}
\end{equation}

\begin{algorithm}[t!]
\begin{algorithmic}[1]
\STATE {\bf Input:} Joint distribution $P(S,A,S')$ corresponding a fully-randomized policy controlling an admissible 2-Exogenous state MDP with maximal exogenous subspace $\mathcal{X}$ defined by $W^*_{exo}$
\STATE {\bf Output:} Matrix $\hat{W}^*_{exo}$ 
\FOR {$d_{exo}=d$ \textbf{down to} $1$}
\STATE Solve the following system of equations for $W_{exo}\in\mathbb{R}^{d,d_{exo}}$:
\begin{equation*}
\begin{gathered}
W_{exo}^{\top}W_{exo}=\mathbb{I}_{d_{exo}}\\
I(S'W_{exo}; [S-SW_{exo}W_{exo}^{\top}, A] \mid  SW_{exo})=0.
\end{gathered}
\end{equation*}
\IF {the above system is feasible with solution $W_{exo}$ of rank $d_{exo}$}
\STATE $\hat{W}^*_{exo} := W_{exo}$
\RETURN $\hat{W}_{exo}^*$
\ENDIF
\ENDFOR
\RETURN null matrix $\mathbf{0}$
\end{algorithmic}
\caption{Oracle-\grds{} for the Full Setting}
\label{alg:rank-descending-oracle}
\end{algorithm}

\begin{theorem}\label{theorem:correctness}
The Oracle-\grds{} algorithm returns a matrix $W_{exo}$ such that
\begin{enumerate}[(a)]
\item the subspace $\mathcal{X}$ defined by $W_{exo}$ and the subspace $\mathcal{E}$ defined as the orthogonal complement of $\mathcal{X}$ form a valid exo/endo decomposition;
\item the subspace $\mathcal{X}$ has maximal dimension over all valid exo/endo decompositions; and
\item the subspace $\mathcal{X}$ is unique and contains all other exogenous subspaces $\tilde{\mathcal{X}}$ that could form valid exo/endo decompositions.
\end{enumerate}
\end{theorem}

\begin{proof}
To prove property (a), first note that we can define the joint distribution $P(X= W_{exo}W_{exo}^\top s,E=s-W_{exo}W_{exo}^\top s)=P(S=s)$, because each column of $W_{exo}$ is a unit vector and they are orthogonal. Because $W_{exo}$ is a feasible solution to the manifold optimization Problem \ref{opt:simplified-objective} and the conditional mutual information is zero, we know from Theorem \ref{theorem:ecmi} that $P(S'|S,A)$ factors as $P(X'|X)P(E'|X,E,A,X')$. Hence, it is a valid exo/endo decomposition according to Theorem~\ref{theorem:exo-DBN}. 

Property (b) follows from the fact that $d_{exo}$ is the largest value that yields a feasible solution to Problem \ref{opt:simplified-objective}. 

To establish property (c), we need three lemmas, which are the vector space versions of Theorem~\ref{theorem:decomposition-union}, Corollary~\ref{cor:decomposition-uniqueness}, and Corollary~\ref{cor:decomposition-containment}. We state them here and give the proofs in Appendix~\ref{app:vector-space-decomposition-lemmas}.

\begin{lemma}[Union of Exo/Endo Decompositions]\label{lemma:linear-union}
Let $[\mathcal{X}_1,\mathcal{E}_1]$ and $[\mathcal{X}_2, \mathcal{E}_2]$ be two full exo/endo decompositions of an MDP $\mathcal{M}$ with state space $\mathcal{S}$, where $\mathcal{X}_1 = \{W_1^{\top} s : s \in \mathcal{S}\}$ and $\mathcal{X}_2 = \{W_2^{\top} s : s \in \mathcal{S}\}$ and where $W_1^\top W_1 = \mathbb{I}_{d_1\times d_1}$ and $W_2^\top W_2 = \mathbb{I}_{d_2\times d_2}$, $1\leq d_1,d_2 \leq d$.
Let $\mathcal{X} = \mathcal{X}_1 + \mathcal{X}_2$ be the subspace formed by the sum of subspaces $\mathcal{X}_1$ and $\mathcal{X}_2$, and let $\mathcal{E}$ be its complement. It then holds that the state decomposition $S=[E,X]$ with $E \in \mathcal{E}$ and $X \in \mathcal{X}$ is a valid full exo/endo decomposition of $\mathcal{S}$.
\end{lemma}

\begin{lemma}[Unique Maximal Subspace]\label{cor:unique-maximal-subspace}
The maximal exogenous vector subspace $\mathcal{X}_{max}$ defined by $W_{exo,max}$ is unique.
\end{lemma}

\begin{lemma}\label{lemma:maximal-containment}
Let $W_{exo,max}$ define the maximal exogenous vector subspace $\mathcal{X}_{max}$, and let $W_{exo}$ define any other exogenous vector subspace $\mathcal{X}$. Then $\mathcal{X} \sqsubseteq \mathcal{X}_{max}$. 
\end{lemma}

These three lemmas establish property (c) and complete the proof of Theorem~\ref{theorem:correctness}.
\end{proof}

This concludes our analysis of the Oracle-\grds{} (Algorithm~\ref{alg:rank-descending-oracle}). 

How well does this analysis carry over to the non-oracle \grds{} algorithm? \grds{} departs from the oracle version in three ways. First, \grds{} employs the CCC in place of conditional mutual information (CMI). This may assign non-zero CCC values to $W_{exo}$ matrices that actually have zero CMI. This will cause \grds{} to under-estimate $d_{exo}$, the dimensionality of the exogenous state space. Second, \grds{} only requires CCC to be less than a parameter $\epsilon$. If $\epsilon$ is large, then \grds{} may stop too soon and over-estimate $d_{exo}$. Hence, by introducing $\epsilon$, \grds{} is able to compensate somewhat for the failures of CCC. Third, the database of transitions is not infinite, so the value of CCC that \grds{} computes may be too high or too low. This in turn may cause $d_{exo}$ to be too small or too large. In our experiments, we will compare the estimated $d_{exo}$ to our understanding of its true value. 

\subsection{Stepwise Algorithm \sras{}}
\label{sec:stepwise-algorithm}
The global scheme computes the entire $W_{exo}$ matrix at once. In this section, we introduce an alternative \textit{stepwise} algorithm, the Stepwise Rank Ascending Scheme (\sras{}, see Algorithm~\ref{alg:Stepwise}), which constructs the matrix $W_{exo}$ incrementally by solving a sequence of small manifold optimization problems. 
\begin{algorithm}[t!]
\begin{algorithmic}[1]
\STATE {\bf Inputs:} A database of transitions $\{(s_i,a_i,r_i,s'_i)\}_{i=1}^N$
\STATE {\bf Output:} The exogenous state projection matrix $W_{exo}$ 
\STATE Initialize $W_{exo}\leftarrow[~]$, $W_{temp}\leftarrow[~]$, $C_x\leftarrow[~]$, $k\leftarrow 0$
\REPEAT
\STATE $N\leftarrow \textrm{orthonormal basis for the null space of }C_x$
\STATE Solve the following optimization problem
\begin{align*}
\hat{w} := \quad\quad\quad&\\
\argmin_{w\in \mathbb{R}^{(d-k)\times 1}}&CCC(\mathbf{S}'[W_{temp},N^{\top}w];\mathbf{A}\mid \mathbf{S}[W_{temp}, N^{\top}w])\\
\textrm{subject to } &w^{\top}w=1
\end{align*}
\STATE $w_{k+1} \leftarrow N^{\top}\hat{w}$ %express $\tilde{w}$ in original $\mathbf{R}^{d_{exo}}}$ space 
\STATE $C_x \leftarrow C_x \cup \{w_{k+1}\}$
\STATE $CCC_{sim}\leftarrow CCC(\mathbf{S}'[W_{temp},w_{k+1}];\mathbf{A}\mid \mathbf{S}[W_{temp}, w_{k+1}])$
\IF {$CCC_{sim} < \epsilon$}
\STATE $W_{temp} \leftarrow W_{temp} \cup \{w_{k+1}\}$
\STATE $\mathbf{E}\leftarrow\mathbf{S}-\mathbf{S}W_{temp}W_{temp}^{\top}$
\STATE $CCC_{full}\leftarrow CCC(\mathbf{S}'W_{temp};[\mathbf{E},\mathbf{A}]\mid \mathbf{S}W_{temp})$
\IF{$CCC_{full} < \epsilon$}
\STATE $W_{exo} \leftarrow W_{temp}$
\ENDIF
\ENDIF
\STATE $k\leftarrow k+1$
\UNTIL $k=d$
\RETURN $W_{exo}$
\end{algorithmic}
\caption{Stepwise Rank Ascending Scheme: \sras{}}
\label{alg:Stepwise}
\end{algorithm}

\sras{} maintains the current partial solution $W_{exo}$, a temporary matrix $W_{temp}$ that may be extended to update $W_{exo}$, a set of all candidate column vectors generated so far, $C_x$, and an orthonormal basis $N$ for the null space of $C_x$. ($N$ is a matrix with a number of columns equal to the dimension of the null space of $C_x$.) The set of candidate column vectors $C_x$ contains all of the column vectors in $W_{exo}$ and possibly some additional vectors that were rejected for violating the full CCC constraint, as we will discuss below.

Suppose we have already found the first $k$ columns of $W_{exo} = [w_1, w_2, \ldots, w_k]$. To ensure that the new column $w_{k+1}$ is orthogonal to all $k$ previous vectors, we restrict $w_{k+1}$ to lie in the space defined by $N$ by requiring it to have the form $w_{k+1} = N^{\top}w$. This ensures that it is orthogonal to all columns of $W_{exo}$ and to any additional vectors in $C_x$. 

In Line 6, we compute a new candidate vector $\hat{w}$ by solving a simplified CCC minimization problem on the $(d-k)\times1$-dimensional Stiefel manifold. Recall that the full objective $I(X' ;[E,A]\mid X)=0$ seeks to enforce the conditional independence $X'\independent E, A \mid X$. This requires us to know $X$ and $E$, whereas at this point in the algorithm, we only know a portion of $X$, and we therefore do not know $E$ at all. We circumvent this problem by using the \textit{simplified objective} $I(X'_k; A\mid X_1, \ldots, X_k)$ (approximated via the CCC). This objective ensures that $A$ has no effect on the exogenous variables $X'$ in the next time step, which eliminates the edge $A \rightarrow X'$, but it does not protect against the possibility that $A$ causes a change in some chain of endogenous variables that affect $X$ in some subsequent time step. Hence, the simplified objective is a necessary but not sufficient condition for $X$ to be a valid exogenous subspace. See Appendix~\ref{app:simplified_setting} for a detailed discussion of this point.

Lines 7 and 8 compute the new candidate vector $w_{k+1}$ by mapping $\hat{w}$ into the null space defined by $N$ and then adding it to $C_x$. In Line 9, we compute $CCC_{sim}$, the value of the simplified objective (which is the same as the value that minimized the objective in Line 6). In Line 10, we check whether this is less than $\epsilon$. If not, we increment $k$ and loop back to Line 5 and find another $\hat{w}$ vector. But if $CCC_{sim} < \epsilon$, then in Lines 11-13, we compute the corresponding $\bf{E}$ matrix and compute $CCC_{full}$, the CCC of the full objective. In Line 14, we check whether $CCC_{full} < \epsilon$. If so, then we have a valid new column to add to $W_{exo}$. If not, we increment $k$ and loop back to Line 5. 

Recall that not all subsets of the maximal exogenous subspace are themselves valid exogenous subspaces that satisfy the full CCC constraint (Theorem~\ref{theorem:bad-subsets}). Hence, it is important that \sras{} does not terminate when adding a candidate vector to $W_{exo}$ causes the full constraint to be violated. Note, however, that every subset of the maximal exogenous subspace must satisfy the simplified objective, because otherwise, the action variable is directly affecting one of the exogenous state variables. 

To allow \sras{} to continue making progress when the full constraint is violated, the algorithm maintains the matrix $W_{temp}$. This matrix contains all of the candidates that have satisfied the simplified objective.  If a subsequent candidate $w_{k+1}$ allows $W_{temp}$ to satisfy the full constraint, then we set $W_{exo}$ to $W_{temp}$ and continue. The algorithm terminates when $k=d$. 

The primary advantage of \sras{} compared to \grds{} is that the CCC minimization problems have dimension $(d-k)\times 1$. However, \grds{} can halt as soon as it finds a $W_{exo}$ that satisfies the full CCC objective, whereas \sras{} must solve all $d$ problems. We can introduce heuristics to terminate $d$ early. For example, we can monitor the residual variance $\|\hat{\mathbf{X}}\cdot w_R-\mathbf{R}\|_2^2$ of the reward regression. This decreases monotonically as columns are added to $W_{exo}$, and when those decreases become very small, we can terminate \sras{}. This can make \sras{} very efficient when the exogenous subspace has low rank $d_{exo}$ relative to the rank $d$ of the full state space. In such cases, \grds{} must solve $d - d_{exo} + 1$ large manifold optimization problems of dimension at least $d \times d_{exo}$, whereas \sras{} must only solve $d_{exo}$ problems of dimension $(d-k)\times 1$.  

What can we say about the correctness of \sras{}? First, in an oracle version of \sras{} (where the MDP is admissible, the data are collected using a fully-randomized policy, and CMI is computed instead of CCC), the $W_{exo}$ matrix returned by \sras{} defines a valid exo/endo decomposition. This is because it satisfies the full CMI constraint. However, it would not necessarily define the maximal exogenous subspace, because the $w$ vectors found using the simplified objective and stored in $W_{temp}$ might not be a subset of a satisfying $W_{exo}$ matrix.  Of course, because the actual \sras{} algorithm introduces the CCC approximation and only requires the CCC to be less than $\epsilon$, we do not have any guarantee that the $W_{exo}$ matrix returned by \sras{} defines a valid exogenous subspace. We now turn to experimental tests of the algorithms to see whether they produce useful results despite their several approximations.

We conducted a series of experiments to understand the behavior of our algorithms. In addition to \grds{} and \sras{}, we defined a third algorithm, Simplified-\grds{} that applies the simplified objective $CCC(S'W_{Exo} ; A | SW_{exo})$ in Line 6 of Algorithm~\ref{alg:Global} but then still uses the full objective in Line 7. Like \sras{}, Simplified-\grds{} will always return a valid $W_{exo}$, but it may not be maximal. 

\section{Experimental Study}\label{sec:experiments}

In this section, we present experiments to address the following research questions:
\begin{itemize}
    \item[RQ1:] Do our methods speed up reinforcement learning in terms of sample complexity? In terms of total CPU time?
    \item[RQ2:] Do our methods discover the correct maximal exogenous subspaces?
    \item[RQ3:] How do the algorithms behave when the MDPs are changed to have different properties: (a) rewards are nonlinear, (b) transition dynamics are nonlinear, (c) action space is combinatorial, and (d) states and actions are discrete?
    \item[RQ4:] Which of the three decomposition methods is the best to use in practical applications?
\end{itemize}

\subsection{Experimental Details}
\label{sec:setup}
 We compare five methods:
 \begin{itemize}
    \item \textit{Baseline}: Reinforcement learning applied to the full reward (sum of exogenous and endogenous components)
    \item \textit{\grds{}}: The Global Rank Descending Scheme
    \item \textit{Simplified-\grds{}}: \grds{} using the simplified objective
    \item \textit{\sras{}}: The Stepwise Rank Ascending Scheme
    \item \textit{Endo Reward Oracle}: Reinforcement learning applied with the oracle endogenous reward.
\end{itemize}
As the reinforcement learning algorithm, we employ the PPO implementation from stable-baselines3 \citep{stable-baselines3} in PyTorch \citep{pytorch}, and we model the MDPs in the OpenAI Gym framework \citep{gym}. We use the default PPO hyperparameters in stable-baselines3, which include a clip range of 0.2, a value function coefficient of 0.5, an entropy coefficient of 0, and a generalized advantage estimation (GAE) parameter of 0.95.
The policy and value networks have two hidden layers of 64 tanh units each. For PPO optimization, we employ the default Adam optimizer \citep{kingma} in PyTorch with default hyperparameters $\beta_1=0.9, \beta_2=0.999, \text{eps}=1\times 10^{-5}$ and a default learning rate of $lr_{PPO}=0.0003$. The batch size for updating the policy  and value networks is 64 samples. The discount factor in the MDPs is set to $\gamma=0.99$. The default number of steps between each policy update is $K=1536$.  The number of steps $L$ after which we compute the exo/endo decomposition and the total number of training steps in the experiment $N$ vary per experiment, but their default values are $L=3000$ and $N=60000$. We summarize all hyperparameters in Table \ref{table:hyperparams}.

In each experimental run, we maintain two instances of the Gym environment. Training is carried out in the primary instance. After every policy update, we copy the policy to the second instance and evaluate the performance of the policy (without learning) for 1000 steps. To reduce measurement variance, these evaluations always start with the same random seed.

To solve the manifold optimization problems, we apply the solvers implemented in the Pymanopt package \citep{pymanopt}. We use the Steepest Descent solver with line search with the default Pymanopt hyperparameters. For the CCC constraint, $\epsilon$ is set to $0.05$. The Tikhonov regularizer inside the CCC is set to $\lambda=0.01$. These hyperparameters were determined empirically via random search \citep{random_search}.

We employ the default library hyperparameters for PPO, Adam optimization, and manifold optimization to enable a fair comparison of the different methods. Furthermore, for online reinforcement learning algorithms, it is not feasible to perform extensive hyperparameter searches because of the cost (and risk) of interacting in the real world. Algorithms that only perform well after extensive hyperparameter search are not usable in practice. Hence, we wanted to minimize hyperparameter tuning.

To perform reward regression, we employ standard least-squares linear regression without regularization for our main experiments (Section~\ref{sec:main-experiments}). For all subsequent experiments, we switch to neural network regression, which we will describe in Section~\ref{sec:nonlinear-rewards}.

In all our MDPs, we use the default values in Table \ref{table:hyperparams} for all hyperparameters except for the number of decomposition steps $L$ and the number of training steps $L$. The former is the most critical hyperparameter; it is discussed in detail in Appendix~\ref{app:practical-considerations}. The latter is set to a number that is high enough for our methods to converge or be near the limit. Finally, in a few settings we found it beneficial to use a regression learning rate of 0.0006 instead of 0.0003 for better convergence.

For each setting, we report the values of the hyperparameters that are different from their default values. We run all experiments on a c5.4xlarge EC2 machine on AWS\footnote{https://aws.amazon.com/ec2/instance-types/c5/.}.

\begin{table}[t!]
{\footnotesize
\begin{center}
\begin{tabular}{ c c c c c c } 
 \hline
 Used for & Description & Symbol & Default Value & Fixed \\ 
 \hline\hline
 \multirow{4}{*}{PPO} & clipping parameter & - & 0.2 & Yes \\
   & value function coefficient & - & 0.5 & Yes \\
      & entropy coefficient & - & 0 & Yes \\
      & GAE parameter & - & 0.95 & Yes \\
 \hline
   \multirow{3}{*}{\begin{tabular}{l}Policy \& Value Nets\end{tabular}} & number of layers & - & 2 & Yes \\
  & units per layer & - & 64, 64 & Yes\\
  & activation function & - & tanh & Yes \\
 \hline
  \multirow{3}{*}{\begin{tabular}{l}PPO Optimization\\with Adam\end{tabular}} & Adam learning rate & $lr_{PPO}$ & 0.0003 & Yes \\
  & Adam coefficients & $\beta_1,\beta_2,\textrm{eps}$ & 0.9, 0.99, 1e-5 & Yes\\
  & batch size & - & 64 & Yes \\
 & L2 regularization & - & 0 & Yes \\
 \hline
   \multirow{4}{*}{\begin{tabular}{l}Reinforcement\\Learning\end{tabular}} & discount factor & $\gamma$ & 0.99 & Yes \\
  & policy update steps & $K$ & 1536 & Yes\\
  & total training steps & $N$ & 60000 & No\\
  & steps for decomposition & $L$ & 3000 & No\\
  & steps for exo regression& $M$ & 256 & Yes\\
  & evaluation steps & - & 1000 & Yes \\
 \hline
   \multirow{2}{*}{\begin{tabular}{l}Manifold\\Optimization\end{tabular}} & CCC threshold & $\epsilon$ & 0.05 & Yes \\
  & Tikhonov regularizer & $\lambda$ & 0.01 & Yes\\
 \hline
  \multirow{3}{*}{\begin{tabular}{l}Exo regression Net\end{tabular}} & number of layers & - & 2 & Yes \\
  & units per layer & - & 50, 25 & Yes\\
  & activation function & - & relu & Yes \\
 \hline
  \multirow{4}{*}{\begin{tabular}{l}Exo Regression\\Optimization\\with Adam\end{tabular}} & Adam learning rate & $lr_{regr}$ & 0.0003 & No \\
  & Adam coefficients & $\beta_1,\beta_2,\textrm{eps}$ & 0.9, 0.99, 1e-8 & Yes\\
  & batch size & - & 256 & Yes \\
  & L2 regularization & - & 0.00003 & Yes \\
 \hline
\end{tabular}
\end{center}
\caption{\label{table:hyperparams}Hyperparameters for High-D setting.}}
\end{table}

\subsection{Performance Comparison on High-Dimensional Linear Dynamical MDPs}
\label{sec:linear-setting}\label{sec:main-experiments}

To address RQ1 and RQ2, we define a set of high-dimensional MDPs with linear dynamics. 
Each MDP, by design, has $m$ endogenous and $n$ exogenous variables, so that $e_t\in\mathbb{R}^m$ and $x_t\in\mathbb{R}^n$. There is a single scalar action variable $a_t$ that takes 10 discrete values $(-1, -0.777, -0.555, \ldots, 0, \ldots, 0.555, 0.777, +1)$. The policy chooses one of these values at each time step. The exo and endo transition functions are
\begin{equation*}
\begin{split}
&x_{t+1} = M_{exo}\cdot x_{t} + \varepsilon_{exo}\\
&e_{t+1} = M_{end}\cdot \begin{bmatrix}e_{t}\\ x_{t}\end{bmatrix} + M_a\cdot a_t + \varepsilon_{end},
\end{split}
\end{equation*}
where $M_{exo}\in\mathbb{R}^{n\times n}$ is the transition function for the exogenous MRP; $M_{end}\in\mathbb{R}^{m \times m}$ is the transition function for the endogenous MDP involving $e_{t}$ and $x_{t}$; $M_a\in\mathbb{R}^{m}$ is the coefficient vector for the action $a_{t}$, and it is set to a vector of ones. Hence, the effect of the action $a_t$ is to add an amount specified by $a_t$ to every dimension of the endogenous vector $e_{t+1}$. The exogenous noise is $\varepsilon_{exo}\in\mathbb{R}^{n}$, and its elements are distributed according to $\mathcal{N}(0,0.09)$. The endogenous noise is $\varepsilon_{end}\in\mathbb{R}^{m}$, and its elements are distributed according to $\mathcal{N}(0,0.04)$.
The observed state vector $s_t\in\mathbb{R}^{m+n}$ is a linear mixture of the hidden exogenous and endogenous states defined as
$$s_t = M\cdot\begin{bmatrix}e_{t}\\ x_{t}\end{bmatrix},$$
where $M\in\mathbb{R}^{(m+n)\times(m+n)}$. The elements in  $M_{exo}$, $M_{end}$, and $M$ are generated according to $\mathcal{N}(0,1)$ and then each row of each matrix is normalized to sum to 0.99 for stability\footnote{Notice that all matrices $M_{exo}$, $M_{end}$, and $M$ in our synthetic linear MDPs are stochastic. Future work could explore more general classes of MDPs.}. The elements of  the initial endo and exo states are randomly initialized from a uniform distribution over $[0, 1]$.

In our first set of experiments, the reward at time $t$ is
$$R_t = R_{exo,t} + R_{end,t},$$
where $R_{exo,t} = -3\cdot\textrm{avg}(x_{t})$ is the exogenous reward, $R_{end,t} = e^{-|\textrm{avg}(e_t) - 1|}$ is the endogenous reward, and $\textrm{avg}(\cdot)$ denotes the average over a vector's elements. For this class of MDPs, the optimal policy seeks to minimize the average of $x_t$ while driving the average of $e_{t}$ to $1$. Notice that the transition dynamics of this MDP do not factor---the exogenous variables influence the dynamics of all state variables. Note also that this MDP provides non-zero reward at every time step. This is not important for the decomposition algorithms, as they ignore the reward, but it does provide many training examples for the reward regression.

We experiment in total with 6 MDPs with different choices for the numbers $n$ and $m$ of the exogenous and endogenous variables, respectively: (i) 5-D state with $m=2$ endo variables and $n=3$ exo variables; (ii) 10-D state with $m=5$ and $n=5$; (iii) 20-D state with $m=10$ and $n=10$; (iv) 30-D state with $m=15$ and $n=15$; (iv) 45-D state with $m=22$ and $n=23$; and (vi) 50-D state with $m=25$ endo variables and $n=25$ exo variables. For the 50-D setting, we use $N=100000$ training steps and $L=10000$ decomposition steps due to its higher dimensionality. Similarly, we use $N=80000$ and $L=5000$ for the 45-D MDP, and $L=5000$ for the 30-D MDP. On the other hand, we set $N=50000$ and $L=2000$ for the 5-D MDP. For each MDP, we run 20 replications with different random seeds and report the average results and standard deviations.

\subsubsection{Main Comparison Experiments}
\begin{figure}[t!]
     \centering
     \begin{subfigure}[t]{0.45\textwidth}
         \includegraphics[scale=0.4]{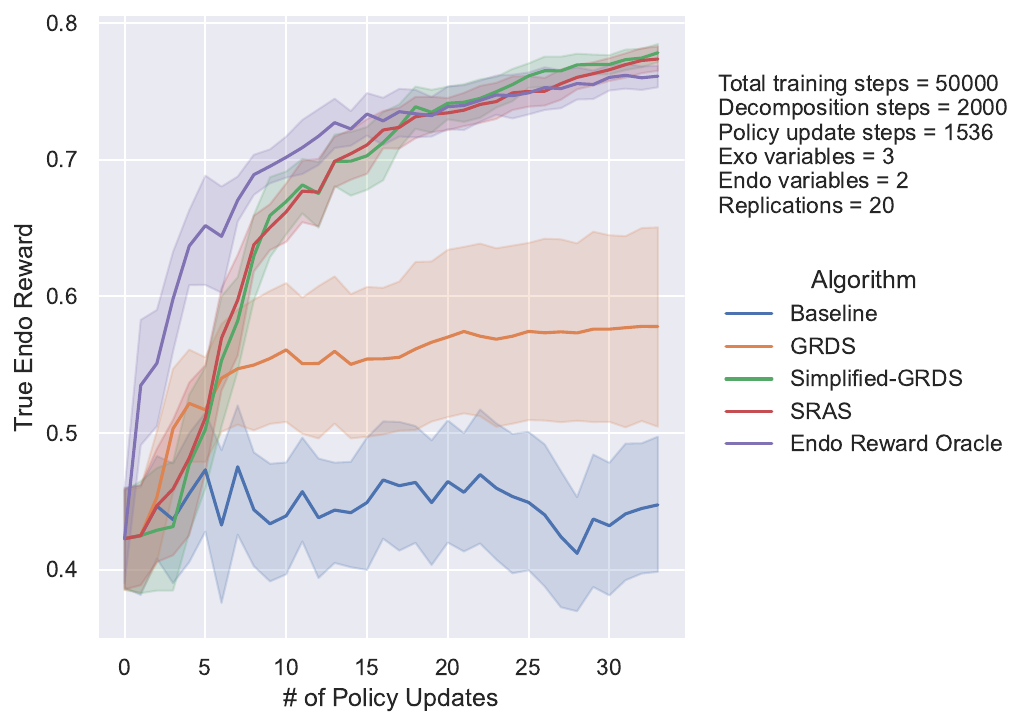}
         \caption{5-D MDP ($m=2,n=3$).}
         \label{fig:3x2}
     \end{subfigure}
     \hspace{0.0\textwidth}
     \begin{subfigure}[t]{0.45\textwidth}
         \includegraphics[scale=0.4]{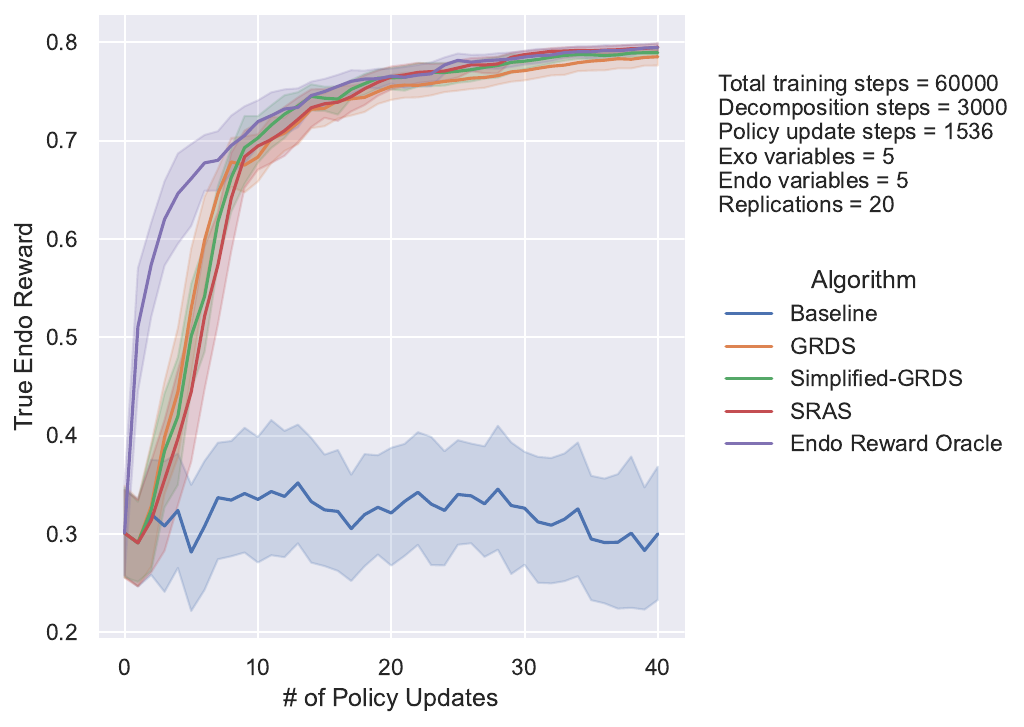}
         \caption{10-D MDP ($m=5,n=5$).}
         \label{fig:5x5}
     \end{subfigure}

     \bigskip
     \begin{subfigure}[t]{0.45\textwidth}
         \includegraphics[scale=0.4]{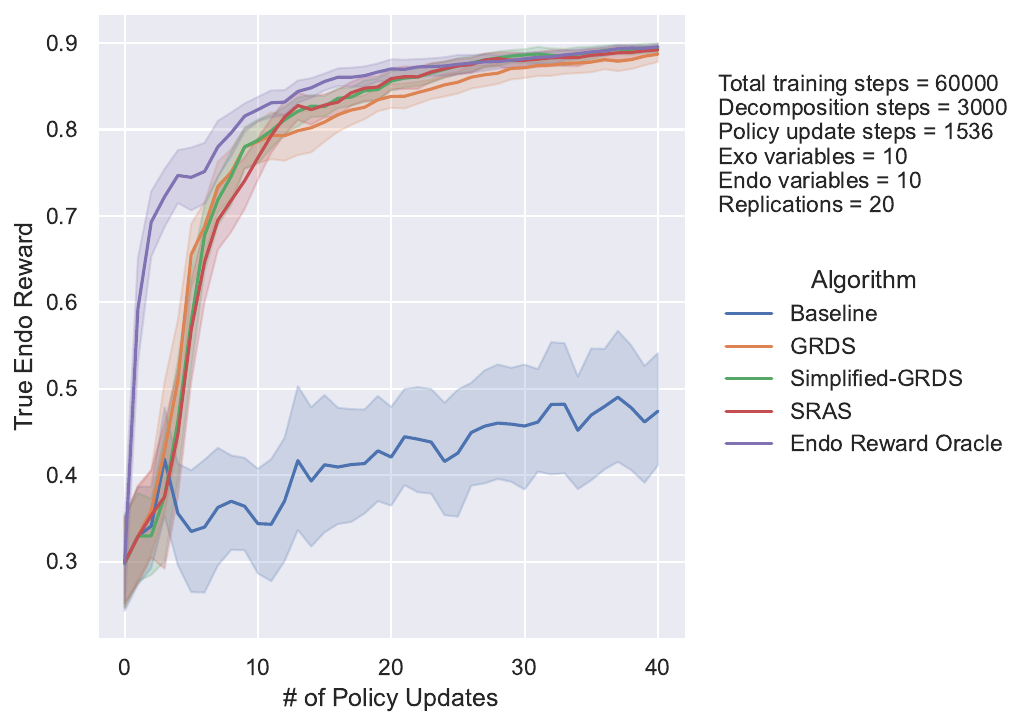}
         \caption{20-D MDP ($m=10,n=10$).}
         \label{fig:10x10}
     \end{subfigure}
     \hspace{0.0\textwidth}
     \begin{subfigure}[t]{0.45\textwidth}
         \includegraphics[scale=0.4]{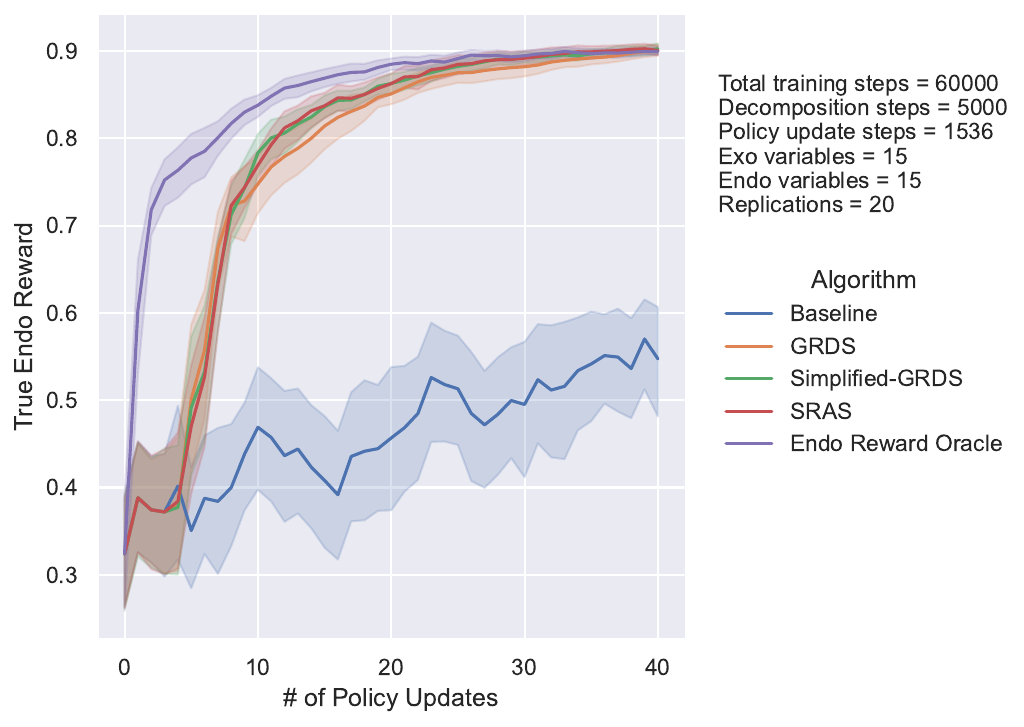}
         \caption{30-D MDP ($m=15,n=15$).}
         \label{fig:15x15}
     \end{subfigure}
     
     \bigskip
     \begin{subfigure}[t]{0.45\textwidth}
         \includegraphics[scale=0.4]{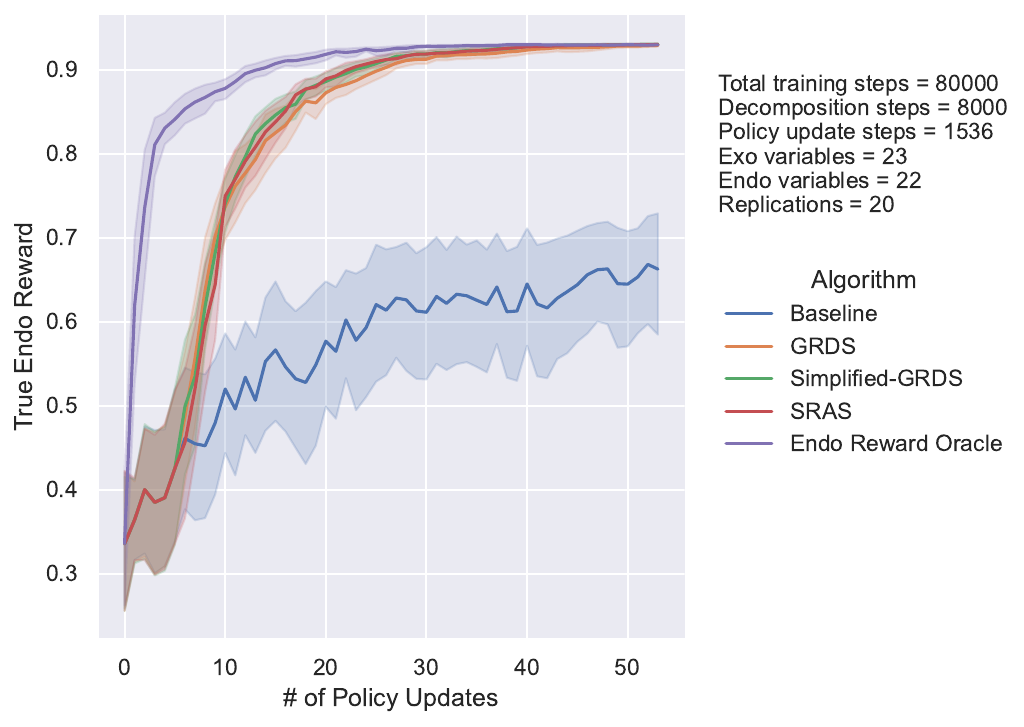}
         \caption{45-D MDP ($m=22,n=23$).}
         \label{fig:23x22}
     \end{subfigure}
     \hspace{0.0\textwidth}
     \begin{subfigure}[t]{0.45\textwidth}
         \includegraphics[scale=0.4]{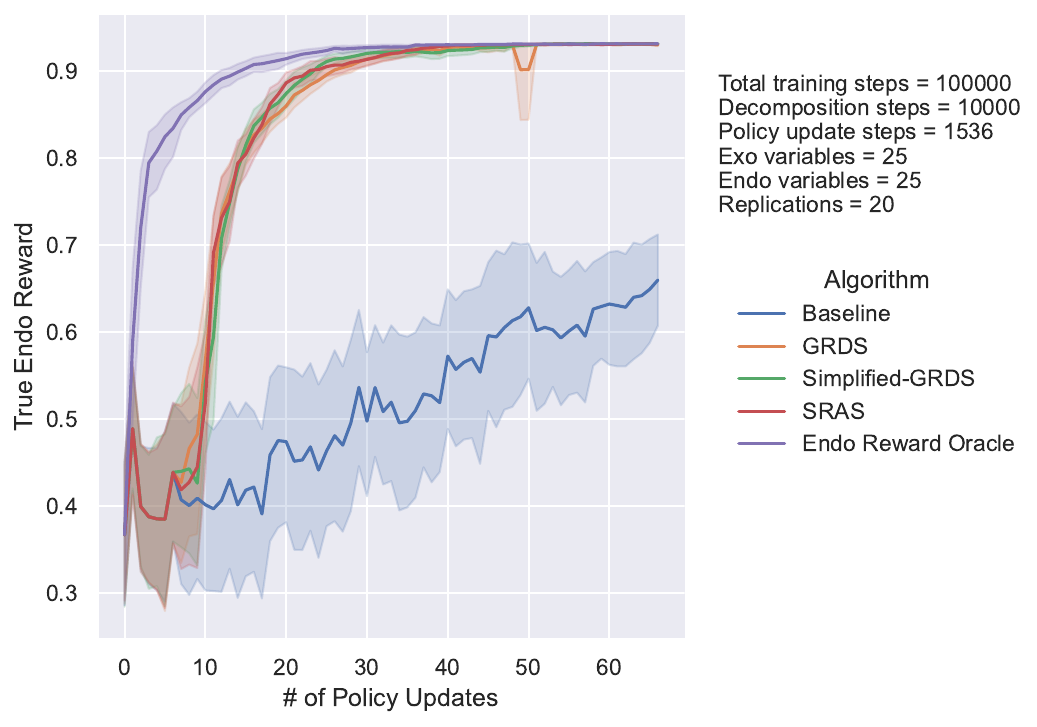}
         \caption{50-D MDP ($m=25,n=25$).}
         \label{fig:25x25}
     \end{subfigure}
    \caption{Comparison of various methods in high-dimensional linear MDPs.}
    \label{fig:linear_global_stepwise}
\end{figure}

To address RQ1, we report the performance for each of the 6 MDPs in Figure \ref{fig:linear_global_stepwise}. 
In all 6 MDPs, all of our methods far-outperform the baseline. Indeed, in the 5-D and 10-D MDPs, the baseline does not show any sign of learning, and in the larger problems, the baseline's performance has attained roughly half of the performance of our methods after 65 policy updates. A simple linear extrapolation of the baseline learning curve for the 50-D problem suggests that it will require 132 policy updates to attain the performance of the other methods. This is more than 3 times as long as the 40 updates our methods require. Hence, in terms of sample complexity, our methods are much more efficient than the baseline method.

On these MDPs, the Simplified-\grds{} and \sras{} methods are able to match the performance of the Endo Reward Oracle, which is given the correct endogenous reward from the very start. \grds{} performs very well on all MDPs except for the 5-D one, where it is still able to outperform the baseline.

RQ1 also asks whether our methods are superior in terms of CPU time. Table~\ref{table:exp1} reports the CPU time required by the various methods. The Baseline and Endo Reward Oracle consume identical amounts of time, so the table only lists the Baseline CPU time. We observe that even the slowest of our methods (\sras{} on the 50-D problem) requires only 45\% more time than the Baseline. However, if we again extrapolate the baseline to 132 policy updates, where its performance would match our methods, that would require 2997 seconds of CPU time, which is 49\% more than \sras{}. Hence, even if there is zero cost to collecting training samples in the real world, our methods are still much faster. 

\begin{table}[t!]
\caption{\label{table:exp1}Average and standard deviation for the rank of the discovered exo subspace, total execution time, and decomposition time for the high-dimensional linear MDPs.}
{\footnotesize
\begin{center}
\begin{tabular}{ c c c c c c c } 
 \hline
 Total State & Exo State & Endo State & \multirow{2}{*}{Method} & Exo Subspace & Total Time & Decomposition \\ 
 Variables & Variables & Variables & & Rank & (secs) & Time (secs) \\ 
 \hline\hline
 \multirow{4}{*}{5} & \multirow{4}{*}{3} & \multirow{4}{*}{2} & Baseline & - & 245.9$\pm$14.7 & - \\
   &  &  & \grds{} & 4.0$\pm$4.0 & 294.9$\pm$113.0 & 6.3$\pm$4.6 \\ 
   &  &  & Simplified-\grds{} & 4.0$\pm$0.0 & 297.0$\pm$115.4 & 9.6$\pm$5.0 \\ 
   &  &  & \sras{} & 3.95$\pm$0.22 & 297.1$\pm$113.3 & 14.0$\pm$14.0 \\ 
 \hline
 \multirow{4}{*}{10} & \multirow{4}{*}{5} & \multirow{4}{*}{5} & Baseline & - & 345.3$\pm$10.48 & - \\
   &  &  & \grds{} & 8.65$\pm$0.48 & 413.5$\pm$148.0 & 22.6$\pm$18.0 \\ 
   &  &  & Simplified-\grds{} & 9.0$\pm$0.0 & 413.7$\pm$144.4 & 13.1$\pm$9.1 \\ 
   &  &  & \sras{} & 8.75$\pm$0.54 & 434.3$\pm$157.6 & 34.2$\pm$46.7 \\ 
 \hline
 \multirow{4}{*}{20} & \multirow{4}{*}{10} & \multirow{4}{*}{10} & Baseline & - & 450.2$\pm$34.8 & - \\
   &  &  & \grds{} & 18.1$\pm$1.18 & 525.3$\pm$167.9 & 41.8$\pm$47.8 \\ 
   &  &  & Simplified-\grds{} & 19.0$\pm$0.0 & 513.6$\pm$141.3 & 8.6$\pm$4.4 \\ 
   &  &  & \sras{} & 18.4$\pm$0.66 & 562.4$\pm$188.5 & 86.0$\pm$72.3 \\ 
 \hline
 \multirow{4}{*}{30} & \multirow{4}{*}{15} & \multirow{4}{*}{15} & Baseline & - & 509.9$\pm$43.0 & - \\
   &  &  & \grds{} & 28.25$\pm$1.22 & 597.5$\pm$185.6 & 56.9$\pm$84.1 \\ 
   &  &  & Simplified-\grds{} & 29.0$\pm$0.0 & 584.3$\pm$136.9 & 12.5$\pm$6.0 \\ 
   &  &  & \sras{} & 27.9$\pm$1.58 & 688.9$\pm$276.5 & 177.5$\pm$234.9 \\ 
 \hline
 \multirow{4}{*}{45} & \multirow{4}{*}{23} & \multirow{4}{*}{22} & Baseline & - & 895.4$\pm$159.3 & - \\
   &  &  & \grds{} & 43.4$\pm$0.86 & 1041.7$\pm$287.6 & 84.9$\pm$154.2 \\ 
   &  &  & Simplified-\grds{} & 44.0$\pm$0.0 & 1006.3$\pm$207.8 & 13.3$\pm$8.4 \\ 
   &  &  & \sras{} & 44.0$\pm$0.0 & 1323.4$\pm$716.1 & 605.1$\pm$538.7 \\ 
 \hline
 \multirow{4}{*}{50} & \multirow{4}{*}{25} & \multirow{4}{*}{25} & Baseline & - & 1472.4$\pm$126.0 & - \\
   &  &  & \grds{} & 48.45$\pm$0.86 & 1659.6$\pm$282.7 & 81.7$\pm$105.8 \\ 
   &  &  & Simplified-\grds{} & 49.0$\pm$0.0 & 1634.5$\pm$239.1 & 15.7$\pm$9.2 \\ 
   &  &  & \sras{} & 49.0$\pm$0.0 & 2009.2$\pm$840.3 & 667.8$\pm$616.9 \\ 
 \hline
\end{tabular}
\end{center}
}
\end{table}

\subsubsection{Rank of the Discovered Exogenous Subspace}
\label{sec:linear-rank}

RQ2 asks whether our methods find the correct maximal exogenous subspaces. Our experiments revealed a surprise. Although we constructed the MDPs with the goal of creating an $n$-dimensional exogenous space and an $m$-dimensional endogenous space, our methods usually discover exogenous spaces with $n+m-1$ dimensions. Upon further analysis, we realized that because the effect of the action variable is 1-dimensional, it can only affect a 1-dimensional subspace of the $n+m$-dimensional state space. Consequently, the effective maximal exogenous subspace has dimension $n+m-1$. The results in Table~\ref{table:exp1} show that the Simplified-\grds{} always finds an exogenous subspace of the correct dimension. The exogenous space computed by \sras{} is sometimes slightly smaller on the smaller MDPs, and the space computed by \grds{} is sometimes slightly smaller on the larger MDPs. We believe the failures of \sras{} are due to the approximations that we discussed in Section~\ref{sec:stepwise-algorithm}. We suspect the failures of \grds{} reflect failures of the manifold optimization to find the optimum in high-dimensional problems. 

The fact that the exogenous space has dimension $n+m-1$ explains the relative amount of CPU time consumed by the different algorithms. \sras{} is often the slowest, because it must solve $n+m$ optimization problems whereas \grds{} and Simplified-\grds{} must only solve two (large) manifold optimization problems before terminating.

\subsection{Exploring Modifications of the MDPs}
\label{sec:nonlinear-rewards}

\begin{figure}[t!]
     \centering
     \begin{subfigure}[t]{0.46\textwidth}
         \includegraphics[scale=0.38]{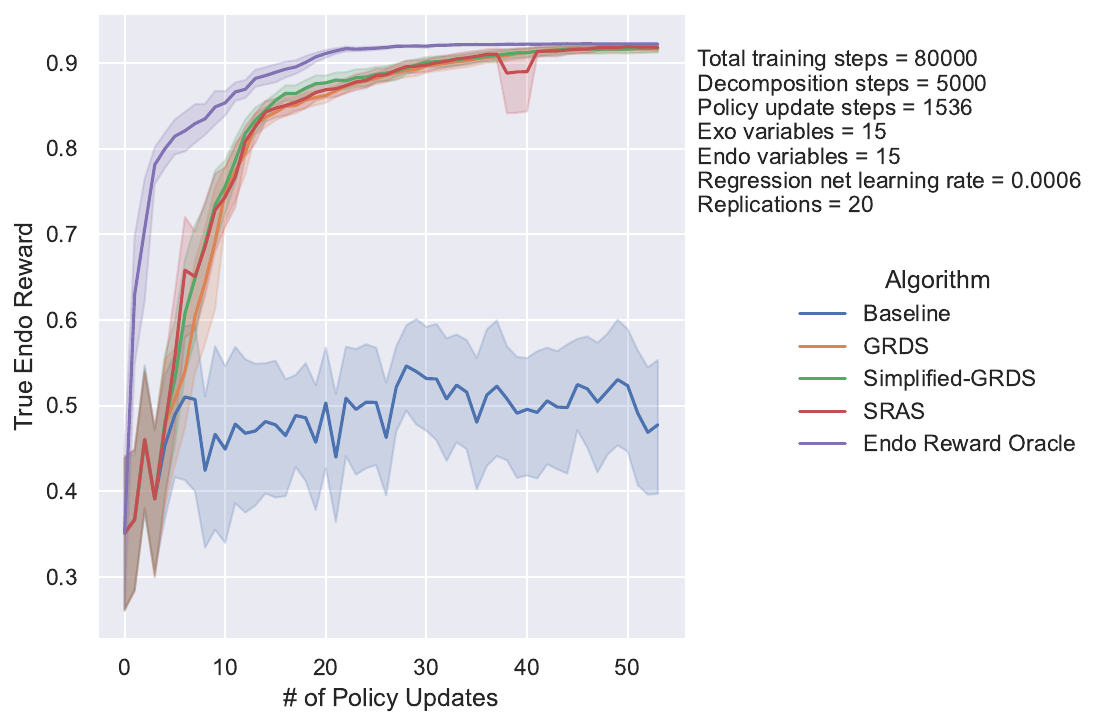}
         \caption{$R_{exo,t}^1$.}
         \label{fig:nonlinear_0}
     \end{subfigure}
     \hspace{0.0\textwidth}
     \begin{subfigure}[t]{0.44\textwidth}
         \includegraphics[scale=0.38]{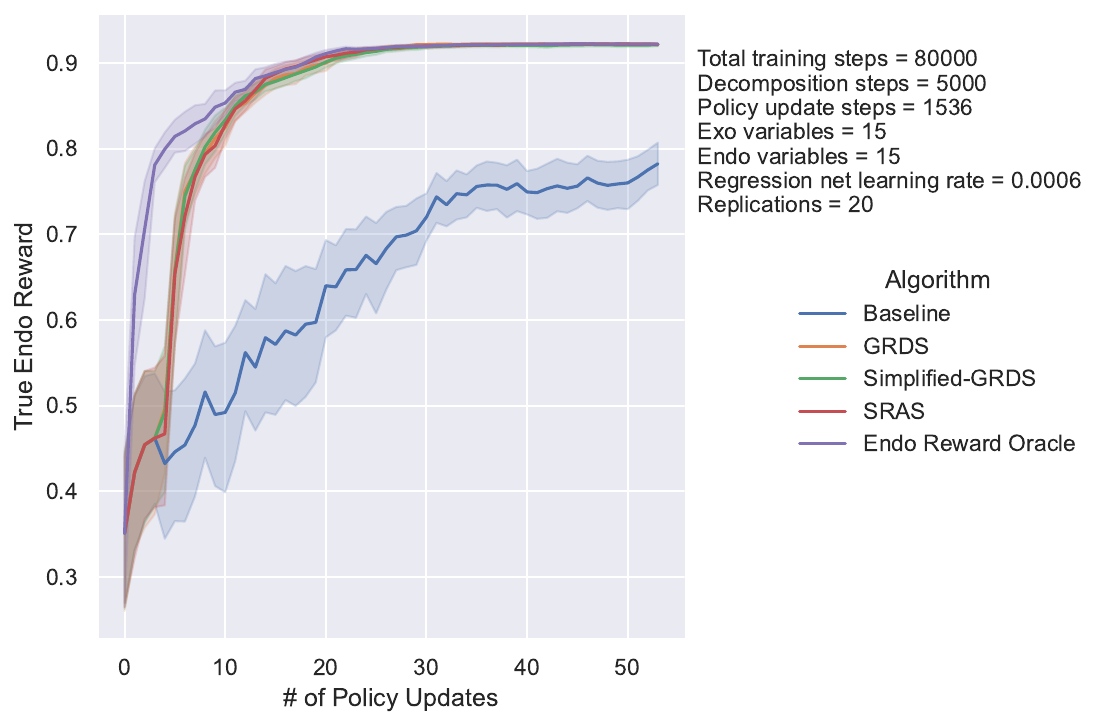}
         \caption{$R_{exo,t}^2$.}
         \label{fig:nonlinear_1}
     \end{subfigure}     
     
     \bigskip
     \begin{subfigure}[t]{0.46\textwidth}
         \includegraphics[scale=0.38]{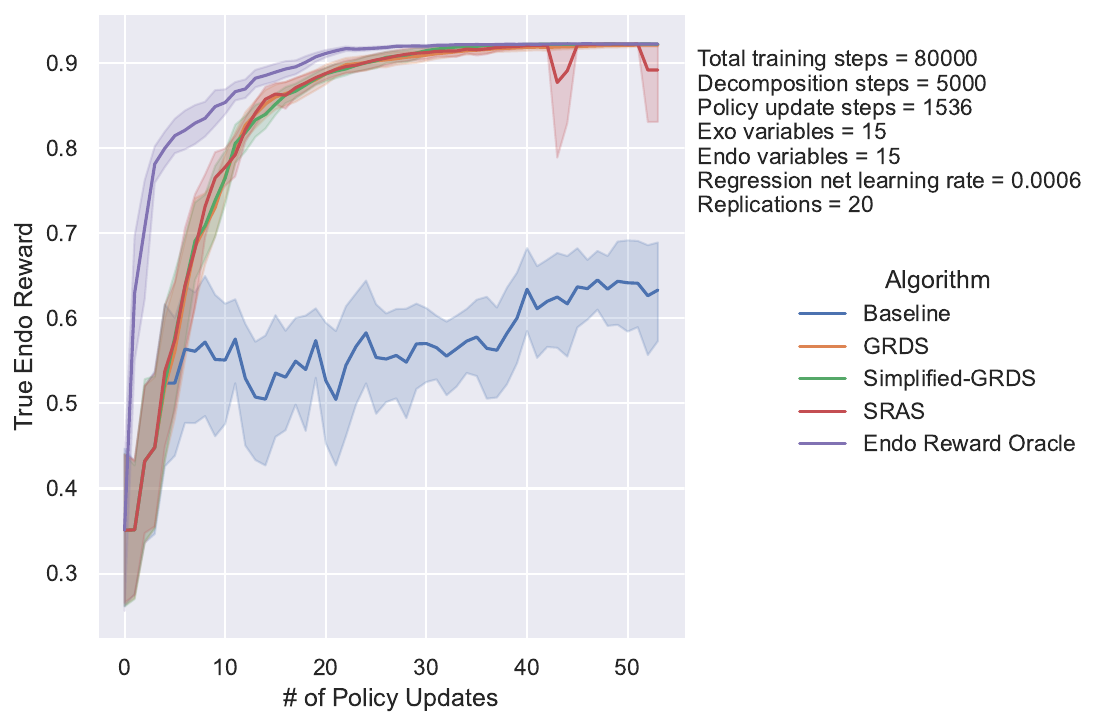}
         \caption{$R_{exo,t}^3$.}
         \label{fig:nonlinear_3}
     \end{subfigure}
     \hspace{0.0\textwidth}
     \begin{subfigure}[t]{0.44\textwidth}
         \includegraphics[scale=0.38]{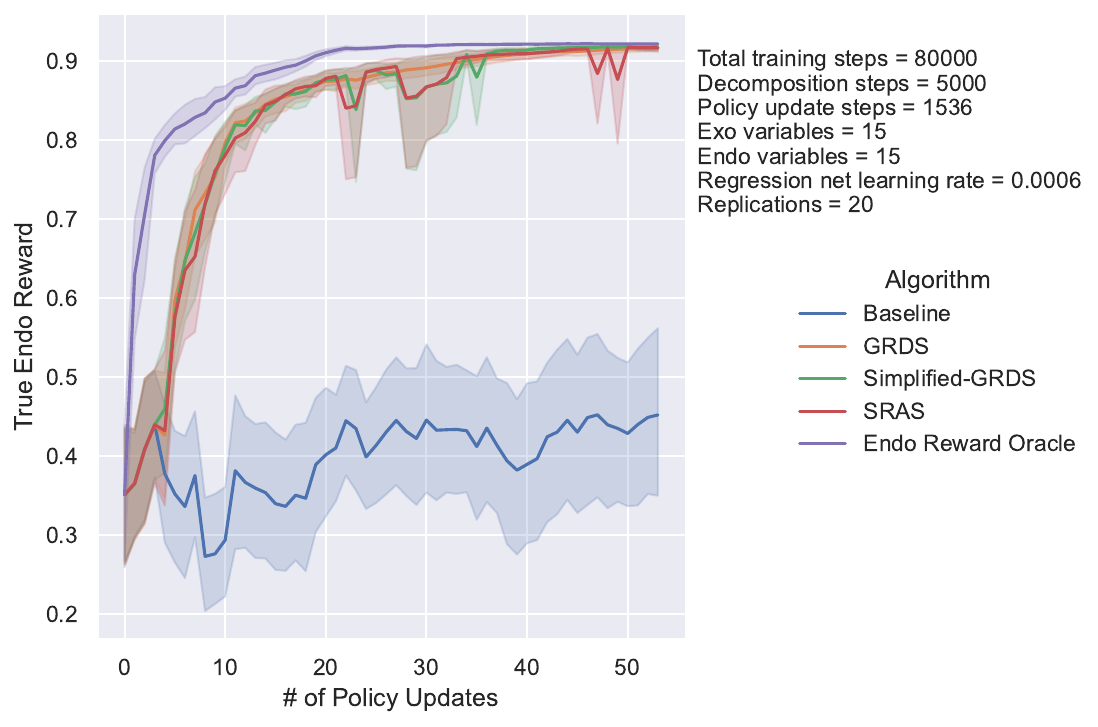}
         \caption{$R_{exo,t}^4$.}
         \label{fig:nonlinear_5}
     \end{subfigure}
    \caption{RL performance for MDPs with nonlinear exo reward functions.}
    \label{fig:nonlinear_rewards}
\end{figure}

To address RQ3, we now study the performance of our methods when they are applied to MDPs that depart in various ways from the linear dynamical MDPs studied thus far:
\begin{itemize}
\item[(a)] Rewards are nonlinear functions of the state, 
\item[(b)] Transition dynamics are nonlinear,
\item[(c)] The action space is combinatorial, and
\item[(d)] The states and actions are discrete.
\end{itemize}

Because we are introducing various non-linearities, we investigated alternative methods for performing the reward regression (details in Appendix~\ref{app:impact-exo-regression}). Based on our experiments, we adopted the following online neural network reward regression procedure. The reward network is implemented in sklearn \citep{sklearn} with 2 hidden layers of 50 and 25 units, respectively, and ReLU activations. We train with Adam using the default Adam parameters. The learning rate is set by default to $lr_{regr}=0.0003$ and the L2 regularization to $3\times 10^{-5}$. The batch size is set to 256. In Phase 1 of Algorithm~\ref{alg:framework}, we train the net with the $L$ collected samples until convergence (or a maximum number of 125 epochs). During Phase 2, we perform online learning by updating the neural net every $M=256$ training steps with a single pass over the last 256 samples. 

\subsubsection{Nonlinear Exogenous Reward Functions}
Because we have adopted online neural network reward regression, we expect that our methods should be able to fit nonlinear exogenous reward functions.  We consider the high-dimensional linear setting of Section \ref{sec:linear-setting} with $m=n=15$. We perform exo/endo decomposition after $L=5000$ steps and train for a total of $N=80000$ steps. We found it beneficial to use a learning rate for the exogenous reward regression of 0.0006 instead of the default 0.0003; the higher learning rate can help the exogenous reward neural net adapt faster. We perform 20 replications with different seeds. Furthermore, we replace the linear exogenous reward function $R_{exo,t}$ by the following four choices:
\begin{itemize}
\vspace{-0.2cm}\item $R_{exo,t}^1=\textrm{clip}(6\cdot(\textrm{avg}(x_t) + \frac{1}{3}\cdot\textrm{avg}(x_t^2) - \frac{2}{15}\cdot\textrm{avg}(x_t^3)), -5.0, 5.0)$, a $3^{rd}$ degree polynomial.
\vspace{-0.2cm}\item $R_{exo,t}^2=-3 \cdot e^{-|\textrm{avg}(x_t)|^{1.5}}$, a function of $\textrm{avg}(x_t)$ with a single mode.
\vspace{-0.2cm}\item $R_{exo,t}^3=-3 \cdot (e^{-|\textrm{avg}(x_t + 1.5)|^{2}} - e^{-|\textrm{avg}(x_t - 1.5)|^{2}})$, a function of $\textrm{avg}(x_t)$ with two modes.
\vspace{-0.2cm}\item $R_{exo,t}^4=-3 \cdot (e^{-|\textrm{avg}(x_t + 1)|^{2}} + \frac{3}{2}\cdot e^{-|\textrm{avg}(x_t - 1.5)|^{2}} - \frac{5}{3}e^{-|\textrm{avg}(x_t)|^{2}})$, a function of $\textrm{avg}(x_t)$ with three modes.
\end{itemize}
Figures \ref{fig:nonlinear_0}-\ref{fig:nonlinear_5} plot the results. 

We generally observe that all methods match the Endo Reward Oracle's performance and outperform the baseline by a large margin. This confirms that the nonlinear reward regression is able to fit these nonlinear reward functions. As we have observed before, the RL performance of \sras{} is a bit unstable, perhaps because it is not always able to detect the maximal exogenous subspace. The Simplified-\grds{} method also shows a tiny bit of instability. 

Table \ref{table:nonlinear_rewards} reports the average rank of the discovered exogenous subspaces. The true exogenous space has rank 29, but the exogenous reward only depends on 15 of those dimensions. The Simplified-\grds{} method is most consistently able to find the true rank, whereas \sras{} and \grds{} struggle to capture that last dimension. 

\begin{table}[t!]
{\footnotesize
\begin{center}
\begin{tabular}{ c c c c c } 
 \hline
 Exo Reward & \multirow{2}{*}{Method} & Exo Subspace & Total Time & Decomposition \\ 
 Function & & Rank & (secs) & Time (secs) \\ 
 \hline\hline
 \multirow{4}{*}{$R_{exo,t}^1$} & Baseline & - & 1520.0$\pm$153.0 & - \\
   & \grds{} & 28.05$\pm$1.02 & 1874.6$\pm$357.3 & 155.8$\pm$122.1 \\ 
   & Simplified-\grds{} & 29.0$\pm$0.0 & 1808.6$\pm$278.7 & 40.5$\pm$11.3 \\ 
   & \sras{} & 27.95$\pm$1.56 & 1914.4$\pm$510.3 & 251.8$\pm$249.5 \\ 
 \hline
 \multirow{4}{*}{$R_{exo,t}^2$} & Baseline & - & 1510.9$\pm$136.2 & - \\
   & \grds{} & 28.3$\pm$0.9 & 1860.4$\pm$359.9 & 150.3$\pm$155.1 \\
   & Simplified-\grds{} & 29.0$\pm$0.0 & 1797.8$\pm$264.4 & 31.9$\pm$7.7 \\ 
   & \sras{} & 27.95$\pm$1.56 & 1925.4$\pm$516.3 & 297.7$\pm$244.1 \\ 

 \hline
 \multirow{4}{*}{$R_{exo,t}^3$} & Baseline & - & 1518.1$\pm$146.4 & - \\
   & \grds{} & 28.0$\pm$1.0 & 1879.6$\pm$340.0 & 150.3$\pm$118.1 \\
   & Simplified-\grds{} & 29.0$\pm$0.0 & 1818.0$\pm$262.2 & 28.6$\pm$8.5 \\ 
   & \sras{} & 27.95$\pm$1.56 & 1932.4$\pm$505.4 & 272.9$\pm$265.5 \\ 
 \hline
 \multirow{4}{*}{$R_{exo,t}^4$} & Baseline & - & 1513.8$\pm$103.7 & - \\
   & \grds{} & 28.35$\pm$0.79 & 1849.8$\pm$366.1 & 123.2$\pm$111.5 \\
   & Simplified-\grds{} & 28.35$\pm$0.79 & 1812.0$\pm$268.9 & 37.1$\pm$11.3 \\ 
   & \sras{} & 29.0$\pm$0.0 & 1926.7$\pm$560.5 & 307.4$\pm$320.7 \\ 
 \hline
\end{tabular}
\end{center}
\caption{\label{table:nonlinear_rewards}Average and standard deviation for rank of discovered exo subspace, total execution time, and decomposition time for the MDPs with nonlinear exo rewards.}}
\end{table}

\subsubsection{Nonlinear State Transition Dynamics}

So far, we have considered MDPs with linear state transitions for the endogenous and exogenous states. It is a natural question whether our algorithms can handle more general MDPs. In this section, we provide experimental results on a more general class of nonlinear MDPs. Even though we lack a rigorous theoretical understanding, our results hint at the potential of the CCC objective to discover useful exo/endo state decompositions even when the dynamics are nonlinear.

In the experiments in this section, we introduce nonlinear dynamics, but we still configure the exogenous and endogenous state spaces so that they are linear projections of the full state space.  We study the following three MDPs, which are defined according to the recipe in Section \ref{sec:linear-setting} with the following modifications:
\begin{itemize}
\vspace{-0.2cm} \item $\mathcal{M}_1$ is a 10-D MDP with $m=n=5$ and a single action variable. The exo and endo state transitions are
\begin{equation*}
\begin{split}
&x_{t+1} = \textrm{clip}(M_{exo}\cdot x_{t} + \frac{1}{3}\cdot N_{exo}\cdot x^2_{t} - \frac{2}{15}\cdot K_{exo}\cdot x^3_{t}, -4, 4) + \varepsilon_{exo}\\
&e_{t+1} = M_{end}\cdot \begin{bmatrix}e_{t}\\ x_{t}\end{bmatrix} + M_a\cdot a_t + \varepsilon_{end},
\end{split}
\end{equation*}
where $M_{exo},m_e,M_a$ and $\varepsilon_{exo},\varepsilon_{end}$ are defined as in Section \ref{sec:linear-setting}. Furthermore, the two matrices $N_{exo}\in\mathbb{R}^{n,n}$ and $K_{exo}\in\mathbb{R}^{n,n}$ are generated following the same procedure as $M_{exo}$. $\mathcal{M}_1$ has nonlinear exogenous dynamics but linear endogenous dynamics.
\vspace{-0.2cm} \item $\mathcal{M}_2$ is exactly the same as $\mathcal{M}_1$ except that the endogenous transition function now has a nonlinear dependence on the action:
\begin{equation*}
\begin{split}
&e_{t+1} = M_{end}\cdot \begin{bmatrix}e_{t}\\ x_{t}\end{bmatrix} + M_a\cdot a_t + N_a\cdot a_t^2 + \varepsilon_{end}.
\end{split}
\end{equation*}
The entries in $N_a\in\mathbb{R}^{m}$ are sampled from the uniform distribution over $[0.5, 1.5)$.
\vspace{-0.2cm} \item $\mathcal{M}_3$ is a 10-D MDP with $m=n=5$ and a single action variable. The exogenous and endogenous state transitions functions are
\begin{equation*}
\begin{split}
&x_{t+1} = \textrm{clip}(5\cdot \textrm{sign}(x_{t})\cdot \sqrt{|x_t|} - \sin(x_t), -2, 2) + \varepsilon_{exo}\\
&e_{t+1} = M_{end}\cdot \begin{bmatrix}e_{t}\\ x_{t}\end{bmatrix} + \sin(3\cdot a_t) + \varepsilon_{end}.
\end{split}
\end{equation*}
The entries in the noise vectors $\varepsilon_{exo}$ and $\varepsilon_{end}$ are sampled from $\mathcal{N}(0,0.16)$ and  $\mathcal{N}(0,0.09)$, respectively. Like $\mathcal{M}_2$, $\mathcal{M}_3$'s  exo and endo transition functions are both nonlinear.
\end{itemize}

\begin{figure}[t!]
     \centering
     \begin{subfigure}[t]{0.45\textwidth}
         \includegraphics[scale=0.4]{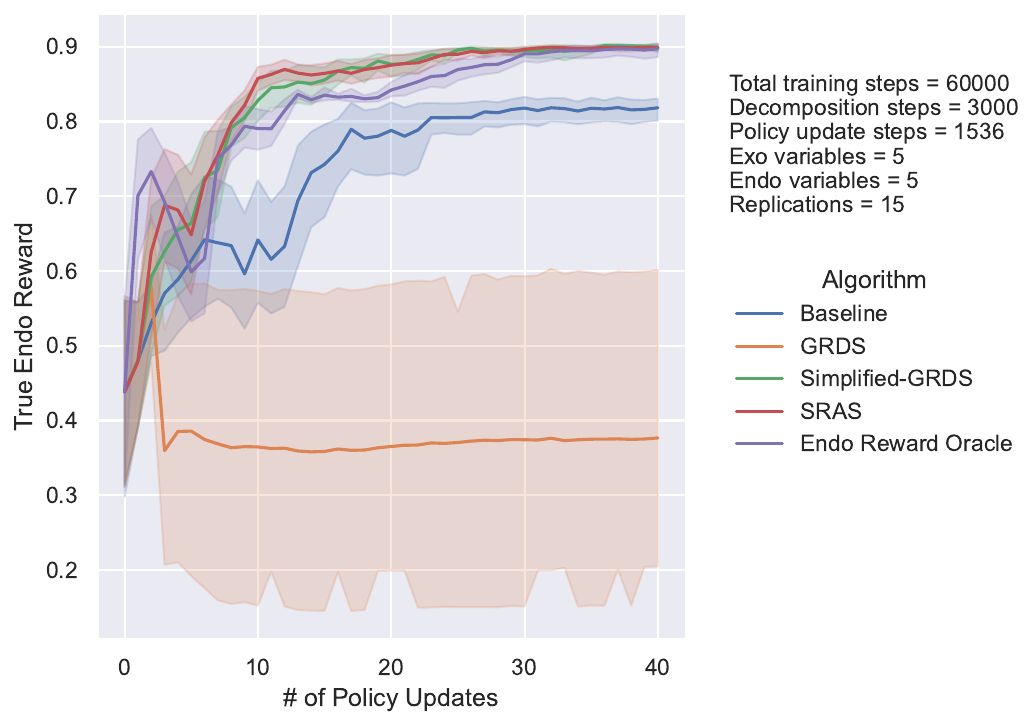}
         \caption{$\mathcal{M}_1$.}
         \label{fig:M1}
     \end{subfigure}
     \hspace{0.0\textwidth}
     \begin{subfigure}[t]{0.45\textwidth}
         \includegraphics[scale=0.4]{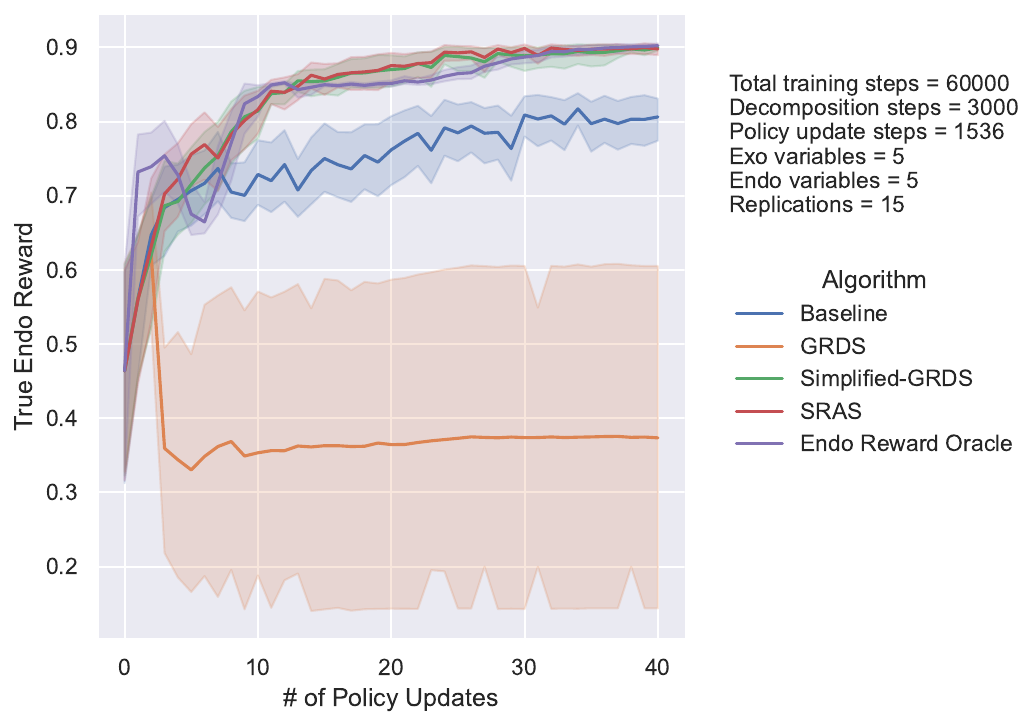}
         \caption{$\mathcal{M}_2$.}
         \label{fig:M2}
     \end{subfigure}

     \bigskip
     \begin{subfigure}[t]{0.45\textwidth}
         \includegraphics[scale=0.4]{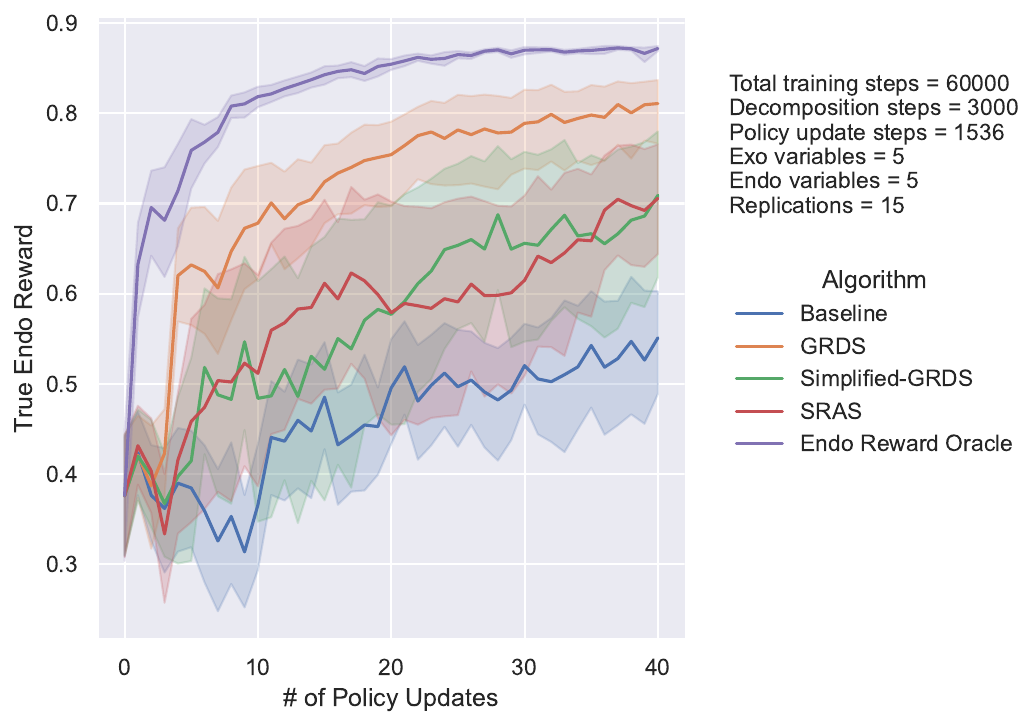}
         \caption{$\mathcal{M}_3$.}
         \label{fig:M3a}
     \end{subfigure}
     \hspace{0.0\textwidth}
     \begin{subfigure}[t]{0.45\textwidth}
         \includegraphics[scale=0.4]{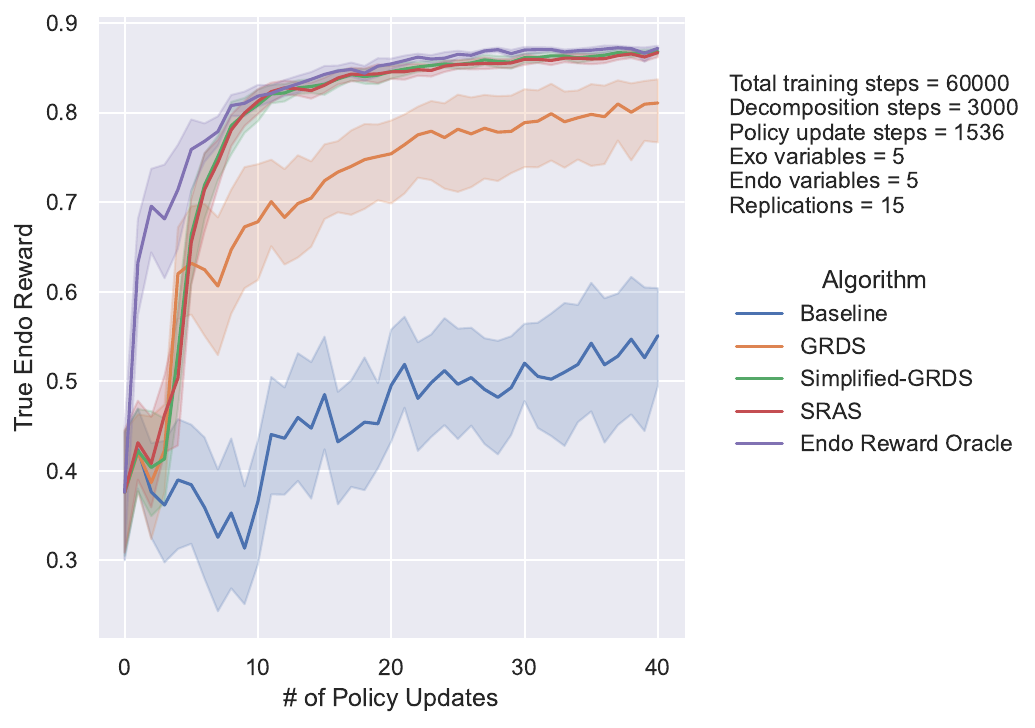}
         \caption{$\mathcal{M}_3$ $(\epsilon=0.1)$.}
         \label{fig:M3b}
     \end{subfigure}
    \caption{RL performance for MDPs with nonlinear state transitions.}
    \label{fig:nonlinear_states}
\end{figure}

Figures \ref{fig:M1}-\ref{fig:M3a} plot the RL performance over 15 replications with different seeds and Table~\ref{table:nonlinear_states} reports the ranks of the discovered exogenous subspaces. Consider first MDPs $\mathcal{M}_1$ and $\mathcal{M}_2$. The Simplified-\grds{} and \sras{} algorithms perform very well. They converge quickly, and they are able to match the Endo Reward Oracle's performance. In both settings, the Baseline converges to a suboptimal value and suffers from high variance. Most shocklingly, \grds{} performs catastrophically and exhibits very high variance even though it discovers an exogenous subspace of the correct rank. 

Now consider $\mathcal{M}_3$ in Figure \ref{fig:M3a}. On this problem, Simplified-\grds{} and \sras{} perform poorly (although still better than the baseline), while \grds{} performs much better. However, none of the methods is able to match the Endo Reward Oracle. Table~\ref{table:nonlinear_states} reveals that all of the methods, but particularly Simplified-\grds{} and \sras{}, are failing to discover the correct exogenous subspace. This suggests that the CCC computation when applied to the simplified objective is not finding good solutions. To evaluate this possibility, we reran the $\mathcal{M}_3$ experiment with a larger value of $\epsilon=0.1$ instead of its default value of $0.05$.  With this change, the results improve dramatically for Simplified-\grds{} and \sras{}, and they are able to match the performance of the Endo Reward Oracle. However, the performance of \grds{} does not improve, which suggests that it is not able to find good solutions to the large manifold optimization problems that it is solving. This could be because the CCC objective is confused by the nonlinear dynamics or it could be that the optimization is trapped in local minima. 

\begin{table}[t!]
{\footnotesize
\begin{center}
\begin{tabular}{ c c c c c c } 
 \hline
 \multirow{2}{*}{MDP} & Exo/Endo State & \multirow{2}{*}{Method} & Exo Subspace & Total Time & Decomposition \\ 
 & Variables & & Rank & (secs) & Time (secs) \\ 
 \hline\hline
 \multirow{4}{*}{$\mathcal{M}_1$} & \multirow{4}{*}{5/5} & Baseline & - & 665.3$\pm$39.9 & - \\
   &  & \grds{} & 9.0$\pm$0.0 & 890.4$\pm$140.6 & 55.0$\pm$34.0 \\ 
   &  & Simplified-\grds{} & 9.0$\pm$0.0 & 906.7$\pm$154.6 & 72.2$\pm$19.8 \\ 
   &  & \sras{} & 9.0$\pm$0.0 & 951.2$\pm$253.2 & 172.8$\pm$73.2 \\ 
 \hline
 \multirow{4}{*}{$\mathcal{M}_2$} & \multirow{4}{*}{5/5} & Baseline & - & 717.0$\pm$30.3 & - \\
   &  & \grds{} & 9.0$\pm$0.0 & 959.5$\pm$137.1 & 61.9$\pm$41.2 \\ 
   &  & Simplified-\grds{} & 9.0$\pm$0.0 & 981.2$\pm$138.0 & 85.0$\pm$18.1 \\ 
   &  & \sras{} & 9.0$\pm$0.0 & 1019.8$\pm$271.9 & 187.4$\pm$72.6 \\ 
 \hline
 \multirow{4}{*}{\begin{tabular}{l}$\quad\mathcal{M}_3$\\$(\epsilon=0.05)$\end{tabular}} & \multirow{4}{*}{5/5} & Baseline & - & 768.2$\pm$50.1 & - \\
   &  & \grds{} & 7.67$\pm$1.44 & 1033.0$\pm$241.2 & 125.8$\pm$126.7 \\ 
   &  & Simplified-\grds{} & 3.60$\pm$3.28 & 988.1$\pm$160.4 & 38.7$\pm$19.0 \\ 
   &  & \sras{} & 3.47$\pm$3.36 & 1081.3$\pm$370.7 & 168.8$\pm$70.7 \\ 
 \hline
 \multirow{4}{*}{\begin{tabular}{l}$\quad\mathcal{M}_3$\\$(\epsilon=0.1)$\end{tabular}} & \multirow{4}{*}{5/5} & Baseline & - & 768.2$\pm$50.1 & - \\
   &  & \grds{} & 7.67$\pm$1.44 & 1033.0$\pm$241.2 & 125.8$\pm$126.7 \\ 
   &  & Simplified-\grds{} & 9.0$\pm$0.0 & 918.3$\pm$129.7 & 17.7$\pm$4.6 \\ 
   &  & \sras{} & 9.0$\pm$0.0 & 1023.3$\pm$341.5 & 230.4$\pm$69.7 \\ 
 \hline
\end{tabular}
\end{center}
\caption{\label{table:nonlinear_states}Average and standard deviation for rank of discovered exo subspace, total execution time, and decomposition time for the MDPs with nonlinear state transitions.}}
\end{table}

\subsubsection{Combinatorial Action Spaces}

A limitation of our experiments so far has been that the action space is one-dimensional. We have seen that this implies that the maximal exogenous subspace has dimension $n+m-1$ rather than $n$ as originally intended. In this section, we describe experiments where we introduce higher-dimensional action spaces. For example, with a 5-dimensional action space, the policy must now select one of 10 values for each of the 5 action variables. In effect, the MDP now has $10^5$ primitive actions.  

To design MDPs with high-dimensional action spaces, we modify the general linear setting of Section \ref{sec:linear-setting}, so that the action $a_t\in\mathbb{R}^l$ is an $l$-D vector, and the matrix $M_a$ multiplying $a_t$ is in $\mathbb{R}^{m,l}$. As in the single-action setting, each action variable takes 10 possible values evenly-spaced in $[-1,1]$. We consider 6 MDPs with a variety of different structures, which may in principle appear in real applications:
\begin{itemize}
\vspace{-0.2cm}\item 10-D MDP with $m=n=5$ and $l=5$. The action matrix $M_a$ is \textit{dense}, meaning that all its entries are nonzero. We sample the entries in $M_a$ from the uniform distribution over $[0,1)$ and subsequently normalize each row of $M_a$ to sum to 0.99 for stability. We apply the decomposition algorithms after $L=6000$ steps and train for a total of $N=200000$ steps.
\vspace{-0.2cm}\item 20-D MDP with $m=n=10$ and $l=10$. The action matrix $M_a$ is dense, and generated as above. We set $L=10000$ and $N=200000$.
\vspace{-0.2cm}\item 30-D MDP with $m=n=15$ and $l=8$. The action matrix $M_a$ is \textit{partial dense}, meaning that only $l=8$ out of the $m=15$ endogenous states are controlled, but these 8 states are controlled by all actions. $M_a$ is generated as above, except that the rows corresponding to non-controlled endo variables are 0. We set $L=15000$ and $N=200000$.
\vspace{-0.2cm}\item 30-D MDP with $m=n=15$ and $l=8$. The action matrix $M_a$ is \textit{partial disjoint}, meaning that only $l=8$ out of the $m=15$ endo states are controlled, but each of these 8 states is controlled by a distinct action variable. We sample the $l=8$ nonzero entries of $M_a$ from the uniform distribution over $[0.5, 1.5)$. We set $L=15000$ and $N=200000$.
\vspace{-0.2cm}\item 35-D MDP with $m=20$, $n=15$ and $l=10$. The action matrix $M_a$ is partial disjoint, that is, only 10 endogenous state variables are directly controlled through the 10 actions (each by a distinct action variable), whereas the remaining ones are controlled indirectly through the other endogenous states; furthermore, the endo transition matrix $M_e$ is \textit{sparse} with sparsity (fraction of nonzeros) 14.3\%. Specifically, $M_e$ is generated as in Section \ref{sec:linear-setting} except that only a small part of the matrix (equal to 14.3\%) is initialized to nonzero values. We set $L=20000$ and $N=200000$.
\vspace{-0.2cm}\item 70-D MDP with $m=40$, $n=30$ and $l=20$. The action matrix $M_a$ is partial disjoint, i.e., only 20 endogenous state variables are directly controlled through the 20 actions whereas the remaining ones are controlled indirectly through the other endogenous states. The endo transition matrix $M_e$ is sparse with sparsity 14.3\%. We set $L=35000$ and $N=300000$.
\end{itemize}

\begin{figure}[t!]
     \centering
     \begin{subfigure}[t]{0.46\textwidth}
         \includegraphics[scale=0.38]{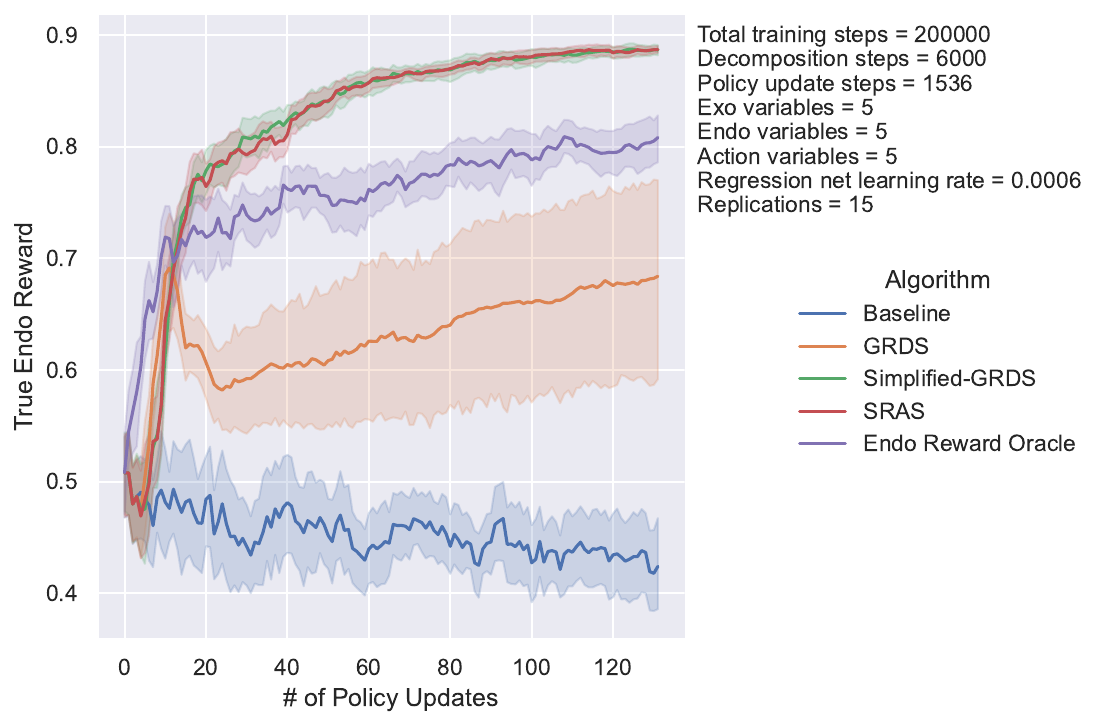}
         \caption{Dense 10-D MDP.}
         \label{fig:5x5x5_full}
     \end{subfigure}
     \hspace{0.0\textwidth}
     \begin{subfigure}[t]{0.44\textwidth}
         \includegraphics[scale=0.38]{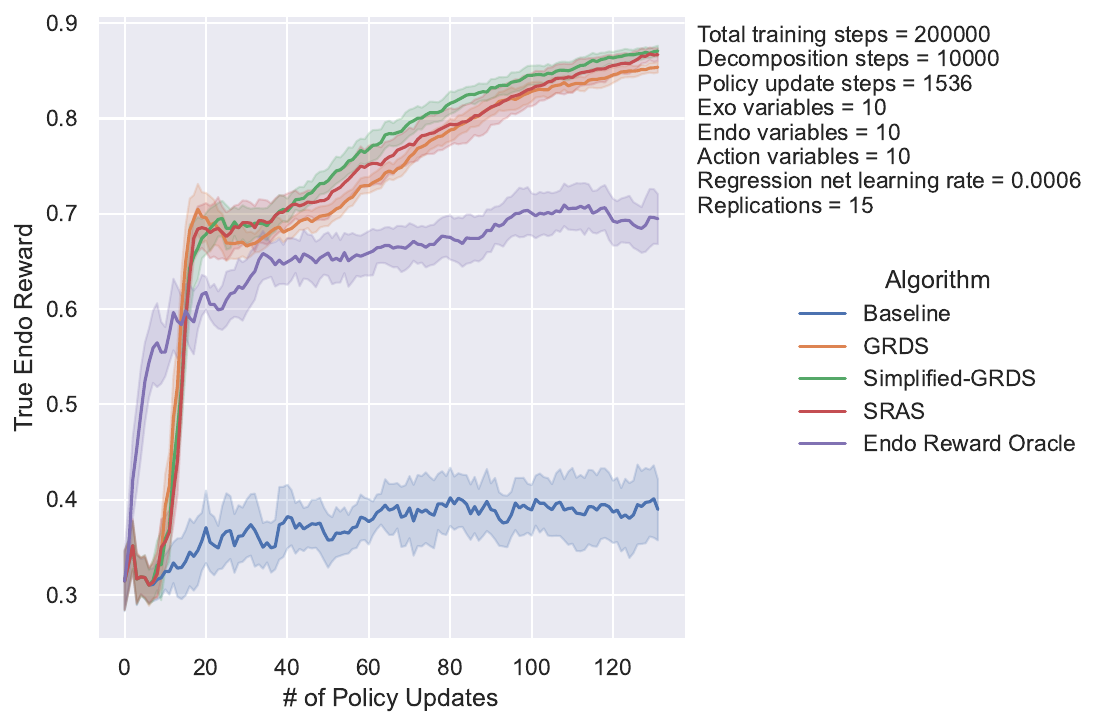}
         \caption{Dense 20-D MDP.}
         \label{fig:10x10x10_full}
     \end{subfigure}

     \bigskip
     \begin{subfigure}[t]{0.45\textwidth}
         \includegraphics[scale=0.38]{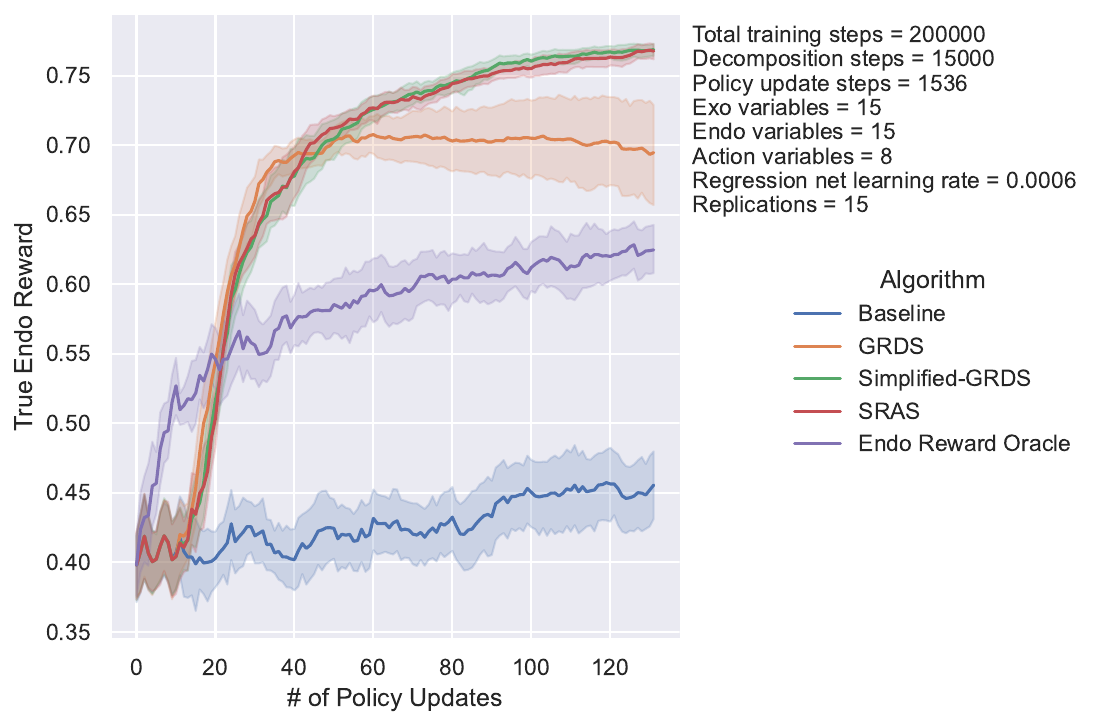}
         \caption{Partial dense 30-D MDP.}
         \label{fig:15x15x8_partial_full}
     \end{subfigure}
     \hspace{0.0\textwidth}
     \begin{subfigure}[t]{0.45\textwidth}
         \includegraphics[scale=0.38]{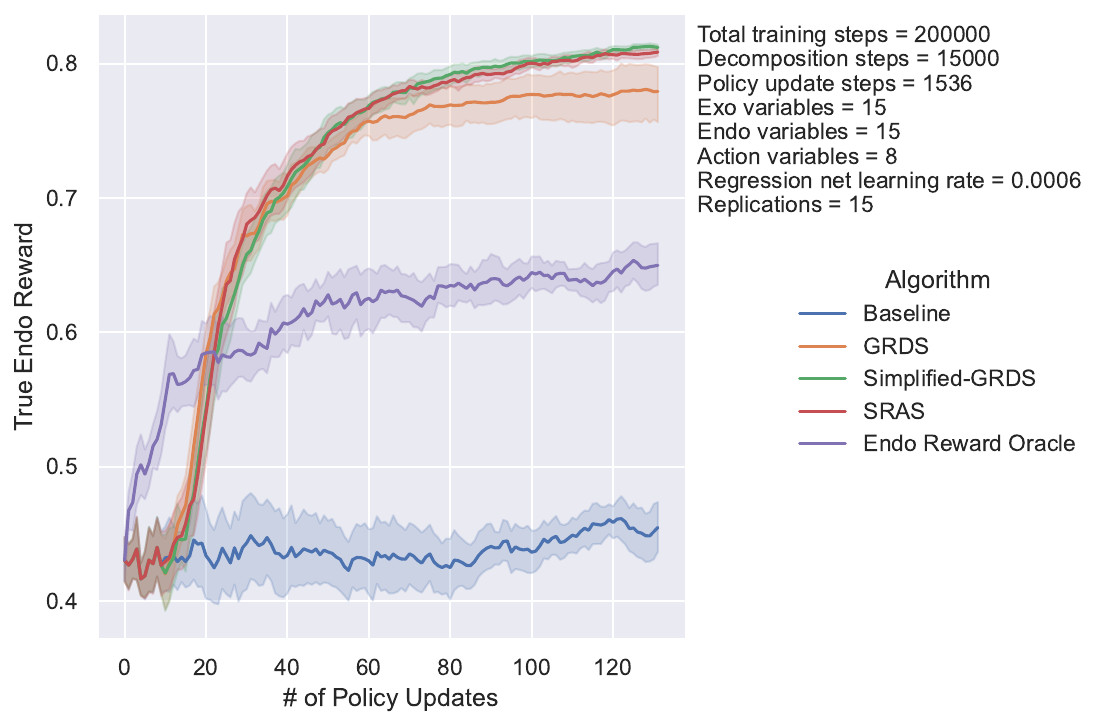}
         \caption{Partial disjoint 30-D MDP.}
         \label{fig:15x15x8_partial_disjoint}
     \end{subfigure}
     
     \bigskip
     \begin{subfigure}[t]{0.46\textwidth}
         \includegraphics[scale=0.38]{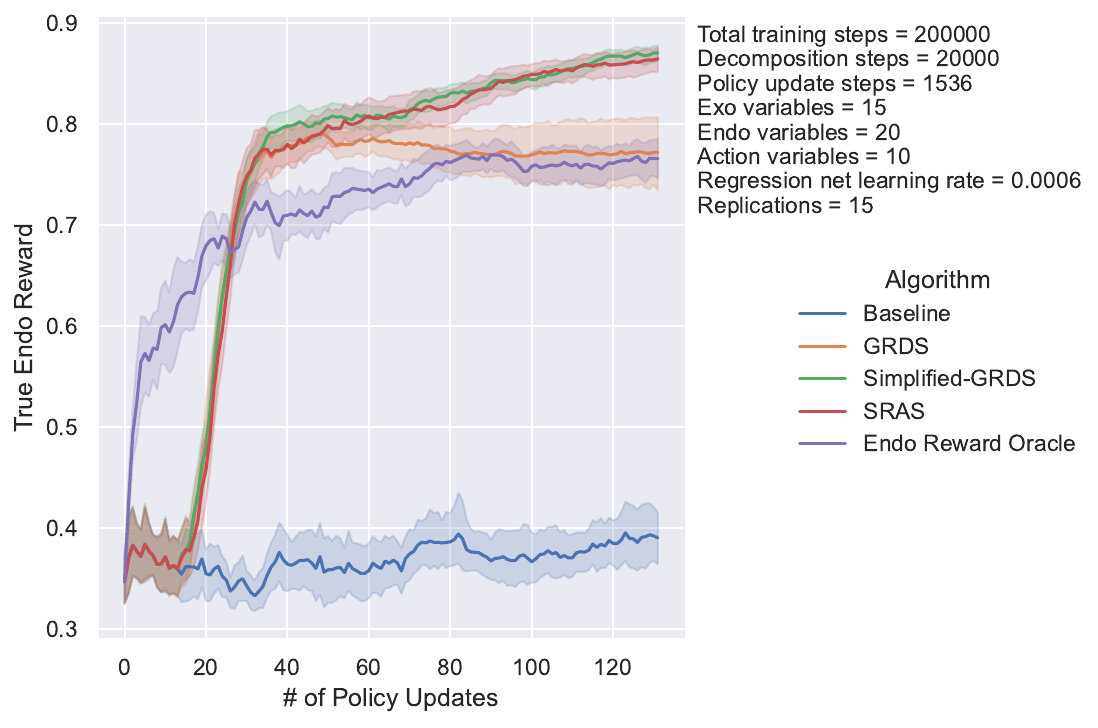}
         \caption{Partial disjoint sparse 35-D MDP.}
         \label{fig:15x20x10_partial_disjoint}
     \end{subfigure}
     \hspace{0.0\textwidth}
     \begin{subfigure}[t]{0.44\textwidth}
         \includegraphics[scale=0.38]{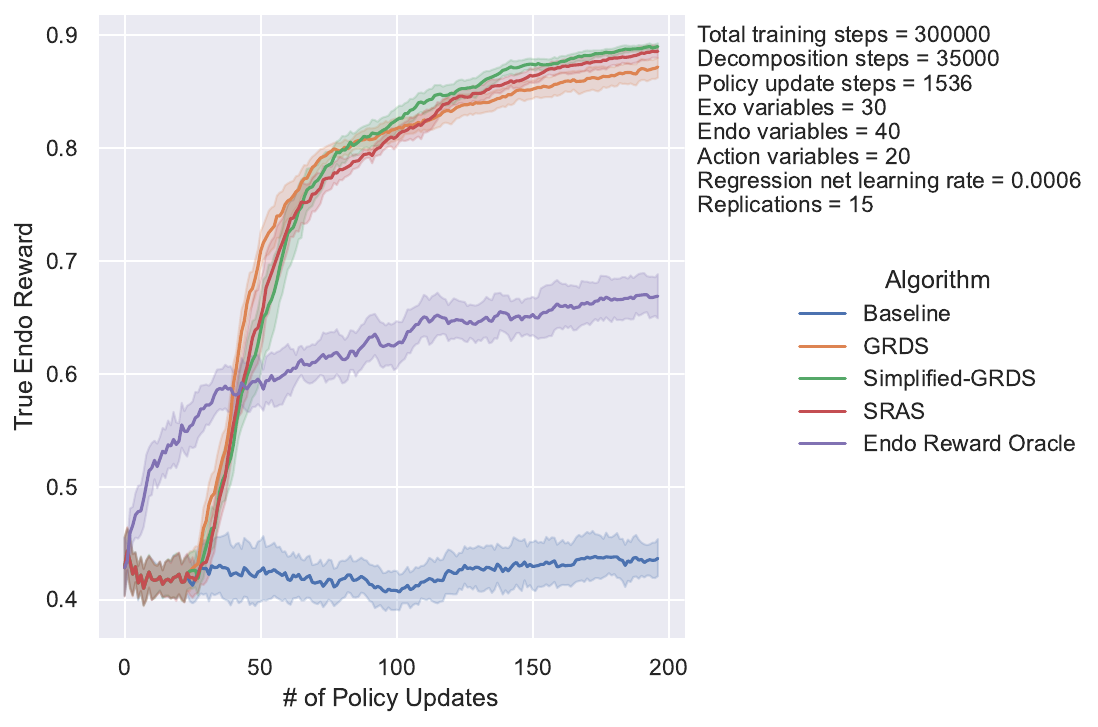}
         \caption{Partial disjoint sparse 70-D MDP.}
         \label{fig:30x40x20_partial_disjoint}
     \end{subfigure}
    \caption{RL performance for MDPs with multiple action variables.}
    \label{fig:multiple_actions}
\end{figure}
\begin{table}[t!]
{\footnotesize
\begin{center}
\begin{tabular}{ c c c c c c c } 
 \hline
 Action & Exo/Endo & \multirow{2}{*}{Method} & Exo Subspace & Total Time & Decomposition \\ 
 Variables & State Variables & & Rank & (secs) & Time (secs) \\ 
 \hline\hline
\multirow{4}{*}{\parbox{1.5cm}{5 (dense)}} & \multirow{4}{*}{5/5} & Baseline & - & 3576.8$\pm$293.2 & - \\
   &  & \grds{} & 7.47$\pm$0.50 & 4835.2$\pm$960.6 & 69.8$\pm$39.1 \\ 
   &  & Simplified-\grds{} & 7.47$\pm$0.50 & 4864.5$\pm$979.3 & 148.1$\pm$70.2 \\ 
   &  & \sras{} & 7.27$\pm$0.44 & 4914.6$\pm$1126.0 & 402.8$\pm$134.0 \\ 
 \hline
 \multirow{4}{*}{\parbox{1.5cm}{10 (dense)}} & \multirow{4}{*}{10/10}  & Baseline & - & 6162.7$\pm$478.0 & - \\
   &  & \grds{} & 15.6$\pm$0.49 & 7745.5$\pm$1299.4 & 396.5$\pm$228.8 \\ 
   &  & Simplified-\grds{} & 16.0$\pm$0.0 & 7731.9$\pm$1292.3 & 387.0$\pm$145.0 \\ 
   &  & \sras{} & 16.0$\pm$0.0 & 8064.7$\pm$2006.7 & 1546.7$\pm$483.1 \\ 
 \hline
 \multirow{4}{*}{\begin{tabular}{l}8 (partial\\ $\quad$dense)\end{tabular}} & \multirow{4}{*}{15/15} & Baseline & - & 8734.0$\pm$878.1 & - \\
   &  & \grds{} & 27.0$\pm$0.0 & 10400.0$\pm$1599.6 & 397.0$\pm$244.3 \\ 
   &  & Simplified-\grds{} & 27.0$\pm$0.0 & 10150.5$\pm$1452.2 & 556.7$\pm$121.8 \\ 
   &  & \sras{} & 26.93$\pm$0.25 & 11448.9$\pm$4052.9 & 3702.0$\pm$1730.1 \\ 
 \hline
 \multirow{4}{*}{\begin{tabular}{l}8 (partial\\ $\quad$disjoint)\end{tabular}} & \multirow{4}{*}{15/15} & Baseline & - & 8388.9$\pm$706.1 & - \\
   &  & \grds{} & 22.0$\pm$0.0 & 10399.5$\pm$2200.5 & 1136.4$\pm$433.9 \\ 
   &  & Simplified-\grds{} & 22.0$\pm$0.0 & 10362.2$\pm$1687.5 & 1520.7$\pm$304.4 \\ 
   &  & \sras{} & 22.0$\pm$0.0 & 10389.7$\pm$1877.7 & 1224.7$\pm$506.5 \\ 
 \hline
 \multirow{4}{*}{\begin{tabular}{l}10 (partial\\ $\quad$disjoint\\ $\quad$sparse)\end{tabular}} & \multirow{4}{*}{15/20} & Baseline & - & 12088.9$\pm$973.6 & - \\
   &  & \grds{} & 23.2$\pm$0.4 & 16126.3$\pm$1096.9 & 2799.9$\pm$473.5 \\ 
   &  & Simplified-\grds{} & 5.47$\pm$1.89 & 18627.7$\pm$3103.6 & 7019.8$\pm$1427.1 \\ 
   &  & \sras{} & 6.2$\pm$1.51 & 13984.1$\pm$6876.4 & 1976.3$\pm$1504.9 \\ 
 \hline
 \multirow{4}{*}{\begin{tabular}{l}20 (partial\\ $\quad$disjoint\\ $\quad$sparse)\end{tabular}} & \multirow{4}{*}{30/40} & Baseline & - & 35134.1$\pm$2877.3 & - \\
   &  & \grds{} & 50.0$\pm$0.0 & 43345.1$\pm$3470.3 & 10256.2$\pm$2719.9 \\ 
   &  & Simplified-\grds{} & 50.0$\pm$0.0 & 44187.4$\pm$3672.7 & 10670.1$\pm$3015.7 \\ 
   &  & \sras{} & 49.0$\pm$0.0 & 44350.5$\pm$3540.4 & 11691.1$\pm$2925.2 \\ 
 \hline
\end{tabular}
\end{center}
\caption{\label{table:multiple_actions}Mean and standard deviation of the rank of the discovered exo subspace, total execution time, and decomposition time for MDPs with multiple action variables.}}
\end{table}

Figures \ref{fig:5x5x5_full}-\ref{fig:30x40x20_partial_disjoint} plot RL performance for these 6 MDPs. We run 15 replications with different seeds and a reward regression learning rate of 0.0006. A first observation is that in all 6 settings the baseline struggles and shows very slow improvement over time time (e.g., Figure \ref{fig:10x10x10_full}). Its performance appears to decay on the smallest of these MDPs (Figure \ref{fig:5x5x5_full}). The Endo Reward Oracle performs visibly better than the Baseline and is able to attain higher rewards. However, it exhibits high variance and it improves very slowly (e.g., Figure \ref{fig:10x10x10_full}).

Surprisingly, the Simplified-\grds{} and \sras{} methods substantially outperform the Endo Reward Oracle. It appears that our methods are able to discover additional exogenous dimensions that reduce the variance of the endogenous reward below the level of the reward oracle. Table~\ref{table:multiple_actions} confirms that, with the exception of the 35-dimensional sparse, partial disjoint MDP, the algorithms are all discovering exogenous spaces of the expected size. For example, in the largest MDP, which has 70 dimensions and a 20-dimensional action space, \grds{} and Simplified-\grds{} both discover a 50-dimensional exogenous space. The algorithms are challenged by the fourth MDP ($n$=15, $m=20$, $l=10$). On this MDP, \grds{} finds a 23.2-dimensional exo space on average, but Simplified-\grds{} and \sras{} only find exo spaces with an average dimension of 5.47 and 6.2, respectively. They also exhibit high variation in the number of discovered dimensions. Despite these failures, Simplified-\grds{} and \sras{} perform very well on all six of these MDPs. \grds{} struggles on the dense MDPs, but does quite well on the sparse and partial disjoint MDPs. 

\subsubsection{Discrete MDPs}

\begin{figure}[b!]
     \centering
         \includegraphics[scale=0.8]{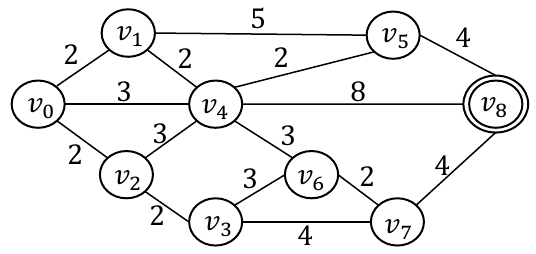}
    \caption{Graph for discrete MDP.}
    \label{fig:discrete-MDP}
\end{figure}

The final variation in MDP structure that we studied was to create an MDP with discrete states. Figure \ref{fig:discrete-MDP} shows a simple routing problem defined on a road network.  There are 9 endogenous states corresponding to the nodes of the network. This MDP is episodic; each episode starts in the starting node, $v_0$, and ends in the terminal node $v_8$. Each edge in the network has a corresponding traversal cost, and the goal of the agent is to reach the terminal node while minimizing the total cost. 
There are 4 exogenous state variables; each of them is independent of the others and evolves as $x_{t+1,i} = 0.9\cdot x_{t,i} + \varepsilon_{exo}$, where $\varepsilon_{exo}$ is distributed according to $\mathcal{N}(0,1)$ and $i\in\{1,2,3,4\}$. These exogenous state variables are intended to model global phenomena such as amount of automobile traffic, fog, snow, and pedestrian traffic. These quantities evolve independently of the navigation decisions of the agent and can be ignored by the optimal policy.

The reward function is the sum of two terms:
\[
r_t = -\textrm{cost}(s_t \rightarrow s_{t+1}) - \sum_{i=1}^4 x_{t,i}.\]
The first term is the endogenous reward $R_{end,t}$ and the second term is the exogenous reward $R_{exo,t}$.

The actions at each node consist in choosing one of the outbound edges to traverse. We restrict the set of actions to move only rightward (i.e., toward states with higher subscripts). For instance, there are three available actions at node $v_0$ corresponding to the three outgoing edges, but only a single action at node $v_4$. The cost of traversing an edge is shown by the edge weights in Figure ~\ref{fig:discrete-MDP}. The MDP is deterministic: a given action (i.e., edge selection) at a given node always results in the same transition. The observed state consists of the 1-hot encoding for the 9 endo states plus the 4-D continuous exo state variables.

We apply episodic RL with PPO. Since the MDP is small, we modify some of the hyperparameters as follows. We set the total training steps to $N=10000$ and the policy update steps to $K=128$. We perform decomposition after $L=300$ episodes instead of the default value of 3,000 steps. The PPO batch size is set to 32. Every time we update the policy, we execute 300 evaluation episodes (in a separate instance of the MDP environment). For reward regression, we compute a Single Linear Regression rather than Online Neural Net Regression (see Section \ref{app:impact-exo-regression}). We perform 15 replications, each with a different random seed.

\begin{figure}[t!]
     \centering
     \begin{subfigure}[t]{0.45\textwidth}
         \includegraphics[scale=0.4]{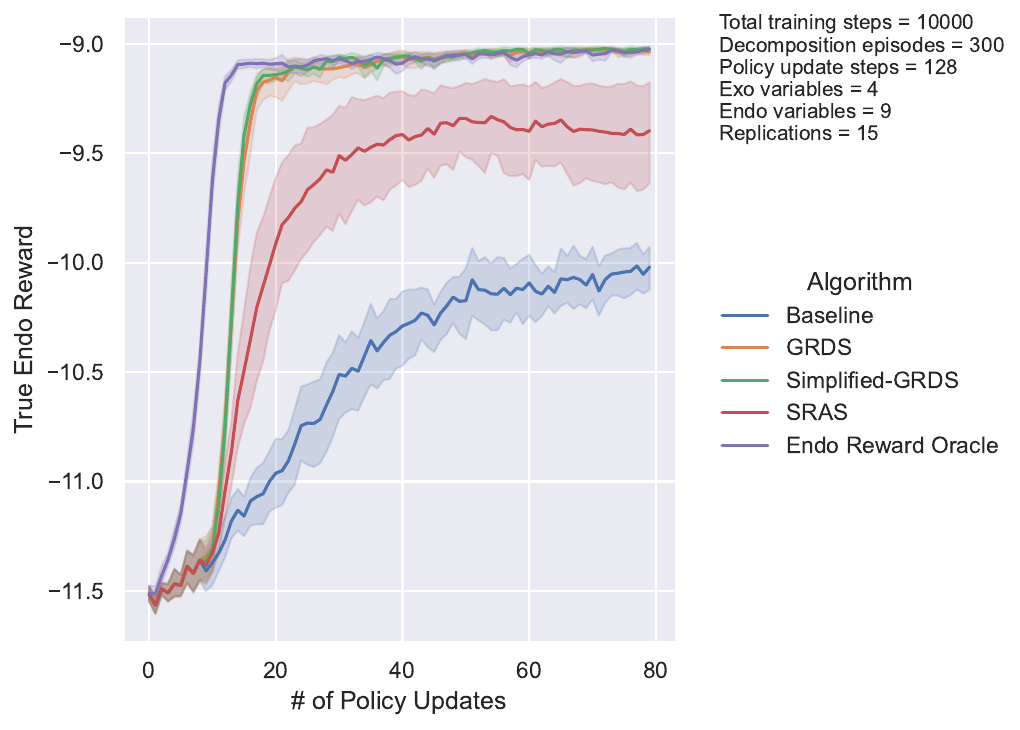}
         \caption{Deterministic MDP.}
         \label{fig:deterministic}
     \end{subfigure}
     \hspace{0.0\textwidth}
     \begin{subfigure}[t]{0.45\textwidth}
         \includegraphics[scale=0.4]{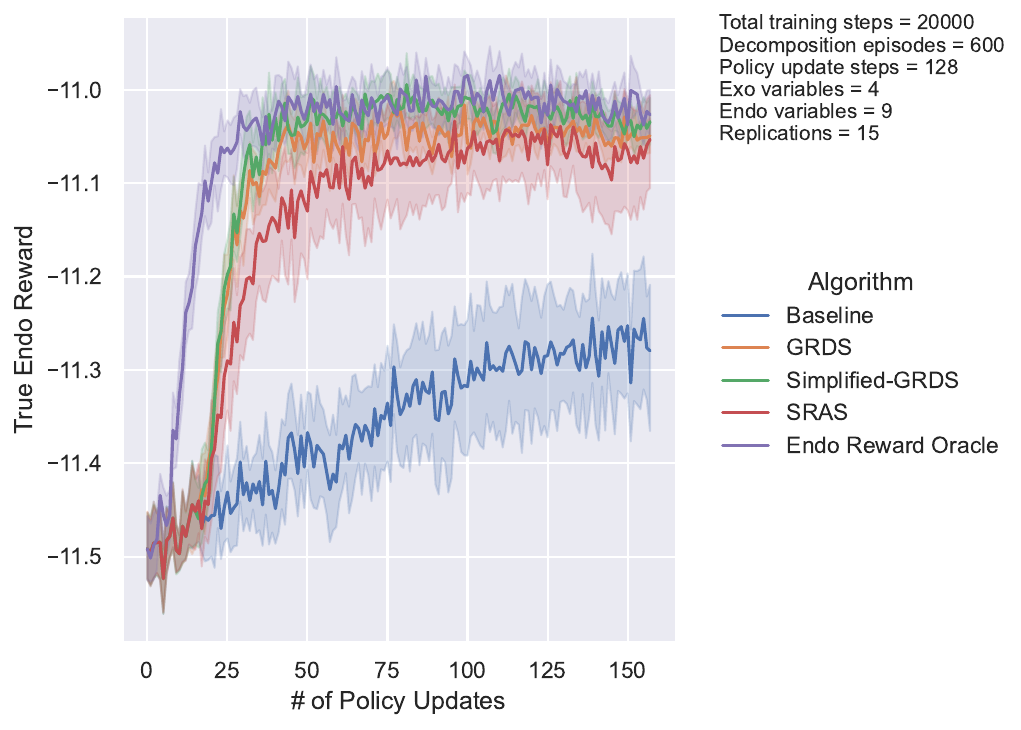}
         \caption{Non-deterministic MDP.}
         \label{fig:non-deterministic}
     \end{subfigure}
    \caption{RL performance for discrete MDPs.}
    \label{fig:discrete-plots}
\end{figure}

Figure \ref{fig:deterministic} plots RL performance. Note that our methods perform on par with the Endo Reward Oracle and exhibit very little variance, with the exception of \sras{}. On the other hand, the Baseline makes very slow progress.  Table \ref{table:discrete-MDP} reports the rank of the discovered exogenous subspace as well as the total time and decomposition time. \grds{} and Simplified-\grds{} discover large exogenous subspaces of rank 11 and 10, respectively. The computed exogenous subspace includes the 4 exogenous state variables we defined as well as linear combinations of the endogenous states that do not depend or have only weak dependence on the action. In contrast, \sras{} discovers an exo subspace of very low rank, which explains its suboptimal performance. 

The strong performance of our methods might be due to the deterministic dynamics of the problem. To test this, we created a stochastic version of the road network MDP. The transition probabilities are specified as follows:
\begin{itemize}
\item Taking action 0 / 1 / 2 at node $v_0$ leads to nodes $v_1, v_2, v_4$ with probabilities $(0.5, 0.3, 0.2)$ /$(0.3, 0.5, 0.2)$ / $(0.3, 0.2, 0.5)$, respectively.
\item Taking action 0 / 1 at node $v_1$ leads to nodes $v_4, v_5$ with probabilities $(0.6, 0.4)$ / $(0.5, 0.5)$, respectively.
\item Taking action 0 / 1 at node $v_2$ leads to nodes $v_3, v_4$ with probabilities $(0.5, 0.5)$ / $(0.3, 0.7)$, respectively.
\item Taking action 0 / 1 at node $v_3$ leads to nodes $v_6, v_7$ with probabilities $(0.7, 0.3)$ / $(0.4, 0.6)$, respectively.
\item Taking action 0 / 1 / 2 at node $v_4$ leads to nodes $v_5, v_6, v_8$ with probabilities $(0.6, 0.2, 0.2)$ /$(0, 1, 0)$ / $(0.3, 0.2, 0.5)$, respectively.
\item There is only one action (action 0) available at nodes $v_5, v_6, v_7$, and this leads with probability 1 to nodes $v_8, v_7, v_8$, respectively.
\item Node $v_8$ remains a terminal node.
\end{itemize}

\begin{table}
{\footnotesize
\begin{center}
\begin{tabular}{ c c c c c c } 
 \hline
 \multirow{2}{*}{MDP} & Exo/Endo State & \multirow{2}{*}{Method} & Exo Subspace & Total Time & Decomposition \\ 
 & Variables & & Rank & (secs) & Time (secs) \\ 
 \hline\hline
 \multirow{4}{*}{Deterministic} & \multirow{4}{*}{4/9} & Baseline & - & 225.6$\pm$16.7 & - \\
   &  & \grds{} & 9.47$\pm$0.50 & 271.4$\pm$55.8 & 68.4$\pm$26.9 \\
   &  & Simplified-\grds{} & 11.0$\pm$0.0 & 229.3$\pm$18.0 & 10.2$\pm$1.9 \\ 
   &  & \sras{} & 2.13$\pm$1.63 & 272.9$\pm$39.2 & 72.3$\pm$13.5 \\ 
 \hline
 \multirow{4}{*}{Non-deterministic} & \multirow{4}{*}{4/9} & Baseline & - & 419.9$\pm$10.8 & - \\
   &  & \grds{} & 9.66$\pm$0.70 & 503.6$\pm$102.2 & 109.1$\pm$42.7 \\
   &  & Simplified-\grds{} & 11.0$\pm$0.0 & 446.6$\pm$21.5 & 15.6$\pm$2.1 \\ 
   &  & \sras{} & 2.8$\pm$1.38 & 488.0$\pm$56.0 & 82.7$\pm$9.7 \\ 
 \hline
\end{tabular}
\end{center}
\caption{\label{table:discrete-MDP}Average and standard deviation for rank of discovered exo subspace, total execution time, and decomposition time for the discrete MDPs.}}
\end{table}

We employ the same hyperparameters as in the deterministic setting, except that we set the total training steps $N:=20000$ and the number of decomposition episodes $L:=600$. 

Figure \ref{fig:non-deterministic} plots the RL performance over 15 trials (each with a separate random seed). We observe that Simplified-\grds{} and \grds{} perform only slightly worse than the Endo Reward Oracle, while \sras{} exhibits slightly lower average performance and higher variance. The Baseline improves very slowly and lags far behind the other methods. Due to the stochastic transitions, all methods (and particularly the Baseline) exhibit larger variance than in the deterministic MDP in Figure \ref{fig:deterministic}. 

Regarding the rank of the computed subspace, we notice from Table \ref{table:discrete-MDP} that the ranks of the discovered exo subspaces are the same as for the deterministic MDP. This demonstrates the robustness of the algorithms. Simplified-\grds{} finds the largest exogenous space with \grds{} close behind. \sras{} finds much smaller spaces, which partially explains its slightly poorer performance. It is important to remember that not all dimensions of the discovered exogenous subspaces may be relevant to reward regression. \sras{} may only be discovering 2.8 dimensions, on average, but these are enough to give it a huge performance advantage over the Baseline. 

The total CPU cost and decomposition cost are higher than the deterministic setting, which reflects the added cost of running online reward regression for twice as many steps. 

In this problem, the rank-descending methods worked better than \sras{}, but even \sras{} gave excellent results, and the total CPU time required is not significantly higher than the Baseline.

\section{Discussion}
\subsection{RQ1: Sample Complexity and CPU Time}
The experiments show that the exo-endo decomposition approach greatly reduces the amount of training data (exploration steps) and the amount of CPU time required to match or exceed the endogenous oracle. The improvements are so dramatic that we were not able to run the Baseline method long enough for it to match the level of RL performance achieved by our methods, so we estimated this by linear extrapolation.

\subsection{RQ2: Correctness of the Decomposition Methods}
Our methods surprised us by finding exogenous subspaces that were larger than we had naively expected. Analysis showed that these subspaces did indeed have maximal dimension. Our experiments measured the correctness of the subspaces by the resulting RL performance. In nearly all cases, our methods matched the performance of the Endo Reward Oracle that was told the correct subspace for the endogenous reward.

\subsection{RQ3: Performance Beyond Continuous State Spaces, Linear Dynamics, and Linear Rewards}
Our methods, combined with neural network regression, handled the nonlinear rewards well. When the transition dynamics are nonlinear, our methods sometimes struggled, but we found that they worked well if we increased the $\epsilon$ parameter to allow the CCC objective to be substantially nonzero. Our methods gave excellent results in combinatorial action spaces, but these problems required much more CPU time (for all methods) compared to one-dimensional action spaces. Our methods---especially Simplified-\grds{}---gave excellent results in discrete state MDPs.

\subsection{RQ4: Comparison of Decomposition Algorithms}
The performance of \grds{} was uneven. It failed catastrophically on MDPs $\mathcal{M}_1$ and $\mathcal{M}_2$ with non-linear dynamics, and it was often worse than Simplified-\grds{} and \sras{} on other problems. There was only one problem, $\mathcal{M}_3$, a highly nonlinear MDP, where \grds{} out-performed the other two methods. But when we increased the $\epsilon$ parameter in the CCC computation to 0.10, the other methods were able to far out-perform \grds{}. We hypothesize that the poor performance of \grds{} results from the difficulty of the large manifold optimization problems that it must solve. This is a shame, because \grds{} is the best-understood algorithm from a theoretical perspective. Perhaps advances in manifold optimization will permit it to match the performance of the other methods.

Simplified-\grds{} and \sras{} gave very similar performance across virtually all of the MDPs. The one case where Simplified-\grds{} clearly out-performed \sras{} was on the deterministic graph traversal problem. On this problem, \sras{} hit a performance ceiling while Simplified-\grds{} was able to match the endogenous oracle. Simplified-\grds{} is also somewhat faster (in CPU time) than \sras{}. 

Based on this comparison, Simplified-\grds{} appears to be the best choice. What are the risks of this exo/endo decomposition algorithm?  From a mathematical perspective, when we minimize $I(X';A \mid X)$ (or the corresponding CCC approximation), we are only eliminating edges from $A$ to $X'$. These are the direct effects of $A$. We are not excluding any indirect path by which $A$ might affect $E'$, and $E'$ might then affect $X'$ in some future time step. In the linear dynamical MDPs that we have studied, such indirect effects do not arise, and all effects of $A$ on $X'$ are visible immediately.  Of course, because Simplified-\grds{} verifies the full constraint $I(X';[E,A] \mid X)<\epsilon$, Simplified-\grds{} is still sound, but it may fail to find the maximal exogenous subspace in complex MDPs. 

\section{Limitations}
The main limitation of our approach is that it assumes that there is an additive component of the reward function that depends only on the exogenous state variables. It is unclear how often such an additive component will be present in real applications. In our prior work \citep{dietterich2018}, we reported an experiment with cell phone tower optimization where \grds{} produced a significant speedup for Q-learning. Fortunately, in real applications it is usually easy to execute Phase 1 of Algorithm~\ref{alg:framework} as part of any reinforcement learning process. One can then apply one of our decomposition algorithms to see if there are exogenous variables present and, if so, perform reward regression to see how much of the reward variance can be explained by those exogenous variables. Note that detecting and isolating the exogenous state variables is valuable even if the reward does not additively decompose, as it may possible to reduce or eliminate those variables from the MDP.

The second major limitation of our approach is that the exogenous subspace is assumed to be a linear projection of the full state space. Our algorithms further rely on the CCC rather than conditional mutual information to identify this subspace. An interesting direction for future work would be to develop neural network methods for modeling the dynamics of nonlinear systems, isolating the exogenous dynamics, and performing non-linear reward regression. Nonetheless, experience throughout engineering has shown that linear methods are surprisingly useful even when the underlying systems are nonlinear.

A third potential issue with the exogenous reward regression approach is that it models only the expected value of $R_{exo}$. If the exogenous reward contains a high degree of noise, this is not removed. Rather, it remains in the residual reward $r_i - \hat{m}_{exo}(\xi_{exo}(s_i))$. A general approach to removing noise in the reward function is to fit a regression model to predict the expected value of the immediate reward. In our setting, one could also model the endogenous reward by fitting a model to predict $r_i - \hat{m}_{exo}(\xi_{exo}(s_i))$ from $s_i$ and then replace the observed rewards with their estimated expected values. A weakness of any reward modeling approach, including ours, is that if the estimated reward is biased, this may reduce the quality of the learned policy.

Finally, our experiments were performed on MDPs where both $R_{end}$ and $R_{exo}$ are dense (i.e., nonzero almost everywhere). If the rewards were sparse, the reward regression could be more difficult, and linear regression would not be a good choice. Note that the decomposition algorithms always receive dense training data, because they work only with the observed state at each time step.

\section{Concluding Remarks}
In this paper, we proposed a causal theory of exogeneity in reinforcement learning and showed that in causal models satisfying the full two-step DBN structure, the exogenous variables are causally exogenous. 

We introduced exogenous-state MDPs with additively decomposable rewards and proved that such MDPs can be decomposed into an exogenous Markov reward process and an endogenous MDP such that any optimal policy for the endogenous MDP is an optimal policy for the original MDP. We studied the properties of valid exo/endo decompositions and proved that there is a maximal exogenous subspace that contains all other exogenous subspaces. We also showed that not all subsets of the maximal exogenous subspace define valid exo/endo decompositions. 

We developed two practical algorithms, \grds{} and \sras{}, for the case when the exogenous space is a linear projection of the full state space. Our algorithms use the conditional correlation coefficient (CCC) as a measure for conditional independence and rely on solving a manifold optimization problem. Under the assumption that the exploration policy visits all states and tries all actions infinitely often, we proved that \grds{} discovers the maximal exogenous subspace. We also introduced Simplified-\grds{}, which employs a simplified CCC objective and then checks the solution to see if it satisfies the full CCC objective. We found experimentally that Simplfied-\grds{} gave the best performance on a wide variety of MDPs. The cost of running Simplfied-\grds{} is very small, so we recommend that it be employed routinely in practical applications, as it may be able to discover and exploit exogenous variables to greatly accelerate real-world reinforcement learning. 

\section*{Author Contributions and Acknowledgments}

TD developed the initial formulation and provided some formal analysis. GT developed most of the formal results, invented the CCC, and designed the algorithms. GT also conducted all experiments and their analysis. Both authors contributed equally to writing the paper.

The authors thank Zhitang Chen for introducing us to the \verb+pymanopt+ library and for initial ideas that eventually became the CCC. The authors thank Yuvraj Sharma for performing an initial set of experiments during his internship at Oregon State University.

Part of this work was conducted while George Trimponias was employed at Amazon, Inc. Portions of this material were supported by the US Defense Advanced Research Projects Agency (DARPA) under Contract Numbers HR001119C0112 and HR001120C0022, and under cooperative agreement W911NF-22-2-0149. Any opinions, findings and conclusions or recommendations expressed in this material are those of the authors and do not necessarily reflect the views of the DARPA.

The authors thank the anonymous reviewers for their careful reading of the manuscript and their valuable questions and suggestions, which greatly improved the rigor and readability of the paper.

\bibliography{bibliography}
\newpage
\appendix

\section{Proof and Consequences of Theorem~\ref{theorem:generic-causal-exogeneity}}
\label{app:causal-exogeneity-beyond-DBNs}

In this appendix, we present the proof of Theorem~\ref{theorem:generic-causal-exogeneity} and explore its consequences. For ease of reading, we restate the theorem:\\

\noindent\textbf{Theorem~\ref{theorem:generic-causal-exogeneity}. }\textit{A state variable $S$ is \textrm{causally exogenous} if and only if it is action-disconnected.}\\

\begin{proof}
    We first prove the reverse direction: if $S$ is action-disconnected, then $S$ is causally exogenous. Let $\mathcal{G}$ be the causal graph of the MDP extending from time 0 to time $H$. 

    The plan for the proof is to apply Rule 3 of the do-calculus to remove $\textnormal{do}(A_t=a_t)$ in \eqref{eq:causal-exogeneity}. Rule 3 states that for any causal graph $\mathcal{G}$ 
    \[P(F \mid \doop(G),\doop(H),J)=P(F \mid \doop(G),J), \textrm{ if } F\independent H \mid G \cup J \textrm{ in } \tilde{\mathcal{G}},\]
    where $\tilde{\mathcal{G}}$ is the graph obtained by first deleting all edges pointing into $G$ and then deleting all arrows pointing into $H$ from nodes that are not ancestors of $J$. 
   
    For this purpose, we set $F\gets \{S_{t+1}, \ldots, S_H\}$, $G\gets \emptyset$, $H\gets A_t$, and $J\gets S_t$. Given $G$ is the empty set and $A_t$ cannot have incoming edges from any ancestor of $S_t$, the graph $\tilde{\mathcal{G}}$ will be identical to $\mathcal{G}$ except that the incoming edges to $A_t$ are deleted. To show exogeneity, it then suffices to show that 
    \begin{equation}\label{eq:S-conditional-independence}
        S_{t+1}, \ldots, S_H\independent A_t \mid S_t \textrm{ in } \tilde{\mathcal{G}} \quad \forall t.
    \end{equation}

     We will make use of the $d$-separation theory with trails \citep[see][]{pearl1988,pearl2009}. A trail is a loop-free and undirected (i.e., all edge directions are ignored) path between two nodes in the causal graph. To show that two nodes in the causal graph are conditionally independent, we must show that every trail connecting them is blocked. Consider any trail $L$ connecting node $A_t$ to node $S_{t+\tau}$ in $\tilde{\mathcal{G}}$. Since $S$ is action-disconnected, we know that there can be no directed path from $A_t$ to $S_{t+\tau}$ of the form $A_t\to\cdots\to S_{t+\tau}$. This implies that if a trail $L$ exists, it must necessarily contain a collider node $Z$ of the form $\to Z\gets$. To see why, notice that the first link in $L$ has the form $A_t\to\cdots$ connecting the action at time $t$ to some or all of state variables at time $t+1$. If all subsequent links in $L$ are of the form $\cdots\to\cdots$, then $L$ would be a directed path, which contradicts the assumption that $S$ is action-disconnected. Hence, the edge directionality along $L$ must change at some node, which implies there must be at least one collider node $Z$ in $L$. Let $Z=Z_{t'}$ be the first collider node in trail $L$, with time step index $t'$. It must be the case that $t'>t$, because $Z_{t'}$ is a descendant of $A_t$, and all descendants of $A_t$ must occur at times $\geq t+1$. Node $Z_{t'}$ is obviously neither $S_t$ nor an ancestor of $S_t$ because this would again create a directed trail. At time $t$, $Z_{t'}$ is not yet observed, and any unobserved collider node blocks all trails that pass through it. Therefore $Z_{t'}$ blocks this trail.  This establishes  \eqref{eq:S-conditional-independence}. 

    Next, we prove the forward direction: if $S$ is causally exogenous, then it is action-disconnected. We will prove this by exhibiting a pair of actions $a_1$ and $a_2$ and a parameterization of the MDP transition function such that 
\begin{equation}
    P(S_{t+\tau} \mid S_t, \textnormal{do}(A_t=a_1)) \neq P(S_{t+\tau} \mid S_t, \textnormal{do}(A_t=a_2)).
\end{equation}
    This will show that Equation~\ref{eq:causal-exogeneity} does not hold for $X=S$ in all parameterizations of the causal graph. Assume $S$ is not action-disconnected. Then by the definition of action-disconnected, there must be times $t$ and $t+\tau$, an action $A_t$, and a state variable $S_{t+\tau}$ such that there is a directed path from $A_t$ to $S_{t+\tau}$. This path does not go through $S_t$ because actions can only affect future states in an MDP. Hence, this path constitutes an active (unblocked) trail that allows information to flow. Therefore, we can choose two actions $a_1$ and $a_2$ and a parameterization of $P(S_{t+1}, \ldots, S_H\mid S_t, \textnormal{do}(A_t))$ such that
\begin{equation}
    P(S_{t + \tau} \mid S_t, \textnormal{do}(A_t=a_1)) \neq P(S_{t + \tau} \mid S_t, \textnormal{do}(A_t=a_1)).
\end{equation}
    This implies that Equation (\ref{eq:causal-exogeneity}) with $X=S$ does not hold, which contradicts our assumption that $S$ is causally exogenous.
\end{proof}

\begin{figure}[b!]
     \centering
     \begin{subfigure}[t]{0.40\textwidth}
         \includegraphics[scale=0.8]{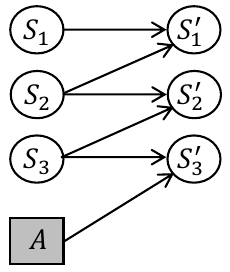}
         \caption{2-time step DBN}
         \label{fig:edge_case_MDP}
     \end{subfigure}
     \hspace{0.0\textwidth}
     \begin{subfigure}[t]{0.40\textwidth}
         \includegraphics[scale=0.8]{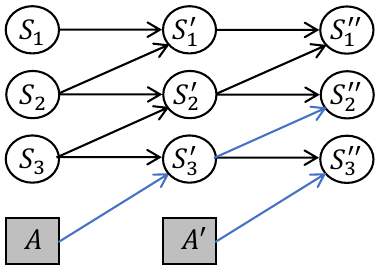}
         \caption{MDP obtained by unrolling the DBN for $H=2$.}
         \label{fig:edge_case_unrolled}
     \end{subfigure}
     \caption{A case where the conditions of Theorem~\ref{theorem:exo-DBN} are violated but there is still a valid endo/exo decomposition}
    \label{fig:edge_case}
\end{figure}

Theorem~\ref{theorem:generic-causal-exogeneity} applies to some MDPs where the DBN-based result of Theorem~\ref{theorem:exo-DBN} does not apply. Consider the causal diagram for the MDP in Figure~\ref{fig:edge_case_unrolled}. Suppose we set $X=\{S_1\}$ and $E=\{S_2,S_3\}$. Then we claim that $(E,X)$ can function as a successful endo/exo decomposition for this MDP because there is no path from either $A$ or $A'$ to any instance of $S_1$ over this short horizon. Hence, it would be safe to perform an exogenous reward regression $\hat{R}_{exo}(X)=f(S_1)$ and construct an Endo-MDP by subtracting this from the full reward function. However, $(E,X)$ violates the definition of a valid exo/endo decomposition (Definition~\ref{def:valid-exo/endo-decomposition}) because in the 2-time step DBN, there is an edge from $E$ to $X$, namely, from $S_2$ to $S'_1$. Hence, the probability distribution $P(E',X' \mid E, X, A)$ does not factor as required. This demonstrates that while the 2-time step structural condition is sufficient, it is not necessary. 

Recall that Theorem~\ref{theorem:MDP-decomposition} relied on the definition of a valid exo/endo decomposition to show that the Bellman optimality equation could be decomposed. Hence, we need another way to show that decompositions such as $(E,X)$ can give correct results. This is provided by the following theorem:

\begin{theorem}[Causally Exogenous Policy Optimization]\label{theorem:causal-exo}
    Let $\mathcal{M}$ be an MDP with state variables $S$, starting state $S_0$, and discount factor $\gamma \in [0,1]$. Let $X$ be a subset of the state variables that are all causally exogenous, and let $E = S \setminus X$ be the remaining variables. Suppose the reward function $R$ can be decomposed into the sum of two terms: $R(S) = R_{exo}(X) + R_{end}(E,X)$. Then the optimal policy can be found by maximizing the endogenous value function
    \begin{equation}
        \pi^* = \argmax_\pi\; \mathbb{E}_\pi\left[\sum_{t=0}^{\infty} \gamma^t R_{end}(E_t,X_t)\right],
    \end{equation}
    where the expectation is taken over the stochastic dynamics of the MDP, the stochasticity (if any) in the policy $\pi$, and any stochasticity in the reward $R_{end}$. 
\end{theorem}

\begin{proof}
    The optimal policy maximizes the expected return
    \[\pi^* = \argmax_\pi\; \mathbb{E}_\pi\left[\sum_{t=0}^{\infty} \gamma^t R(E_t,X_t)\right].\]
    Replacing $R$ by the sum of $R_{exo}$ and $R_{end}$ and distributing the expectation over the sum gives
    \[\pi^* = \argmax_\pi\; \mathbb{E}_\pi\left[\sum_{t=0}^{\infty} \gamma^t R_{exo}(X_t)\right] + \mathbb{E}_\pi\left[\sum_{t=0}^{\infty} \gamma^t R_{end}(E_t,X_t)\right].\]
    Because $X$ is causally exogenous, $X$ is action-disconnected. Hence, $R_{exo}(X)$ is also action-disconnected, and its expected value is therefore independent of the policy $\pi$.  This gives us
    \[\pi^* = \argmax_\pi\; \mathbb{E}\left[\sum_{t=0}^{\infty} \gamma^t R_{exo}(X_t)\right] + \mathbb{E}_\pi\left[\sum_{t=0}^{\infty} \gamma^t R_{end}(E_t,X_t)\right].\]
    Because the first term is independent of $\pi$, it does not affect the optimization. Hence, 
    \[
    \pi^* = \argmax_\pi\; \mathbb{E}_\pi\left[\sum_{t=0}^{\infty} \gamma^t R_{end}(E_t,X_t)\right].
    \]
\end{proof}

We can apply Theorem~\ref{theorem:causal-exo} to the MDP in Figure~\ref{fig:edge_case_unrolled}. Let $X=\{S_1\}$ and $E=\{S_2,S_3\}$. By inspection we can see that $S_1$ is action-disconnected. If the reward function for this MDP decomposes as $R_{exo}(S_1) + R_{end}(S_1,S_2,S_3)$, then we can obtain the optimal policy by maximizing the expected cumulative reward of $R_{end}$. Hence, even though the DBN for this MDP violates Definitions \ref{def:2-exogeneous-state-MDP} and \ref{def:valid-exo/endo-decomposition}, Algorithm~\ref{alg:framework} can be applied to find the optimal policy.

An advantage of sets of causally-exogenous variables is that any subset $U$ of a set $X$ of causally exogenous variables is itself causally exogenous, because every variable $X$ is action-disconnected. The disadvantage of causal exogeneity is that in general, the MDP must be unrolled to horizon $H$ to verify that a set of variables is causally exogenous. Hence, causal exogeneity cannot be determined by a 2-time step optimization. 

%%%%%%%%%%%%%%%%%%%%%%%%%%%%%%%%%%%%%%%%%%%%%%%%%%%%%%%%%%%%%%%%%%%%%%%%%%%

\section{Conditions Establishing Soundness of Conditional Mutual Information for Discovering Exogenous Subspaces}\label{app:soundness}

What conditions must hold so that the solution to Equation~\ref{opt:linear-decoupled-formulation}, when applied to the collected data $D$, will find a valid exo/endo decomposition? In this section, we address this question for ``tabular'' MDPs---that is, MDPs with finite, discrete state and action spaces. To find valid exo/endo decompositions, we need to ensure that the factorization of Equation \eqref{eqn:full-factorization}
holds in all states of the MDP. This means that our exploration policy needs to visit all states, and it needs to execute all possible actions in each state to verify that the action does not affect (either directly or indirectly) the exogenous variables. We formalize this as follows.

Consider an idealized version of Algorithm~\ref{alg:framework} that collects the tuple dataset $D$ by executing a fixed exploration policy $\pi_x$ for a large number of steps. We will require that $\pi_x$ is fully randomized, according to Definition~\ref{def:fully-randomized}. We will also require that the structure of the MDP is such that a fully-randomized policy will visit every state $s \in \mathcal{S}$ infinitely often. Such MDPs are said to be admissible (see Definition~\ref{def:admissible-mdp}).

Examples of admissible MDPs include episodic MDPs and ergodic MDPs. An episodic MDP begins each episode in a fixed start state $s_0$ and executes a policy until a terminal state is reached. Then it resets to the starting state. It must satisfy the requirement that all policies will reach a terminal state in a finite number of steps. The simplest episodic MDP always terminates after a fixed number of steps $H$, which is called the horizon time of the MDP.  Note that if an episodic MDP contains states that are not reachable from the start state $s_0$ by \textit{any} policy, then these must be deleted from the MDP in order to satisfy the definition of admissibility.

An ergodic MDP has the property that for all policies, every state is reachable from every other state in a finite number of steps, and the time between successive visits to any given state is aperiodic. In the case of ergodic MDPs, an equivalent definition for an admissible policy is that a fully-randomized policy will visit every state in the MDP infinitely often \citep{Puterman1994}.

\begin{theorem}[Asymptotic Soundness of Empirical Conditional Mutual Information]\label{theorem:ecmi}
Let $\mathcal{M}$ be an admissible MDP, and let $D$ be a set of $\langle s,a,r,s'\rangle$ tuples collected by executing fully-randomized policy $\pi_x$ for $n$ steps and recording the state, action, reward, and result state at each step.  Let $(X,E)$ be a proposed exo/endo decomposition of $S$. Let $\hat{P}(E,X,A,X')$ be the maximum-likelihood estimate of the joint distribution of the decomposed $(S,A,S')$ triples, and let $\hat{I}(X';E,A \mid X)$ be the corresponding estimate of the conditional mutual information. (We ignore $E'$, as it is not involved in the conditional mutual information constraint.)
Then if $\lim_{n\rightarrow \infty} \hat{I}(X';E,A \mid X) = 0$, it follows that $(X,E)$ is a valid exo/endo decomposition of $S$.
\end{theorem}
\begin{proof}
For simplicity, we focus on discrete-state, discrete-action MDPs.
We must show that $\hat{I}(X';E,A \mid X) = 0$ implies that the MDP dynamics factors as
\[
P(E',X'\mid E,X,A) = P(E'\mid E,X,A,X') P(X'\mid X).
\]
The empirical conditional mutual information is defined as
\[
\hat{I}(X'; E, A \mid X) = \sum_{e\in E,x\in X,a\in A,x'\in X'} \hat{P}(e,x,a,x') \left[\log\frac{\hat{P}(x',e,a\mid x)}{\hat{P}(x' \mid x)\hat{P}(e,a \mid x)}\right].
\]
Claim: If $\hat{I}(X'; E, A \mid X) = 0$, then 
\begin{equation}
    \hat{P}(X',E,A \mid X) = \hat{P}(X'\mid X) \hat{P}(E,A \mid X) \label{eqn:factor-eax}.
\end{equation}
There are two cases to consider. 
\begin{description}
    \item[Case 1: $\hat{P}(e,x,a,x') > 0$.] In this case, the log expression,
    \[
    \log\frac{\hat{P}(x',e,a\mid x)}{\hat{P}(x' \mid x)\hat{P}(e,a \mid x)},
    \]
    must be zero. This can only occur if the argument of the log is 1. Therefore, the numerator must equal the denominator, and Equation \eqref{eqn:factor-eax} holds.
    \item[Case 2: $\hat{P}(e,x,a,x') = 0$.] In this case, $\hat{P}(x',e,a\mid x)$ is also trivially 0. 
    However, notice that $\hat{P}(x',e,a\mid x)=\hat{P}(e,a\mid x)-\sum_{\tilde{x}\neq x',\hat{P}(\tilde{x},e,a\mid x)>0}\hat{P}(\tilde{x},e,a\mid x)=\hat{P}(e,a\mid x)-\sum_{\tilde{x}\neq x',\hat{P}(\tilde{x},e,a\mid x)>0}\hat{P}(\tilde{x} \mid x)\hat{P}(e,a \mid x)$. Hence, we have that 
    $$\hat{P}(e,a\mid x) \cdot \Big(1 - \sum_{\tilde{x}\neq x',\hat{P}(\tilde{x},e,a\mid x)>0}\hat{P}(\tilde{x} \mid x)\Big) = 0.$$
    From the above equation we then get that either $\hat{P}(e,a\mid x)=0$ or $\hat{P}(x' \mid x)=0$. Equation \eqref{eqn:factor-eax} then follows trivially.
\end{description}
By applying the chain rule of probability, we can rewrite the left-hand side of (\ref{eqn:factor-eax})  as
\[\hat{P}(X',E,A \mid X) = \hat{P}(X'\mid E,X,A)\hat{P}(A\mid E,X)\hat{P}(E\mid X).\]
Similarly, we can rewrite the right-hand side of (\ref{eqn:factor-eax}) as
\[\hat{P}(X'\mid X)\hat{P}(E,A\mid X) = \hat{P}(X'\mid X)\hat{P}(A\mid E,X)\hat{P}(E\mid X).\]
Substituting these into (\ref{eqn:factor-eax}) gives
\begin{equation}\hat{P}(X'\mid E,X,A)\hat{P}(A\mid E,X)\hat{P}(E\mid X) = \hat{P}(X'\mid X)\hat{P}(A\mid E,X)\hat{P}(E\mid X). \label{eqn:factor-eax-sub}
\end{equation}
We wish to cancel matching terms in (\ref{eqn:factor-eax-sub}). To do this, we must show that they are non-zero for all $A$, $E$, and $X$. Because the MDP is admissible, $P(E,X)>0$ for all $E$ and $X$. We can apply the chain rule of probability to rewrite this as $P(E\mid X) P(X) > 0$, hence $P(E\mid X)>0$ for all $E$ and $X$, because all probabilities are non-negative. Furthermore, because $\pi_x(A \mid E,X) = P(A\mid E,X)$ is fully randomized, $P(A \mid E,X)>0$ for all $A$, $E$, and $X$. Hence, for $n$ sufficiently large, 
\begin{align*}
\hat{P}(A|E,X)&> 0 \quad\forall E,X,A \\
\hat{P}(E|X)  &> 0 \quad\forall E,X.
\end{align*}
This authorizes us to cancel these terms from both sides of Equation~\eqref{eqn:factor-eax-sub} to obtain
\[\hat{P}(X'\mid E,X,A) = \hat{P}(X'\mid X).\]
As $n \rightarrow \infty$, all estimates will converge to their true values, and we will obtain
\begin{equation} \label{eqn:almost-full-factorization}
P(X'\mid E,X,A) = P(X'\mid X).
\end{equation}
Now consider the conditional distribution of a general MDP: $P(X',E'\mid E,X,A)$. Apply the chain rule of probability to write this as
\[
P(X',E'\mid E,X,A) = P(E' \mid E,X,A,X') P(X'|E,X,A)
\]
and substitute Equation~\eqref{eqn:almost-full-factorization} on the right-hand side. This gives us the desired full factorization of a 2-Exogenous State MDP. 
\[
P(E',X'\mid E,X,A) = P(E'\mid E,X,A,X') P(X'\mid X).
\]
This completes the proof that $(X,E)$ is a valid exo/endo decomposition.
\end{proof}

%%%%%%%%%%%%%%%%%%%%%%%%%%%%%%%%%%%%%%%%%%%%%%%%%%%%%%%%%%%%%%%%%%%%%%%%%%

\section{Proofs of the Linear Decomposition Lemmas}
\label{app:vector-space-decomposition-lemmas}

In this appendix, we provide the proofs of the three lemmas employed in the proof of Theorem~\ref{theorem:correctness}. \\
\\
\noindent \textbf{Lemma~\ref{lemma:linear-union}} \textit{
Let $[\mathcal{X}_1,\mathcal{E}_1]$ and $[\mathcal{X}_2, \mathcal{E}_2]$ be two full exo/endo decompositions of an MDP $\mathcal{M}$ with state space $\mathcal{S}$, where $\mathcal{X}_1 = \{W_1^{\top} s : s \in \mathcal{S}\}$ and $\mathcal{X}_2 = \{W_2^{\top} s : s \in \mathcal{S}\}$ and where $W_1^\top W_1 = \mathbb{I}_{d_1\times d_1}$ and $W_2^\top W_2 = \mathbb{I}_{d_2\times d_2}$, $1\leq d_1,d_2 \leq d$.
Let $\mathcal{X} = \mathcal{X}_1 + \mathcal{X}_2$ be the subspace formed by the sum of subspaces $\mathcal{X}_1$ and $\mathcal{X}_2$, and let $\mathcal{E}$ be its complement. It then holds that the state decomposition $S=[E,X]$ with $E \in \mathcal{E}$ and $X \in \mathcal{X}$ is a valid full exo/endo decomposition of $\mathcal{S}$.}
\begin{proof}
We wish to follow the same reasoning as in Theorem \ref{theorem:decomposition-union}. The only challenge is that the exogenous or even the endogenous subspaces from the two decompositions may share a common subspace (other than the trivial zero vector space $\mathbf{0}$). 
To address this, we can rewrite the linear spaces $\mathcal{X} = \mathcal{X}_1 + \mathcal{X}_2$ and $\mathcal{E}_1$ and $\mathcal{E}_2$ as the following direct sums
\begin{align}
\mathcal{X} &= \mathcal{X}_1 + \mathcal{X}_2 = \overline{\mathcal{X}} \oplus \hat{\mathcal{X}}_1 \oplus \hat{\mathcal{X}}_2\label{eq:direct-sum}\\
\mathcal{X}_1 &= \hat{\mathcal{X}}_1 \oplus \overline{\mathcal{X}}\label{eq:x1}\\
\mathcal{X}_2 &= \hat{\mathcal{X}}_2 \oplus \overline{\mathcal{X}}\label{eq:x2}\\
\mathcal{E}_1 &= \hat{\mathcal{E}}_1 \oplus \mathcal{E}\label{eq:e1}\\
\mathcal{E}_2 &= \hat{\mathcal{E}}_2 \oplus \mathcal{E}\label{eq:e2},
\end{align}
where
\begin{align}
\overline{\mathcal{X}} &= \mathcal{X}_1 \cap \mathcal{X}_2\label{eq:xbar} \\
\hat{\mathcal{X}}_1 &\sqsubseteq \mathcal{X}_1 \wedge \hat{\mathcal{X}}_1 \cap \mathcal{X}_2\ = \{\mathbf{0}\}  \label{eq:xhat1} \\
\hat{\mathcal{X}}_2 &\sqsubseteq \mathcal{X}_2 \wedge \hat{\mathcal{X}}_2 \cap \mathcal{X}_1\ = \{\mathbf{0}\}  \label{eq:xhat2}\\
\mathcal{E} &= \mathcal{E}_1 \cap \mathcal{E}_2\label{eq:e}\\
\hat{\mathcal{E}}_1 &\sqsubseteq \mathcal{E}_1 \wedge \hat{\mathcal{E}}_1 \cap 
\mathcal{E}_2\ = \{\mathbf{0}\}  \label{eq:ehat1} \\
\hat{\mathcal{E}}_2 &\sqsubseteq \mathcal{E}_2 \wedge \hat{\mathcal{E}}_2 \cap \mathcal{E}_1\ = \{\mathbf{0}\}  \label{eq:ehat2}.
\end{align}
Equation \eqref{eq:direct-sum} says that we can write $\mathcal{X}$ as the direct sum of three vector spaces: the intersection subspace $\overline{\mathcal{X}}$ in Equation \eqref{eq:xbar}, the subspace $\hat{\mathcal{X}}_1$ of $\mathcal{X}_1$ intersecting with $\mathcal{X}_2$ only at $\mathbf{0}$ (Equation \eqref{eq:xhat1}), and the subspace $\hat{\mathcal{X}}_2$ of $\mathcal{X}_2$ intersecting with $\mathcal{X}_1$ only at $\mathbf{0}$ (Equation \eqref{eq:xhat2}). Similarly, $\mathcal{E}_1$ can be written as the direct sum of the intersection $\mathcal{E}$ and a subspace $\hat{\mathcal{E}}_1$ of $\mathcal{E}$ that only intersects with $\mathcal{E}_2$ at $\mathbf{0}$.
Such decompositions are always possible in finite-dimensional vector spaces.

Now we can define random variables that permit us to apply the proof of Theorem \ref{theorem:decomposition-union}. Define random variables for the output spaces: $X=(\overline{X},\hat{X}_1,\hat{X}_2,E,\hat{E}_1,\hat{E}_2)),$ where $\overline{X} \in \overline{\mathcal{X}}$, $\hat{X}_1 \in \hat{\mathcal{X}}_1$, $\hat{X}_2 \in \hat{\mathcal{X}}_2$, $E \in \mathcal{E}$, $\hat{E}_1 \in \hat{\mathcal{E}}_1$, and $\hat{E}_2 \in \hat{\mathcal{E}}_2$.

Because $(X_1,E_1)=((\overline{X}, \hat{X}_1), (E, \hat{E}_1))$ is a valid decomposition, there can be no edges from $E$ or $\hat{E}_1$ or $A$ to any variable in $\overline{X}$ or $\hat{X}_1$. Similarly, because $(X_2,E_2)=((\overline{X}, \hat{X}_2), (E, \hat{E}_2))$ is a valid decomposition, there can be no edges from $E$ or $\hat{E}_2$ or $A$ to any variable in $\overline{X}$ or $\hat{X}_2$. Consequently, there can be no edges from $E$ or $A$ to any of the variables $\overline{X}$ or $\hat{X}_1$ or $\hat{X}_2$. This demonstrates that $(X,E)=((\overline{X}, \hat{X}_1, \hat{X}_2), E)$ is a valid exo/endo decomposition of the state space.
\end{proof}

\noindent \textbf{Lemma~\ref{cor:unique-maximal-subspace}} \textit{
The maximal exogenous vector subspace $\mathcal{X}_{max}$ defined by $W_{exo,max}$ is unique.
}
\begin{proof}
By contradiction. If there were 2 distinct maximal subspaces, then Lemma \ref{lemma:linear-union} would allow us to combine them to get an even larger exogenous vector subspace. This is a contradiction.
\end{proof}

\noindent \textbf{Lemma~\ref{lemma:maximal-containment}}\textit{
Let $W_{exo,max}$ define the maximal exogenous vector subspace $\mathcal{X}_{max}$, and let $W_{exo}$ define any other exogenous vector subspace $\mathcal{X}$. Then $\mathcal{X} \sqsubseteq \mathcal{X}_{max}$. 
}
\begin{proof}
By contradiction. If there were an exogenous subspace $\mathcal{X}$ not contained within $\mathcal{X}_{max}$, then by Lemma~\ref{lemma:linear-union} we could combine the two exo/endo decompositions to get an even larger exogenous subspace. This contradicts the assumption that $\mathcal{X}_{max}$ is maximal.
\end{proof}

%%%%%%%%%%%%%%%%%%%%%%%%%%%%%%%%%%%%%%%%%%%%%%%%%%%%%%%%%%%%%%%%%%%%%

\section{Comparison of Methods for Exogenous Reward Regression}\label{app:impact-exo-regression}
Recall that for the first set of experiments in Section~\ref{sec:linear-setting}, we implemented reward regression using standard linear regression on MDPs defined by controlled linear dynamical systems. Before exploring MDPs with nonlinear rewards and dynamics, we conducted a set of experiments to explore how the type and configuration of exo reward regression affects the performance of our methods. We compare three reward regression configurations. The first configuration is \textit{Single Linear Regression}, which fits a linear model for the exo reward and performs regression only once at the end of Phase 1 of Algorithm~\ref{alg:framework}. The second configuration is \textit{Repeated Linear Regression}. Like Single Linear Regression, it fits a linear model at the end of Phase 1. In addition, it re-fits the model in Phase 2 every 1,000 collected samples using all transition data so far. Note that this is different from online Algorithm \ref{alg:framework}, which updates the exogenous reward function in Phase 2 every $M$ observations using only the last $M$ observations in $D_{exo}$. The goal was to understand whether regular regression with all observed transition data can perform better than a single linear regression at the end of Phase 1. The third configuration is \textit{Online Neural Net Regression}, which fits a neural network to the exo reward data. At the end of Phase 1, we perform reward regression until convergence. During Phase 2, we then perform a single epoch every 256 steps. 

\begin{figure}[t!]
     \centering
     \begin{subfigure}[t]{0.45\textwidth}
         \includegraphics[scale=0.4]{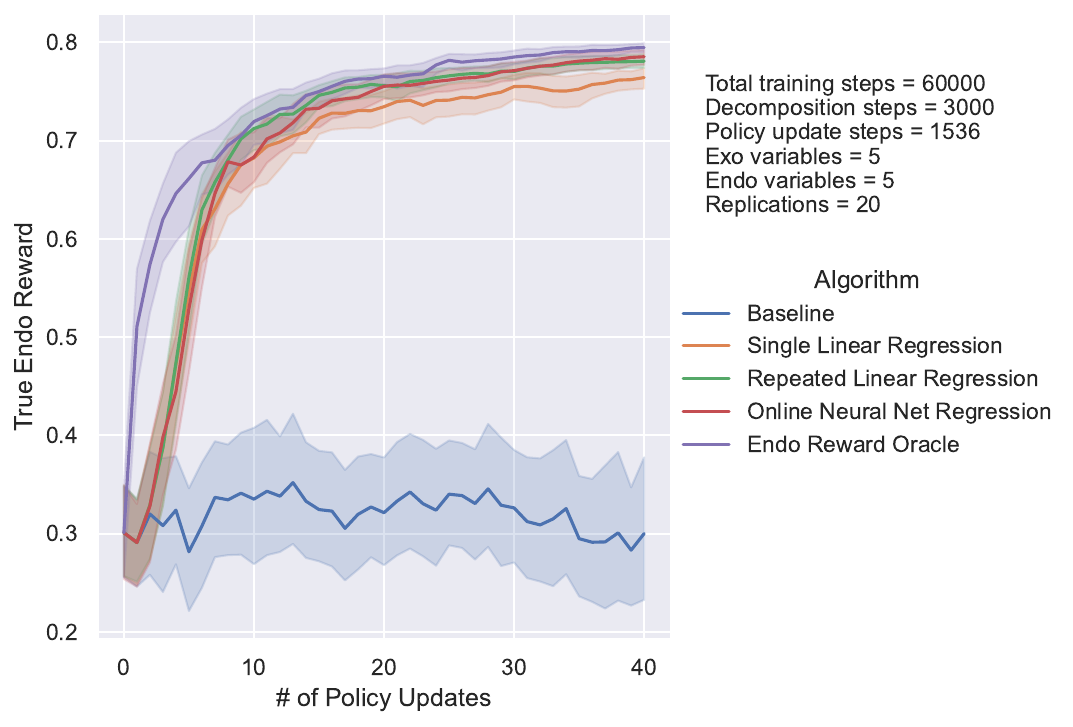}
         \caption{\grds{} ($m=5,n=5$).}
         \label{fig:5x5-global}
     \end{subfigure}
     \hspace{0.0\textwidth}
     \begin{subfigure}[t]{0.45\textwidth}
         \includegraphics[scale=0.4]{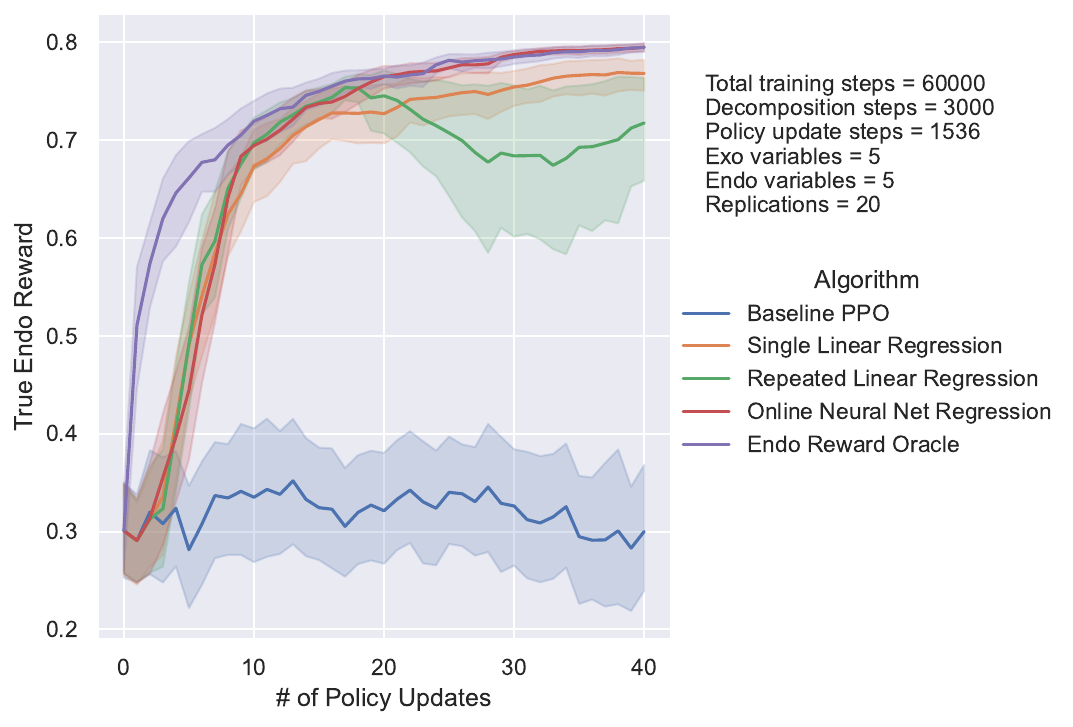}
         \caption{\sras{} ($m=5,n=5$).}
         \label{fig:5x5-stepwise}
     \end{subfigure}
     \bigskip
     \begin{subfigure}[t]{0.45\textwidth}
         \includegraphics[scale=0.4]{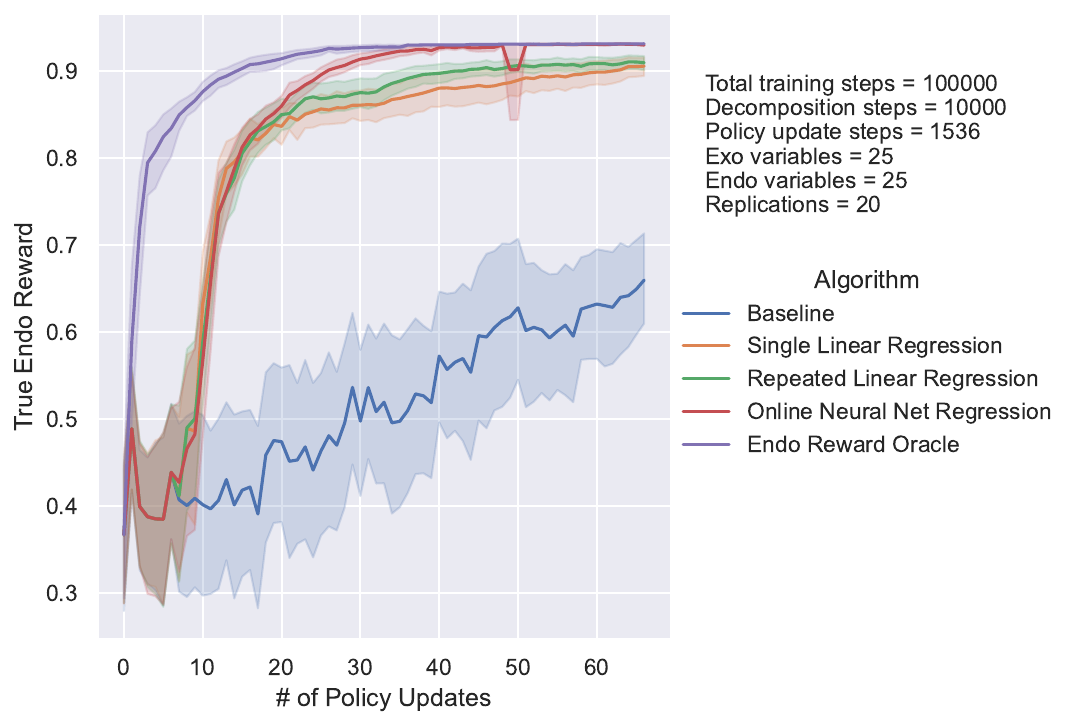}
         \caption{\grds{} ($m=25,n=25$).}
         \label{fig:25x25-global}
     \end{subfigure}
     \hspace{0.0\textwidth}
     \begin{subfigure}[t]{0.45\textwidth}
         \includegraphics[scale=0.4]{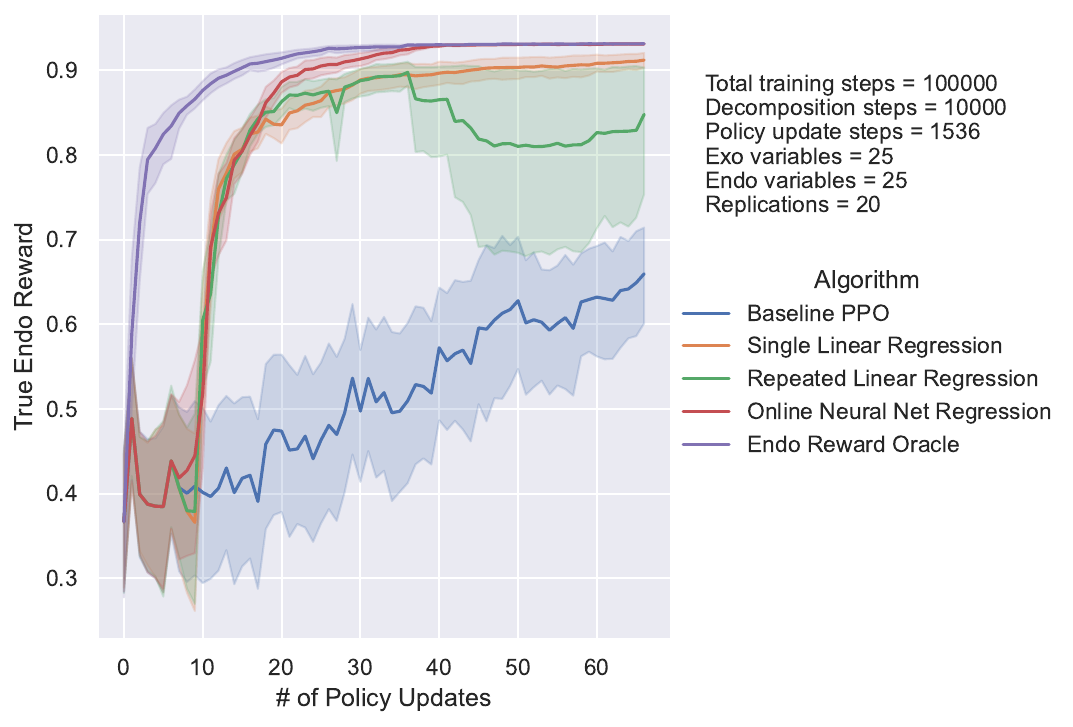}
         \caption{\sras{} ($m=25,n=25$).}
         \label{fig:25x25-stepwise}
     \end{subfigure}
    \caption{Impact of the type of exo reward regression on RL performance.}
    \label{fig:exo_reward_regression}
\end{figure}

We plot RL performance over the 20 replications in Figure \ref{fig:linear_global_stepwise} on two MDPs: (i) the 10-D state MDP (Figures \ref{fig:5x5-global}-\ref{fig:5x5-stepwise}) and (ii) the 50-D state MDP (Figures \ref{fig:25x25-global}-\ref{fig:25x25-stepwise}). We compare the three regression methods applied to \grds{} and \sras{}. The results show that Online Neural Net Regression generally outperforms Single and Repeated Linear Regression, even though the exo reward function $R_{exo,t}$ is a linear function of the exo state. We speculate that this is because it is continually incorporating new data, which in turn may allow PPO to make more progress.  Repeated Linear Regression also incorporates new data, but at a slower rate. Furthermore, when applied to \sras{}, it becomes unstable and exhibits high variance.  Future work might consider imposing strong regularization to improve stability.

Based on the superior performance of online neural network regression, we adopted it as the default reward regression method throughout the paper with the exception of Section~\ref{sec:main-experiments}

\section{The Simplified Objective}
\label{app:simplified_setting}

Our experiments showed that the Simplified-\grds{} and \sras{} algorithms often give excellent performance. Both of these algorithms employ the simplified information theoretic objective $I(X'; A\mid X)=0$ in place of $I(X'; A, E\mid X)=0$. In this appendix, we analyze the properties of the simplified objective.

\begin{figure}[t!]
    \centering
    \includegraphics[scale=0.8]{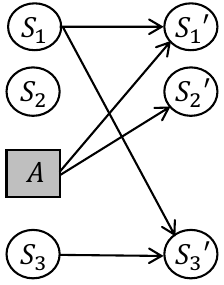}
         \caption{MDP with 3 endogenous state variables $S_1,S_2,S_3$.}
         \label{fig:action-setting-a}
\end{figure}

It is easy to find examples where the simplified objective fails. Consider the DBN in Figure \ref{fig:action-setting-a} of an MDP with the state variables $S_1, S_2$ and $S_3$. Suppose the policy determining the action is random and does not depend on any state, which is why $A$ has no incoming links. 
All state variables can be endogenous ($S_1$ and $S_2$ directly, and $S_3$ indirectly through its dependence on $S_1$).
However, the MDP satisfies the condition $I(S'_3; A\mid S_3)=0$. Hence, we cannot use the simplified objective to safely conclude that a state variable is exogenous when analyzing the 2-time step DBN.

\begin{figure}[b!]
     \centering
     \begin{subfigure}[t]{0.3\textwidth}
         \includegraphics[scale=0.8]{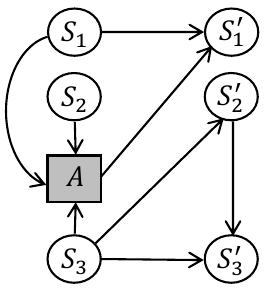}
         \caption{$I(S'_3 ; A \mid S_3)=0$.}
         \label{fig:interesting-setting-4}
     \end{subfigure}
     \hspace{0.15\textwidth}
     \begin{subfigure}[t]{0.3\textwidth}
     	 \centering
         \includegraphics[scale=0.8]{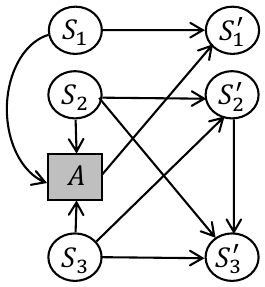}
         \caption{$I(S'_3 ; A \mid S_3)\neq 0$.}
         \label{fig:interesting-setting-2}
     \end{subfigure}
     \caption{State transition diagrams for two MDPs. In both MDPs, $S_2$ and $S_3$ are exogenous.}
     \label{fig:interesting-settings-2-3}
\end{figure}

As we would expect, the simplified setting does not necessarily lead to valid exo/endo decompositions. In this direction, first note that even when the simplified objective returns a set $X$ of variables that are causally exogenous, $X$ and its complement $E=X^c$ may not correspond to a valid exo/endo decomposition $(E,X)$. Consider for example $X=\{S_3\}$ in Figure \ref{fig:interesting-setting-4}, which satisfies $I(X' ; A \mid X)=0$ but $X$ and the corresponding $E=\{S_1,S_2\}$ are not a valid  exo/endo decomposition because $S'_3$ depends on $S'_2$. This is not the case when using the full objective. Furthermore, it is possible for a state variable $X$ to be exogenous even though it fails to satisfy $I(X' ; A \mid X)=0$. Consider $X=\{S_3\}$ in Figure \ref{fig:interesting-setting-2}. For a fixed given policy, it may not satisfy $I(S'_3 ; A \mid S_3)=0$ due to the trail $A\gets S_2\to S'_3$ from $A$ to $S'_3$. 

Despite the fact that the simplified objective $I(X'; A\mid X)=0$ is not sufficient to identify exogenous variables, it can be used as a simpler proxy for the full objective $I(X';[E,A]\mid X)=0$. Any set of state variables $X$ that satisfies the full objective must necessarily satisfy the simplified objective, since the latter has fewer constraints than the former. Of course, the simplified objective may return an over-estimate of the set of exogenous state variables, possibly contaminated with endogenous components. 
For this reason, it is always important to check the decomposition $(E=X^c,X)$ that satisfies the simplified objective against the full objective.

The shortcomings of the simplified objective result from attempting to apply it within the framework of the 2-time step DBN. If we unroll the DBN and consider the $H$-horizon MDP, then the simplified objective corresponds to directly checking that all variables in $X$ are disconnected from $A$ and therefore $X$ is causally exogenous. The next theorem resembles Theorem \ref{theorem:ecmi} for full factorizations.

\begin{theorem}\label{theorem:simplified-objective}
Assume that an $H$-horizon MDP is admissible and data $D$ has been collected by executing a fully-randomized policy for $n$ steps. If 
\begin{equation}\label{eq:simplified-constraints}
\lim_{n\to\infty}\hat{I}(X_{\tau} ; A_t \mid X_{t})=0,\forall t\leq H-1,t+1\leq\tau\leq H,
\end{equation}
then 
\begin{equation}\label{eq:simplified-constraints-2}
P(X_{\tau} \mid X_t, A_t)=P(X_{\tau} \mid X_t), \forall t\leq H-1,t+1\leq\tau\leq H.
\end{equation}
\end{theorem}
\begin{proof} (sketch)
We can show a proof similar to the proof of Theorem \ref{theorem:ecmi} that Equation \eqref{eq:simplified-constraints} is equivalent to the statement that $P(X_{\tau} \mid X_t, A_t)=P(X_{\tau} \mid X_t)$ for all $t$ and $\tau$, as the number of samples approaches infinity. %Under the faithfulness assumption, we can infer that there is no directed path from any $A_t$ to any $X_\tau$. Hence, $X$ is action-disconnected in the causal graph $\mathcal{G}$, and it follows from Theorem~\ref{theorem:generic-causal-exogeneity} that $X$ is causally exogenous.
\end{proof}
Equations \eqref{eq:simplified-constraints-2} are powerful in the context of fully randomized policies, because they imply that the action $A$ cannot have any impact on the evolution of $X$. Under standard assumptions in causality theory (e.g., faithfulness in \citep{pearl1988}), we can then conclude that there are no directed paths from any $A_t$ to any future $X_{\tau}$, implying that $X$ is causally exogenous.

Theorem \ref{theorem:simplified-objective} has the drawback that it must consider all long-range dependencies, and this requires estimating $P(X_t,X_\tau,A_t)$ for all $t \leq H-1$ and all $\tau$ such that $t+1 \leq \tau \leq H$. This makes much less effective use of the data set $D$. In practical applications, it might be fruitful to explore approximate variants of this theorem that employ only a small number $F$ of forward steps. In this case, we could enforce just $F$ constraints per time step, i.e., $I(X_{t+k} ; A_t \mid X_{t})=0,\forall 1\leq k\leq F$, where $F$ is a small number.

%%%%%%%%%%%%%%%%%%%%%%%%%%%%%%%%%%%%%%%%%%%%%%%%%%%%%%%%%%%%%%%%%%%%%%%
\section{Practical Considerations}
\label{app:practical-considerations}

In this section, we switch our attention to practical aspects of our proposed methods. Our goal is to determine how to set the hyperparameters of our method.

\subsection{Impact of Hyperparameters on the Baseline}

First, we investigate the impact of hyperparameters on the Baseline. For this purpose, we consider the High-D linear setting of Section \ref{sec:linear-setting}. Recall that for this setting we used the default PPO and Adam hyperparameters in stable-baselines3, summarized in Table \ref{table:hyperparams}. We now ask  whether the performance of Baseline can be improved with different values for the hyperparameters. If that were the case, then a more careful hyperparameter tuning could be an alternative to our proposed algorithms. 

\begin{figure}[t!]
     \centering
     \begin{subfigure}[t]{0.45\textwidth}
         \includegraphics[scale=0.4]{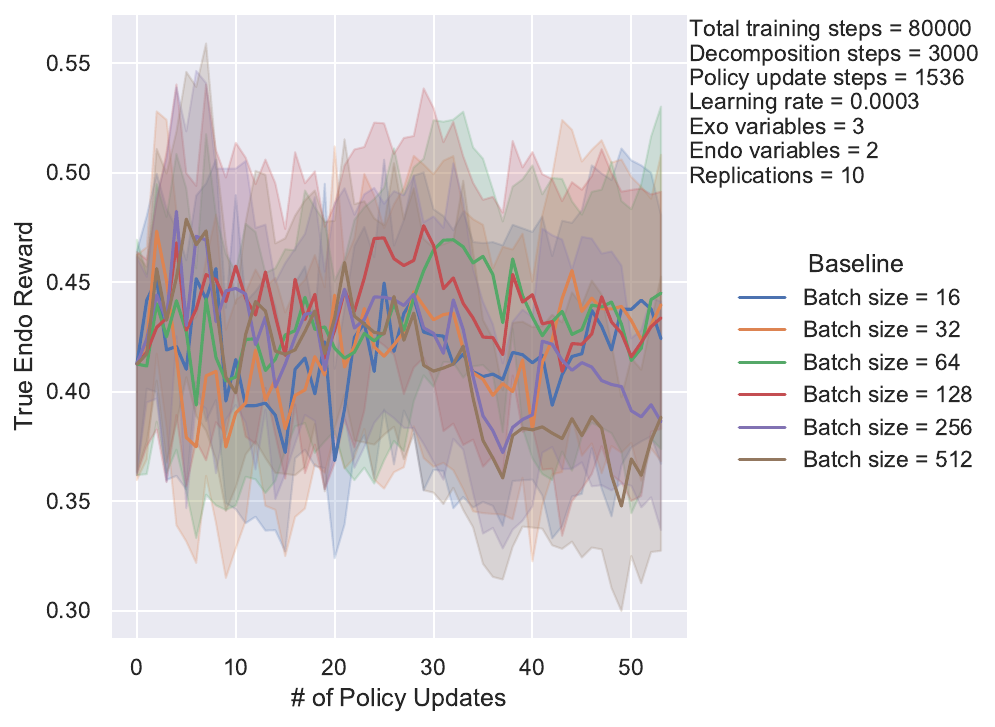}
         \caption{Tuning batch size.}
         \label{fig:baseline-batch-size}
     \end{subfigure}
     \hspace{0.0\textwidth}
     \begin{subfigure}[t]{0.45\textwidth}
         \includegraphics[scale=0.4]{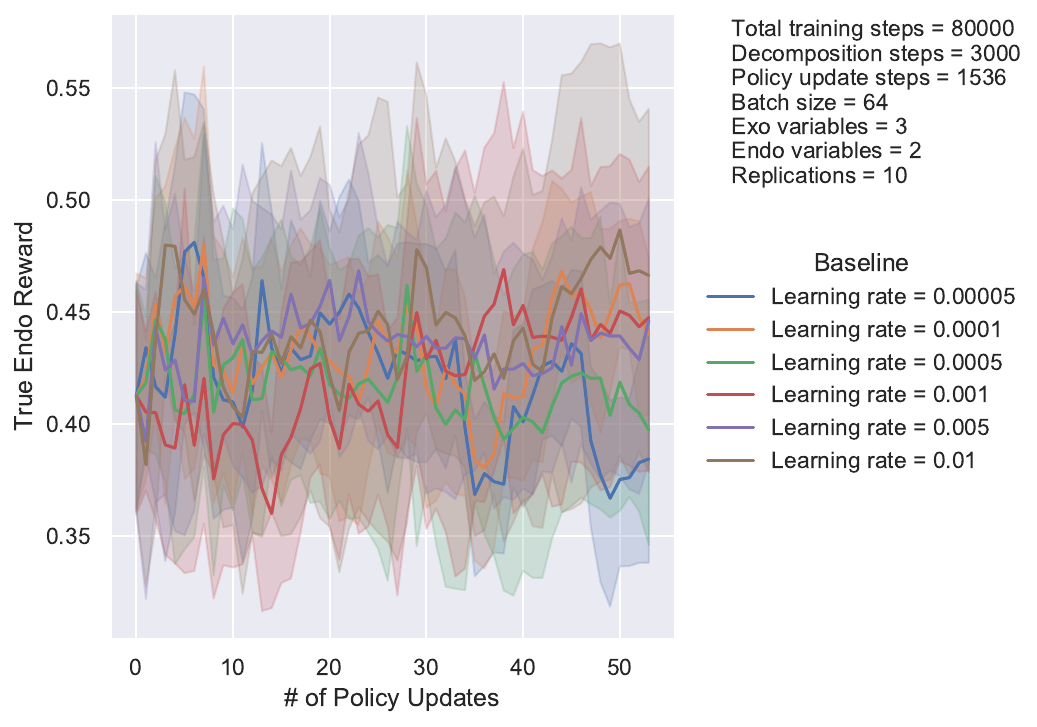}
         \caption{Tuning learning rate.}
         \label{fig:baseline-learning-rate}
     \end{subfigure}
    
    \begin{subfigure}[t]{0.45\textwidth}
         \includegraphics[scale=0.4]{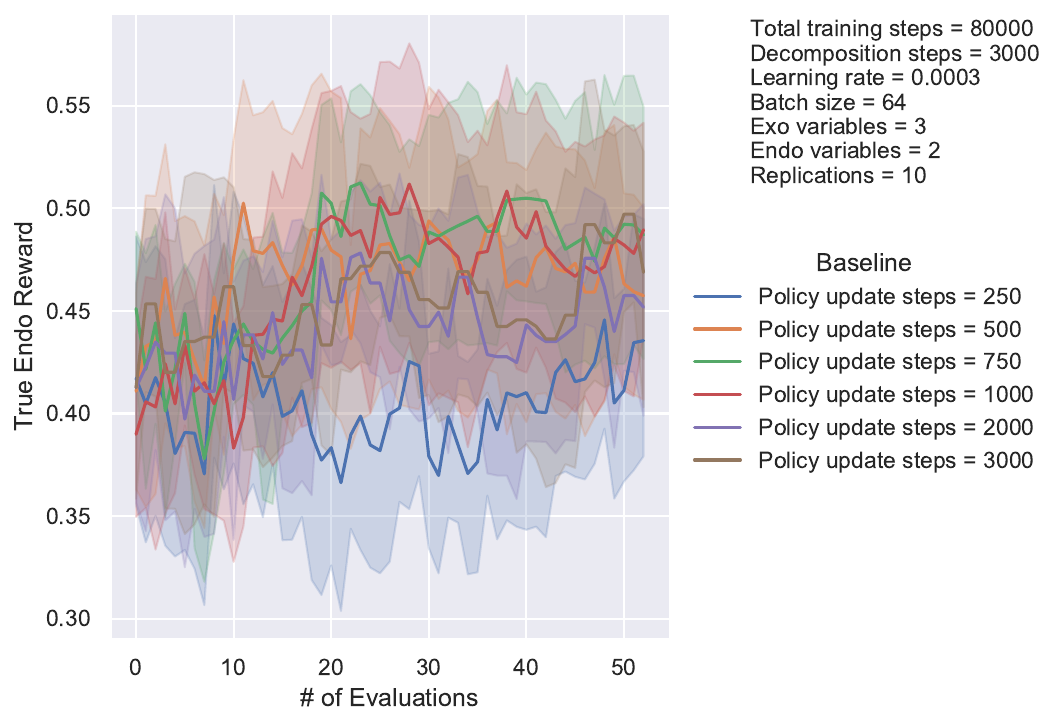}
         \caption{Tuning policy update steps $K$.}
         \label{fig:baseline-policy-update-steps}
     \end{subfigure}
    \caption{RL performance for Baseline of Figure \ref{fig:3x2} under various hyperparameters.}
    \label{fig:baseline-hyperparams}
\end{figure}

We consider the 5-D MDP with 3 exo and 2 endo variables of Figure \ref{fig:3x2}, where the Baseline fluctuates between 0.4 and 0.5 with an average of 0.45, visibly lower than all other methods. To understand whether this can be improved further, we tune 3 critical hyperparameters of PPO optimization. We let the batch size take values in $\{16, 32, 64, 128, 256, 512\}$ (Figure \ref{fig:baseline-batch-size}), the learning rate take values in $\{5\times 10^{-5}, 1\times10^{-4}, 5\times 10^{-4}, 1\times 10^{-3}, 5\times 10^{-3}, 1\times 10^{-2}\}$ (Figure \ref{fig:baseline-learning-rate}), and the number of steps per policy update $K$ take values in $\{250, 500, 750, 1000, 2000, 3000\}$ (Figure \ref{fig:baseline-policy-update-steps}). For each experiment, all hyperparameters except for the tuned one are set to the values in Figure~\ref{fig:3x2}. For Figure~\ref{fig:baseline-policy-update-steps}, we perform policy evaluation every $1536$ steps, instead of after each policy update, to ensure that all curves have the same number of evaluation points. We use 10 independent replications with different seeds. Finally, we increase the number of training steps to $N=80000$ to ensure that the Baseline has enough training budget to converge.

The results demonstrate that none of the parameter combinations can raise performance significantly. Some values for the number of policy update steps manage to slightly improve the average Baseline performance to 0.50 from 0.45, but they still suffer from significant variance. This is in sharp contrast to our algorithms and the Endo Reward Oracle in Figure \ref{fig:3x2}, which all exhibit much lower variance. Given that this experiment only tunes one hyperparameter at a time, we cannot exclude the possibility that there are combinations of hyperparameters that can achieve a higher and more stable performance for the Baseline. However, its low performance under the default hyperparameters and on a range of reasonable values provides evidence that its poor performance mainly stems from the stochasticity of the exogenous rewards and not badly-chosen hyperparameters. 
%This further signals the significance of our methods, which can match or outperform the Endo Reward Oracle without the need hyperparameter optimization.

\subsection{Sensitivity Analysis}
\label{sec:sensitivity-analysis}

\begin{figure}[t!]
     \centering
     \begin{subfigure}[t]{0.45\textwidth}
         \includegraphics[scale=0.4]{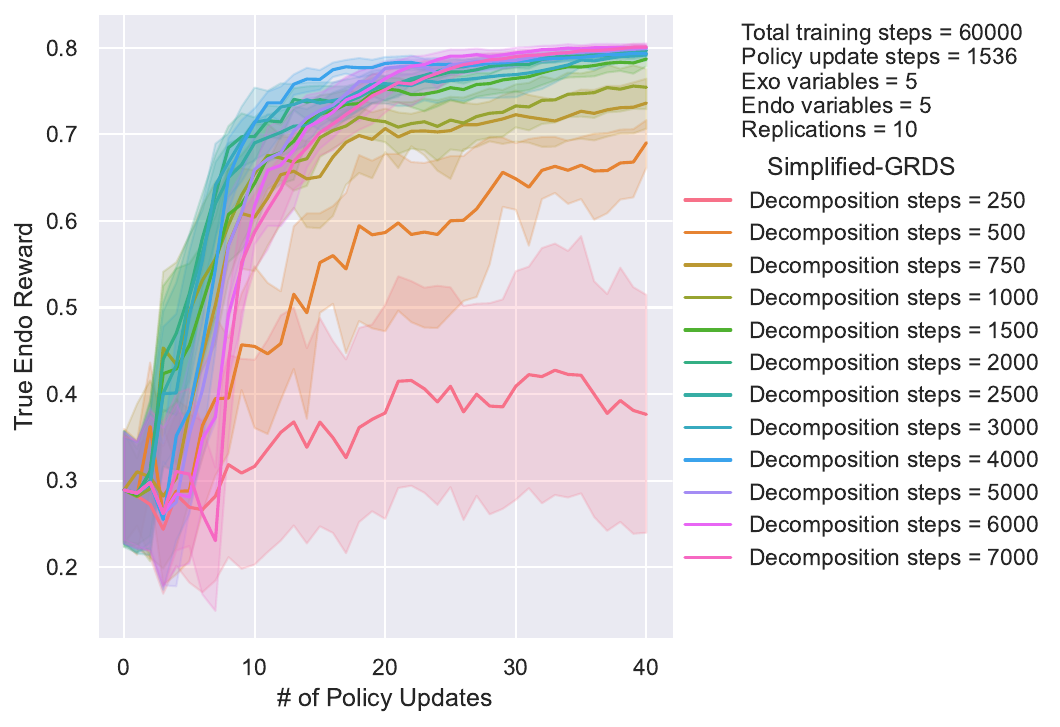}
         \caption{Simplified-\grds{}.}
         \label{fig:sensitivity_analysis_global}
     \end{subfigure}
     \hspace{0.0\textwidth}
     \begin{subfigure}[t]{0.45\textwidth}
         \includegraphics[scale=0.4]{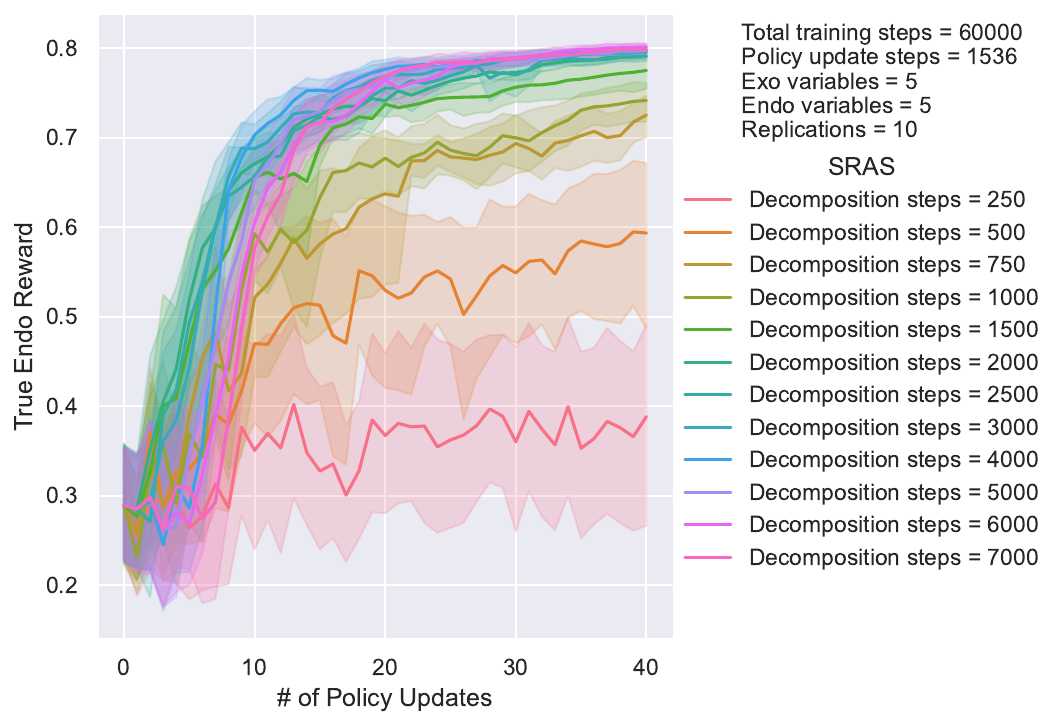}
         \caption{\sras{}.}
         \label{fig:sensitivity_analysis_stepwise}
     \end{subfigure}
     \caption{RL performance for varying values of $L$, the number of steps prior to applying the decomposition algorithms, for the 10D setting of Figure \ref{fig:5x5}.}
    \label{fig:sensitivity_analysis}
\end{figure}

Recall from Section \ref{sec:setup} that the main hyperparameter we need to set for our proposed methods is the number of steps $L$ prior to applying the exogenous subspace discovery algorithms. This specifies the number of $\langle s, a, r, s'\rangle$ tuples that are collected for subspace discovery. In this section, we shed light on the impact of $L$ on RL performance. In this direction, we consider the 10-D setting with 5 exo and 5 endo variables of Figure \ref{fig:5x5}, where our methods and the Endo Reward Oracle converge to a total reward of 0.8 in $N=50000$ steps. In contrast, the Baseline only attains a reward of around 0.3 on average. We perform sensitivity analysis by considering twelve possible values for $L$: $250, 500, 750, 1000, 1500, 2000, 2500, 3000, 4000, 5000, 6000, 7000$. We denote each value by the increasing sequence $L_i,i\in\{0,\dots,11\}$, where $L_0=250$ and $L_{10}=7000$. We report the average RL performance for the Simplified-\grds{} and \sras{} methods in Figure \ref{fig:sensitivity_analysis}(a,b), since they both perform very well. For both methods, when $L_0=250$, performance is barely better than the Baseline. As we increase $L$, performance improves steadily and almost matches the Endo Reward Oracle as soon as we reach 2000 decomposition steps. Increasing $L$ beyond 2000 gives minor benefit and delays the time at which PPO can take advantage of the improved reward function. 

This suggests that the decomposition algorithms have converged after $L=2000$. Can we verify this? The rank of the discovered exogenous space is not informative, because it is always 9-dimensional for all 12 values of $L$ and across all 10 replications. To get a finer-grained measure, we can take advantage of the fact that the complement of the discovered 9-dimensional exogenous space is a 1-dimensional space. This means it can be represented by a direction vector, and we can compare different solutions by computing the angles between these direction vectors. 

\begin{figure}[t!]
     \centering
         \includegraphics[scale=0.5]{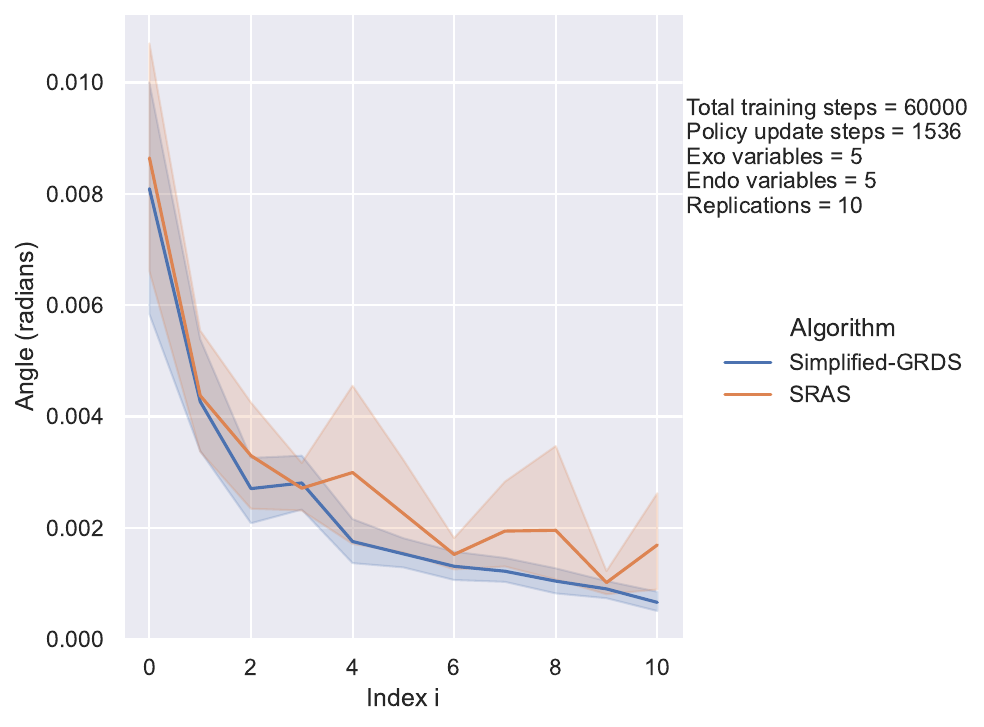}
    \caption{Angle between orthogonal complements of computed exo subspaces corresponding to $L_i$ and $L_{i+1}$ decomposition steps, where the index $i$ ranges from 0 to 10.}
    \label{fig:angle_deltas}
\end{figure}

Figure \ref{fig:angle_deltas} depicts the angle (in radians) between the orthogonal complements of the exogenous subspaces for each consecutive pair of $L_i$ and $L_{i+1}$ values. If the methods had converged, these angles would be zero. They are not zero, which shows that the exogenous subspaces are continuing to change as $L$ increases. But the angles are all very small (less than 0.5 degrees in the largest case), so these changes are not large, and they are converging (with few exceptions) monotonically toward zero. Simplified-\grds{} exhibits smooth convergence, whereas \sras{} shows higher variance and a few bumps. 

The results suggest that manifold optimization performs very well on this problem, even with relatively small numbers of samples. What is then the reason for the different performance levels in Figure \ref{fig:sensitivity_analysis}? The answer lies in the exogenous reward regression. Recall that after $L$ steps, we conclude Phase 1 by computing the decomposition and then fitting the exogenous reward neural net to the $L$ collected observations. Even though different numbers of decomposition steps result in almost identical exogenous subspaces, the subsequent exogenous reward regression can yield dramatically different exo reward models. When the value of $L$ is very low, we have only a limited number of samples for the exo reward regression, and these might not to cover the exo state subspace adequately. As a result, the learned exo reward model may overfit the observations and fail to generalize to other subspaces. The subsequent online neural network reward regression in Phase 2 only processes each new observation once, so learning the correct exo reward model can take many steps, and cause PPO to learn slowly.

To confirm the above, we perform a second experiment. Unlike previous experiments, we decouple state decomposition and exogenous reward regression. Decomposition still takes place after $L_i$ steps. But reward regression is now performed after 3000 samples have been collected (3000 is the default value for $L$ in the high-D experiments). After learning the exogenous reward model with the 3000 samples, we proceed to Phase 2. We experiment with three options for exogenous reward regression in Phase 2: (i) standard online learning where we update the exogenous reward model every $M=256$ steps; (ii) a single linear regression after which the exogenous reward model is never updated; and (iii) repeated linear regression where we fit a new exogenous reward model from scratch every $M=256$ steps. We study the three lowest values for $L_i$ ($L_0=250, L_1=500$ and $L_2=750$), as these were the values in Figure \ref{fig:sensitivity_analysis} with the worst performance.

\begin{figure}[t!]
     \centering
     \begin{subfigure}[t]{0.45\textwidth}
         \includegraphics[scale=0.4]{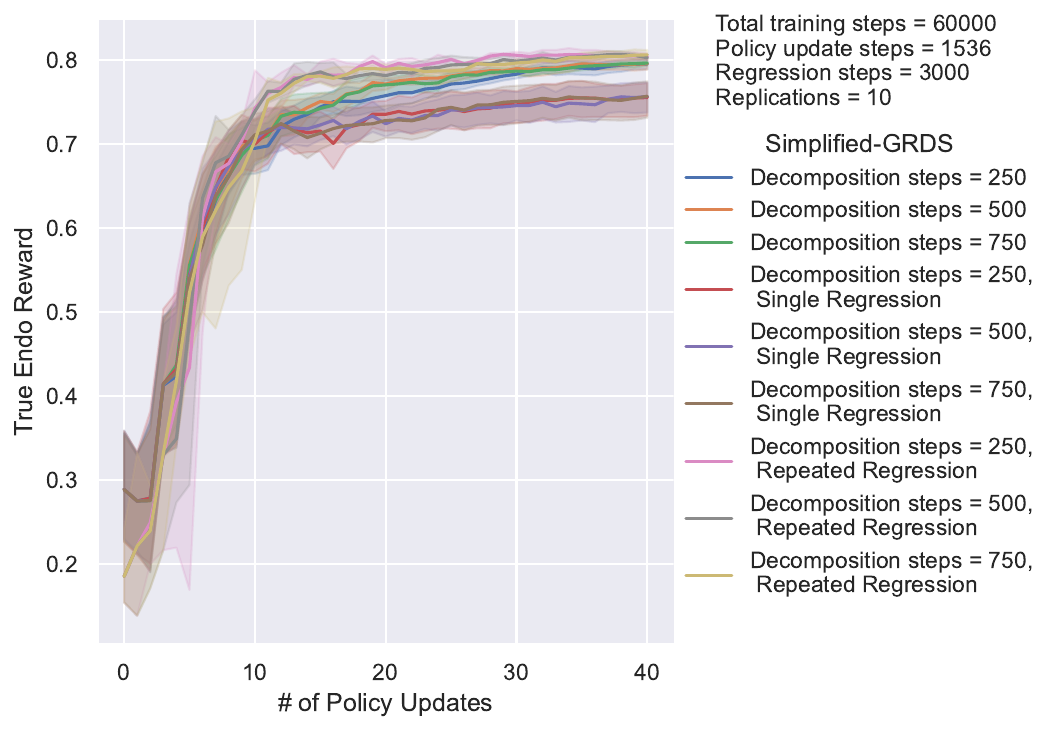}
         \caption{Simplified-\grds{}}
         \label{fig:special_global}
     \end{subfigure}
     \hspace{0.0\textwidth}
     \begin{subfigure}[t]{0.45\textwidth}
         \includegraphics[scale=0.4]{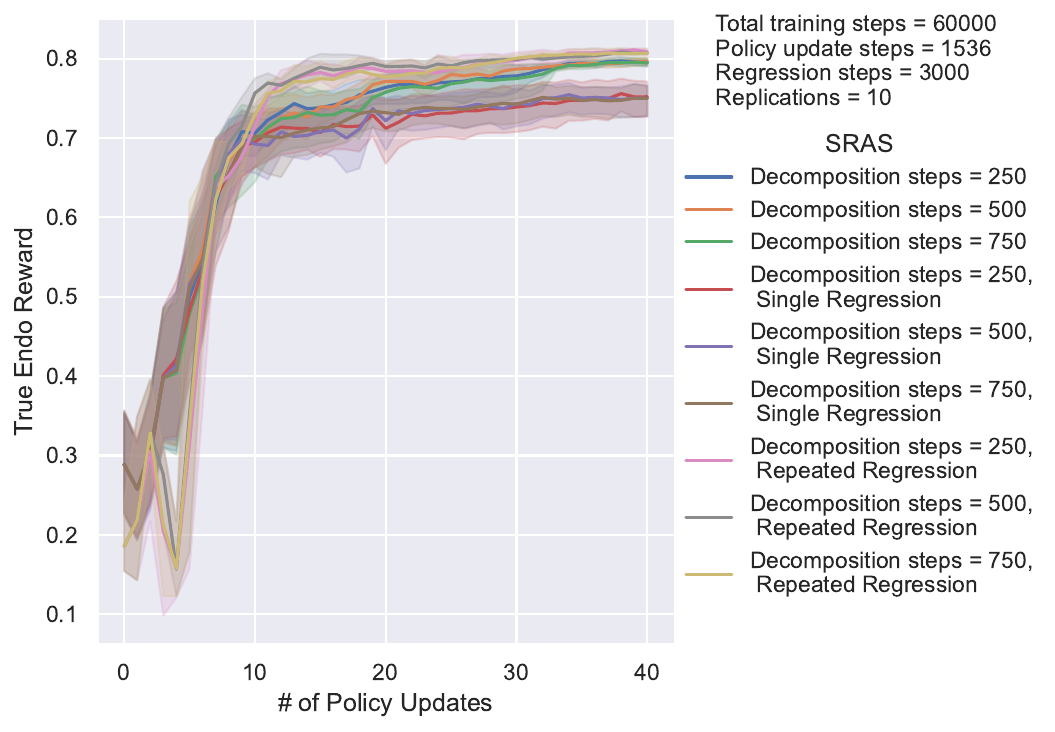}
         \caption{\sras{}}
         \label{fig:special_stepwise}
     \end{subfigure}
    \caption{RL performance when computing the exo/endo decomposition at $L_i = 250, 500,$ and $750$ steps. Reward regression starts at 3000 steps. Default: online neural network regression; Single Regression: single linear regression at 3000 steps; Repeated Regression: linear regression every 256 steps.}
    \label{fig:sensitivity_analysis_2}
\end{figure}

Figure \ref{fig:sensitivity_analysis_2} plots the RL performance averaged over 10 independent trials. We make several observations. First, increasing the number of steps for learning the initial exogenous reward model from $L_i$ to 3000 improves RL performance. With online learning, we match the Endo Reward Oracle's average performance of 0.8. This confirms our previous hypothesis that the reason for the bad performance in Figure \ref{fig:sensitivity_analysis} was the poor initial exogenous reward model. With a single regression, RL performance improves (especially for $L_0=250$ and $L_1=500$), but it performs worse than online learning while having greater variance. 
%This is because online learning can continuously adapt the exogenous reward function to the changing distribution of the observed states and rewards.
%Note that RL performance visibly improves and is able to match the Endo Reward Oracle's average performance of 0.8. This is true for both schemes. This implies that the value for the decomposition steps $L$ must be high enough to ensure that we can both perform an accurate state decomposition as well as learn a good initial exo reward model.
Interestingly, learning a new linear regression model every $M=256$ steps performs the best and slightly outperforms online neural network regression. A plausible reason for this is that online regression only performs a single pass over the data, so it may adapt to the changing state and reward distribution more slowly. On the other hand, repeated linear regression requires 10 times as much computation time as online regression for the settings in Figures \ref{fig:special_global} and \ref{fig:special_stepwise}.

\subsection{Practical Guidelines For State Decomposition and Exo Reward Regression}
This sensitivity analysis suggests the following procedure for computing state space decompositions and reward regressions. Start with a small number for $L$ (e.g., 250) and compute the corresponding exo subspace after $L$ steps. Then, every $\Delta L$ steps (e.g., 250), recompute the exo subspace until the discovered exo subspace stops changing. This can be detected when (i) the rank of the subspace does not change, and (ii) the largest principal angle between consecutive subspaces is close to 0 \citep{principal-angles}. 

As soon as we have an initial exo subspace, we can fit the exogenous reward model, and each time we recompute the subspace, we can re-fit the model. These initial fits could be performed with linear regression or neural network regression. Once the exogenous subspace as converged, we can switch to online neural network regression, because the regression inputs will have stabilized. 

Application constraints may suggest alternative procedures. If each step executed in the MDP is very expensive, then the cost of the multiple decomposition and reward regression computations is easy to justify. However, if MDP transitions are cheap, then we need to take a different approach. If many similar MDPs will need to be optimized, we can use this full procedure on a few of them to determine the value of $L$ at which the exo space converges. We then just use that value to trigger exo/endo decomposition and perform neural network reward regression starting at step $L$ and continuing online. If there is only one MDP to be solved, $L$ could be selected using a simulation of the MDP. In all of our experiments, we have followed this procedure for setting $L$. 

A last question concerns the value for the CCC threshold $\epsilon$. In theory, we should use very low values to minimize the chance of discovering an invalid exo/endo state decomposition, but this could come at the cost of having to perform more steps in \grds{} and \sras{}. We lack theoretical guidance for making this decision. In principle, one could start with a somewhat large value for $\epsilon$ (e.g., 0.1) and perform multiple runs (in simulation or on a sample of MDPs) with progressively smaller values of $\epsilon$ until either the discovered exo subspace converges or RL performance stabilizes.

\bibliographystyle{plainnat}

\end{document}